\documentclass{article}

\usepackage{iclr2027_conference,times}
\usepackage[T1]{fontenc}

\usepackage{amsmath,amsfonts,bm}

\def\eqref#1{equation~\ref{#1}}

\def\1{\bm{1}}

\DeclareMathAlphabet{\mathsfit}{\encodingdefault}{\sfdefault}{m}{sl}
\SetMathAlphabet{\mathsfit}{bold}{\encodingdefault}{\sfdefault}{bx}{n}

\usepackage{amsmath}
\usepackage{amssymb}
\usepackage{booktabs}
\usepackage{multirow}
\usepackage{xcolor}
\usepackage{colortbl}
\usepackage{listings}
\usepackage{graphicx}
\usepackage{wrapfig}
\usepackage{needspace}
\usepackage{hyperref}
\hypersetup{
    hidelinks,
    pdftitle={RoShift: Lazy Shift and Weight-aware Rotation for Activation Sparsity in LLM Inference},
    pdfauthor={Geunsoo Song, Jiwon Seo, Hoeseok Yang, Youngmin Yi}
}
\usepackage{microtype}
\usepackage{subcaption}
\DeclareCaptionFont{tablefont}{\fontsize{9}{10.8}\selectfont}
\usepackage{pdflscape}
\usepackage{algorithm}
\usepackage[noend]{algpseudocode}
\usepackage{url}
\usepackage[normalem]{ulem}

\newcommand{\ymyi}[1]{{\color{black} #1}}
\newcommand{\ymyj}[1]{{\color{black} #1}}
\newcommand{\sgs}[1]{{\color{black} #1}}
\newcommand{\hyang}[1]{{\color{black} #1}}
\newcommand{\revision}[1]{{\color{black}#1}}

\title{RoShift: Lazy Shift and Weight-aware Rotation for Activation Sparsity in LLM Inference}

\author{
Geunsoo Song$^{1}$ \quad Jiwon Seo$^{1}$ \quad Hoeseok Yang$^{2}$\thanks{Co-corresponding authors: \texttt{hoeseok.yang@scu.edu} and \texttt{ymyi@sogang.ac.kr}.} \quad Youngmin Yi$^{1}$\footnotemark[1]\\
{\normalfont $^{1}$Sogang University \quad $^{2}$Santa Clara University}
}

\iclrfinalcopy

\begin{document}

\maketitle
\lhead{Preprint}

\begin{abstract}
Activation sparsity accelerates LLM decoding by skipping unnecessary weight reads and computation.
However, SiLU-based LLMs exhibit little natural sparsity, and imposing high sparsity can substantially degrade accuracy.
We propose RoShift, a training-free method that transforms activations into sparsity-friendly representations.
RoShift introduces lazy shift, which recenters SiLU activation distributions around zero and defers correction to the end of the FFN to obtain additional sparsity.
It also reduces sparsification-induced output error through rotation that jointly accounts for activation and weight information.
By fusing the corrections required for shifting and rotation, RoShift translates these gains into decoding speedup.
At 50\% and 60\% sparsity, RoShift improves average accuracy over TEAL by 1.60 and 3.29 percentage points, respectively, across five models. It achieves up to $1.72\times$ decoding speedup over the dense model on an RTX 5090 at 50\% sparsity and yields a better accuracy-speedup Pareto frontier than TEAL. Code is available at \url{https://github.com/hulahula3247/RoShift}.

\end{abstract}

\section{Introduction}

Large language models (LLMs) achieve strong performance across diverse tasks, but their large parameter counts make inference expensive. Autoregressive decoding generates tokens sequentially within each sequence and repeatedly reads model weights at every step. In particular, \ymyi{at} batch size 1, limited parallelism and weight reuse often make memory bandwidth the performance bottleneck. These characteristics make efficient decoding challenging in edge and on-device settings, where a single device serves one user's requests, as well as on \ymyi{Physical AI platforms}.

Activation sparsity offers a way to reduce these weight reads and computations. In a linear layer, a zero activation makes the corresponding weights irrelevant to the output, allowing their memory accesses and computations to be skipped. ReLU-based models naturally produce many zero activations and can therefore \ymyi{readily} exploit this property, whereas modern LLMs using SiLU \hyang{or GELU} rarely produce exact zeros. Methods such as CATS and TEAL approximate small-magnitude activations as zero to induce sparsity~\citep{lee2024cats,liu2025teal}. At high sparsity, however, this can remove components that contribute substantially to the output, leading to greater accuracy degradation.

Recent studies use rotation to concentrate activation energy in a subset of dimensions~\citep{ashkboos2024slicegpt, liu2025larosa}. \sgs{In particular, LaRoSA applies such rotations to activation sparsification to reduce sparsification error.} Meanwhile, WAS and WINA select components for removal using both activation magnitudes and the importance of the corresponding weights~\citep{wang2025was,chen2026wina}. Combining these approaches introduces a new challenge: rotation changes both activations and their corresponding weights, affecting not only which components are removed but also how their removal changes the output. A sparsity-friendly rotation should therefore account for both the activation distribution and the influence of weights on removal error.

We also observe that \hyang{the channel-wise activation distributions in SiLU-based models are centered away from zero, limiting the components eligible for magnitude-based sparsification around zero.} 
Shifting \ymyi{each channel} \hyang{by a fixed offset} \ymyi{recenters} \hyang{them near zero}, \ymyi{placing} \hyang{more elements within the sparsification threshold}, enabling higher activation sparsity \ymyi{at comparable quality}. 
\sgs{SCAP exploits this property through mode-centering in non-GLU FFNs}~\citep{chua2024scap}.
\sgs{In gated FFNs, however, restoring the original gate activations before multiplying them by the Up outputs prevents the Up and Down projections from exploiting the additional sparsity obtained by recentering. Moreover, preserving the original computation under representation transformations such as shifting and rotation may require additional corrections, whose cost can offset the savings from sparsification.}
We therefore need transformations that reduce sparsification-induced output error while limiting correction overhead sufficiently to deliver practical inference acceleration.

To \ymyi{address these challenges}, we propose RoShift, a training-free activation sparsification method. RoShift recenters post-SiLU activations, while lazy shift defers restoration to the end of the FFN so that the Up and Down projections can exploit the additional sparsity. RoShift also constructs rotations that jointly account for activation statistics and the influence of weights on removal error, reducing sparsification-induced output error. Finally, it fuses the corrections required for shifting and rotation and optimizes sparse kernels to translate the resulting sparsity into practical decoding acceleration.

At 50\% and 60\% non-uniform sparsity, RoShift improves average accuracy over TEAL by 1.60 and 3.29 percentage points, respectively, across five models. It achieves up to $1.72\times$ decoding speedup over the dense model on an RTX 5090 at 50\% sparsity and yields a better accuracy-speedup Pareto frontier than TEAL.

Our main contributions are as follows.

\begin{enumerate}
    \item Lazy shift: We recenter post-SiLU activations and defer the correction to the end of the FFN, providing additional sparsity that the Up and Down projections can exploit.
    \item Activation- and weight-aware rotation: We \ymyi{jointly incorporate} the activation distribution and weight information to construct \sgs{rotations} that reduce sparsification-induced output error.
    \item Correction fusion and efficient GPU execution: We \ymyi{effectively fuse the two correction operations to reduce overhead and use efficient sparse kernels} to \sgs{translate these savings into} practical decoding acceleration.
\end{enumerate}

\section{Background \& Observation}
\label{sec:background}

\subsection{Background: Activation Sparsity}
Activation sparsity \sgs{can} \ymyi{accelerate inference} by skipping weight reads and computations associated with components that do not contribute to the output. We review prior work on activation sparsity in Appendix~\ref{app:related-work}.

Activation sparsity is primarily exploited through input sparsity and output sparsity. Input sparsity exploits zeros in the activations entering a projection. For a linear projection with input activations $X\in\mathbb{R}^{1\times d_{\mathrm{in}}}$ and weights $W\in\mathbb{R}^{d_{\mathrm{in}}\times d_{\mathrm{out}}}$, the output is $Y=XW\in\mathbb{R}^{1\times d_{\mathrm{out}}}$. If $X_i=0$, the corresponding weight row $W_{i,:}$ does not contribute to the output, so its memory accesses and computations can be skipped. In SiLU-based models, where natural zeros are rare, thresholding or Top-K selection induces sparsity by approximating small-magnitude activations as zero.

Output sparsity, in contrast, identifies output dimensions expected to contribute little or nothing to subsequent computations and skips the corresponding weight columns $W_{:,j}$. Let $d$ denote the model dimension and $k$ the FFN intermediate dimension. A gated FFN computes the following for input $X$:
\begin{equation}
    h_1=\sigma(XW_{\mathrm{gate}}),\quad
    h_2=XW_{\mathrm{up}},\quad
    h_3=h_1\odot h_2,\quad
    Y_{\mathrm{FFN}}=h_3W_{\mathrm{down}},
    \label{eq:gated-ffn-background}
\end{equation}
Here, $W_{\mathrm{gate}},W_{\mathrm{up}}\in\mathbb{R}^{d\times k}$ and $W_{\mathrm{down}}\in\mathbb{R}^{k\times d}$. Here, $\sigma$ is an activation function such as SiLU and $h_1$ gates the channel-wise Up outputs. If $h_{1,j}=0$, \ymyi{the computation for Up output} $h_{2,j}$ and \ymyi{the contribution from} the corresponding Down input \ymyi{$h_{3,j}$} can \ymyi{both} be skipped; thus, Up output sparsity becomes Down input sparsity. \sgs{CATS exploits this output sparsity by selecting channels according to the magnitudes of the gate activations $h_1$}~\citep{lee2024cats}.

\sgs{Magnitude-based input sparsification, such as TEAL, selects components for removal based on their input activation magnitudes $|X_i|$}~\citep{liu2025teal}. Removing a component $X_i$, however, removes the output contribution $X_iW_{i,:}$, whose magnitude is $\|X_iW_{i,:}\|_2=|X_i|\cdot\|W_{i,:}\|_2$. Consequently, components with identical activation magnitudes can induce different sparsification errors depending on their corresponding weights. WAS and WINA incorporate both activation and weight information into their sparsification criteria~\citep{wang2025was,chen2026wina}.

Another approach to sparsification is to change the activation basis to obtain a sparsity-friendly representation. Concentrating activation energy in a subset of dimensions through rotation can reduce the energy discarded by Top-K at a fixed sparsity. For an orthogonal matrix $Q\in\mathbb{R}^{d\times d}$ \ymyi{that defines this change of basis}, Eq.~(\ref{eq:rotation-equivalence}) shows that jointly transforming activations and weights preserves the original linear projection output.

{
\setlength{\abovedisplayskip}{0pt}
\setlength{\abovedisplayshortskip}{0pt}
\setlength{\belowdisplayskip}{2pt}
\begin{equation}
    Y=XW=(XQ)(Q^\top W)
    \label{eq:rotation-equivalence}
\end{equation}
}

Within a Transformer block, the channel-wise RMSNorm scales and rotation can be absorbed into the relevant weights offline. Maintaining the residual in the form $XQ$ allows computation in the rotated basis without a separate runtime rotation wherever the same $Q$ is used. LaRoSA constructs layer-wise PCA rotations from calibration activation statistics to concentrate activation energy in a subset of dimensions. When different layers use different $Q_l$, however, the bases must be aligned between layers. Transforming $XQ_l$ into $XQ_{l+1}$ requires an additional operation with the transition matrix \revision{$T_l=Q_l^\top Q_{l+1}\in\mathbb{R}^{d\times d}$}~\citep{liu2025larosa}.

\subsection{Motivation 1: Off-center SiLU Activation Distributions}
\label{sec:motivation-shift}
\begin{figure*}[t]
    \centering
    \includegraphics[width=\textwidth]{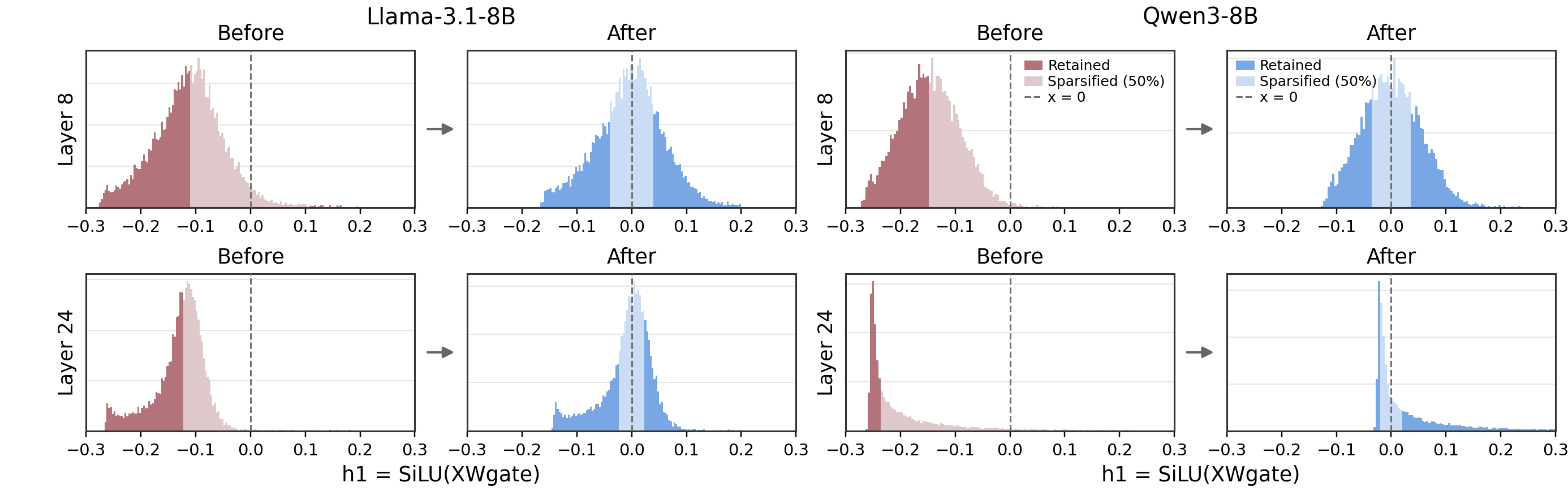}
    \captionsetup{font=small,skip=3pt}
    \caption{Activation distributions of post-SiLU $h_1$ before and after recentering. Light shading denotes the 50\% of components removed by magnitude in each representation; dark shading denotes retained components. Dashed lines indicate zero.}
\label{fig:activation-recentering}
\end{figure*}

Unlike ReLU, SiLU produces nonzero outputs for most inputs and therefore exhibits little natural sparsity. Nevertheless, these activations have asymmetric, off-center distributions. \hyang{To center each channel's distribution at zero, we compute the channel-wise median $h_{1\mathrm{med}}\in\mathbb{R}^{1\times k}$ from calibration activations and subtract it from $h_1$, yielding $\widetilde h_1=h_1-h_{1\mathrm{med}}$.} \hyang{These channel-wise medians vary little across calibration samples, allowing the estimates to serve as fixed shifts that reliably approximate the centers of activation distributions for unseen inputs.} Figure~\ref{fig:activation-recentering} shows that, even when the same 50\% of components are removed, recentering \ymyi{maps} values in densely populated regions of the \ymyi{original} distribution \ymyi{to small residuals} near zero, \ymyi{making them natural candidates for removal.} 
\setlength{\textfloatsep}{6pt}

\ymyi{Preserving the original FFN output, however,} requires a correction for the shift. If $h_1$ is restored before multiplication by $h_2$, the decisions to skip Up and Down computations are again based on the original $h_1$, preventing the additional sparsity from being exploited. To use the sparsity obtained through recentering in both Up and Down, we therefore need a correction that preserves the original FFN output without restoring the shifted activations midway through the computation.

\subsection{Motivation 2: The Need for Weight-aware Rotation}
\label{sec:motivation-rotation}
Weight-aware sparsification selects components for removal using both activations and their corresponding weights, whereas rotation transforms activations into a sparsity-friendly representation. Applying a rotation $Q$ changes the activations to $XQ$ and the weights to $Q^\top W$, so both the removed components and their influence on the output depend on $Q$.

The output error after applying weight-aware Top-K to the rotated activations is
\begin{equation}
    \color{black}
    Y_{\mathrm{error}}
    =
    \left\|
    \left(
        \operatorname{TopK}_{\mathrm{WA}}(XQ;Q^\top W)-XQ
    \right)
    (Q^\top W)
    \right\|_2^2.
    \label{eq:rotation-sparsification-error}
\end{equation}
\revision{where weight-aware Top-K (WA-Top-K) retains components with the largest scores $s_i=|(XQ)_i|\,\|(Q^\top W)_{i,:}\|_2$ at the target sparsity and sets the others to zero.} Thus, concentrating activation energy alone does not necessarily minimize sparsification error, motivating a rotation that considers both activation statistics and weights.

\section{RoShift}
\label{sec:roshift}

\vspace{-0.3\baselineskip}
\subsection{Overview}
\begin{figure}[t]
    \centering
    \includegraphics[width=\linewidth,trim=0 0 0 57,clip]{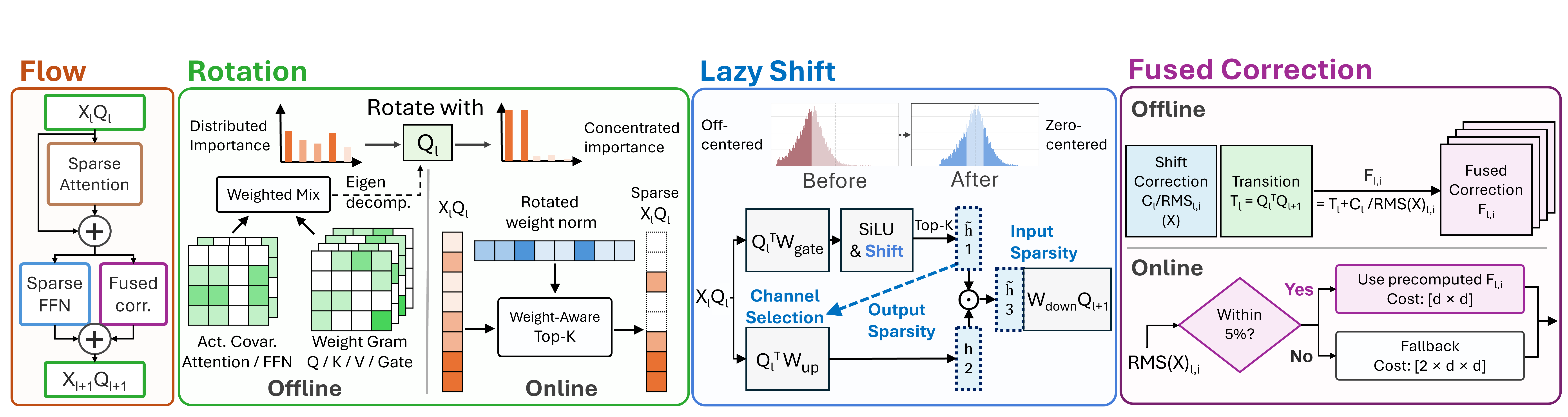}
    \captionsetup{font=small,skip=3pt}
    \caption{
    Overview of RoShift, which \ymyj{introduces lazy shift and weight-aware rotation, tightly integrating both with efficient correction fusion for activation sparsification.}
    }
\label{fig:roshift-overview}
\end{figure}

\vspace{-0.3\baselineskip}
As illustrated in Figure~\ref{fig:roshift-overview}, RoShift is a training-free method that transforms activations \ymyi{through rotation or shifting} into sparsity-friendly representations and sparsifies them to reduce weight reads and computations.
\ymyi{Both transformations require corresponding corrections to preserve the original computations.}
By \ymyi{minimizing their overhead, RoShift} translates \ymyi{activation} sparsity into practical decoding acceleration.

\sgs{During inference, the rotated residual representation enables rotation-based input sparsity for Q/K/V, Gate, and Up,
\ymyi{but} nonlinear operations prevent \ymyi{propagation} beyond these projections.
RoShift \ymyi{therefore uses} rotation-based input sparsity for Q/K/V and Gate, \sgs{while lazy shift provides Up output sparsity that becomes Down input sparsity.}
For Up,
lazy-shift output sparsity empirically achieves higher sparsity than rotation-based input sparsity, so we use the former.
}

\enlargethispage{\baselineskip}
\vspace{-0.3\baselineskip}
\subsection{Lazy Shift}
\vspace{-0.3\baselineskip}
\label{sec:lazy-shift}
Lazy shift makes activation distributions more amenable to sparsification and defers the correction to the end of the FFN, enabling Up and Down to exploit higher sparsity.
Using the per-layer channel-wise median $h_{1\mathrm{med}}$ precomputed during calibration, RoShift forms $\widetilde h_1=h_1-h_{1\mathrm{med}}$. The gated FFN in Eq.~(\ref{eq:gated-ffn-background}) can then be decomposed as follows.
\begin{align}
    Y_{\mathrm{FFN}}
      &= ((\widetilde h_1+h_{1\mathrm{med}})\odot h_2)W_{\mathrm{down}} \notag\\
      &= (\widetilde h_1\odot h_2)W_{\mathrm{down}}+XC,
      \label{eq:lazy-shift-decomposition}\\
    C &= W_{\mathrm{up}}\operatorname{diag}(h_{1\mathrm{med}})W_{\mathrm{down}}.
    \label{eq:lazy-correction-matrix}
\end{align}
\revision{Here, $\widetilde h_3=\widetilde h_1\odot h_2$ denotes the shifted Down input.} The first term in Eq.~(\ref{eq:lazy-shift-decomposition}) \ymyi{evaluates} the FFN with shifted activations, and the second corrects for the shift. The correction matrix $C\in\mathbb{R}^{d\times d}$ is \ymyi{input-}independent and can therefore be precomputed. The FFN can thus operate on shifted activations without restoring $h_1$ midway through the computation, and adding $XC$ to its output exactly compensates for the shift.

We first apply input sparsity to Gate, \ymyi{then} select Up output channels using the recentered gate magnitudes \(|\widetilde h_1|\), skipping Up computations for unselected channels. \ymyi{Because} these channels are also unused as inputs to Down, \ymyi{Up output sparsity becomes Down input sparsity.} If Down is assigned a higher target sparsity, we apply additional input sparsification. We do not sparsify the correction term \(XC\) itself, because the resulting accuracy loss is large relative to the speedup obtained.

Computing $XC$ requires reading and processing an additional $d^2$ weights. In typical gated FFNs, $k$ is several times larger than $d$, so the $d^2$ entries of $C$ are fewer than the $3dk$ weights in the three FFN projections. Appendix~\ref{app:method-details} details this overhead for each model. Nevertheless, additional sparsity does not yield decoding acceleration if its savings are smaller than the correction cost. RoShift reduces this overhead by combining the correction with the inter-layer transition and computing it with optimized kernels, as described in Section~\ref{sec:correction-fusion}.

\vspace{-0.5\baselineskip}
\begin{table}[t]
\centering
\color{black}
\captionsetup{font={tablefont,color=black},skip=5pt}
\caption{Statistics and sparsification errors for No Rotation, PCA, and AW-Rot on Llama-3.1-8B at 50\% sparsity. Values are means over the Q/K/V/Gate projections across 32 layers (128 projections). ``Leading 50\%'' denotes the first half of coordinates in descending eigenvalue order (original order for No Rotation), not the Top-K support. Removed contribution proxy is the token-average fraction of total squared activations, weighted by squared weight-row norms, removed by the actual B128 mask.}
\label{tab:rotation-statistics-diagnostics}
\begingroup
\fontsize{9}{10.8}\selectfont
\setlength{\tabcolsep}{5pt}
\renewcommand{\arraystretch}{1.1}
\begin{tabular}{@{}lrrr@{}}
\toprule
Diagnostic & No Rotation & PCA & AW-Rot \\
\midrule
Leading 50\% activation variance & 49.51\% & \textbf{83.69\%} & 80.49\% \\
Leading 50\% weight energy & 49.97\% & 58.44\% & \textbf{78.92\%} \\
Leading 50\% output-contribution variance & 49.59\% & 90.77\% & \textbf{94.85\%} \\
Removed contribution proxy & 4.41\% & 1.62\% & \textbf{0.89\%} \\
Output NMSE & 2.11\% & 1.33\% & \textbf{0.70\%} \\
\bottomrule
\end{tabular}
\endgroup
\end{table}

\subsection{Activation- and Weight-aware Rotation}
\label{sec:output-aware-rotation}
RoShift constructs sparsity-friendly layer-wise rotations $Q_l\in\mathbb{R}^{d\times d}$ by jointly considering activation statistics and the influence of weights on sparsification-induced output error.
\revision{PCA concentrates activation variance in a subset of dimensions but does not consider whether weight energy is concentrated in the same dimensions. Removing dimensions with low activation variance can therefore still incur large output error due to the corresponding weights. RoShift jointly considers activation variance and weight energy, aiming to concentrate output contributions in a subset of dimensions and reduce the error caused by sparsifying the remaining dimensions.}
Since Top-K selection changes discretely with the rotation, RoShift instead uses each axis's contribution variance to the projection output as a surrogate.

Consider the output contribution of a single rotation axis $q$. In Eq.~(\ref{eq:rotation-equivalence}), this contribution is $Y_q=(Xq)(q^\top W)$. Let $S$ denote the activation covariance and $M=WW^\top$ the weight Gram matrix. The variance of this axis's output contribution is

\vspace{-10pt}
\begin{equation}
    e(q)
    =
    \operatorname{Var}(Xq)\|q^\top W\|_2^2
    =
    (q^\top S q)(q^\top M q).
    \label{eq:rotation-output-variance}
\end{equation}
\vspace{-10pt}

Maximizing $e(q)$ under the unit-norm constraint $q^\top q=1$ yields the Lagrange stationarity condition:
\begin{equation}
    \left[(q^\top Mq)S+(q^\top Sq)M\right]q = \lambda q.
    \label{eq:rotation-stationarity}
\end{equation}

\vspace{-0.4em}
Because the coefficients \(q^\top Mq\) and \(q^\top Sq\) depend on \(q\), unlike standard PCA, no single fixed eigensystem directly solves the problem. RoShift approximates these axis-dependent coefficients with layer-shared mixture coefficients. \ymyi{Accordingly, the resulting eigensystem is a tractable surrogate for the axis-dependent objective rather than its exact closed-form optimizer.} Appendix~\ref{app:rotation-derivation} provides the full derivation.

To apply a single layer-wise rotation across Transformer projections, we combine six statistics: attention and FFN input covariances and Q/K/V/Gate weight Gram matrices. Activations are inverse-RMS normalized, with learned RMSNorm channel scales absorbed into the corresponding weights. After unit-trace normalization, the matrices $B_{l,i}$ are combined as $B_l=\sum_{i=1}^{6}\theta_{l,i}B_{l,i}$, $\theta_{l,i}\ge0$. We eigendecompose $B_l$ and use its eigenvectors in descending eigenvalue order as the columns of $Q_l$.

\revision{After applying this rotation, we impose input sparsity on the Q/K/V and Gate projections using the weight-aware Top-K defined in Section~\ref{sec:motivation-rotation}.} \sgs{Because adjacent layers use different rotations, the residual representation is transformed between layers using the transition matrix \revision{$T_l=Q_l^\top Q_{l+1}\in\mathbb{R}^{d\times d}$}.} Section~\ref{sec:policy-search} describes the search for the layer-wise mixture coefficients $\theta_{l,i}$.

\revision{Table~\ref{tab:rotation-statistics-diagnostics} supports this design on Llama-3.1-8B. Compared with PCA, AW-Rot has slightly lower activation-variance concentration but higher weight-energy concentration, with the fraction of output-contribution variance in the leading 50\% of dimensions increasing from 90.77\% to 94.85\%. Under actual B128 sparsification, the removed contribution proxy decreases from 1.62\% to 0.89\%, and output NMSE from 1.33\% to 0.70\%.}

\vspace{-0.7\baselineskip}

\paragraph{Rotation-Enabled LM-Head Sparsification.} The rotation framework enables additional LM-head sparsification. By concentrating final hidden activations in leading dimensions, it improves early partial-logit estimates, enabling staged evaluation and early termination of unlikely candidates to skip LM-head weight reads and computation. The rotation is absorbed into the LM-head weights offline; details are provided in Appendix~\ref{app:sparse-lm-head}.

\subsection{Correction Fusion and Efficient GPU Execution}
\label{sec:correction-fusion}

\begin{wrapfigure}[13]{R}{0.40\textwidth}
    \vspace{-\intextsep}
    \centering
    \captionsetup{font=small,skip=3pt}
    \includegraphics[width=\linewidth]{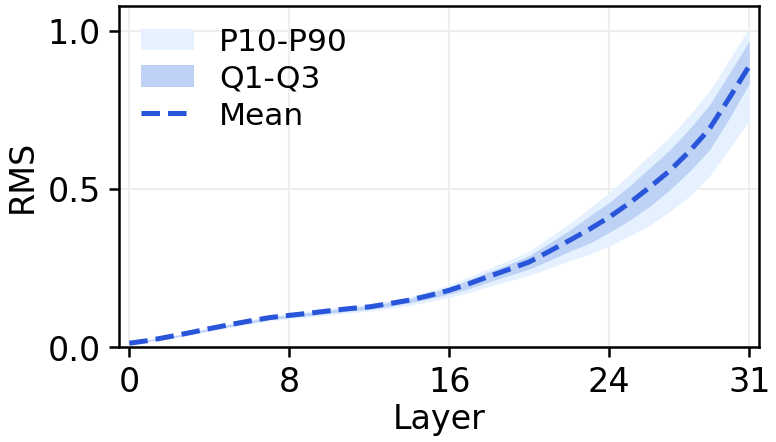}
    \caption{Layer-wise RMS distributions of FFN inputs before RMSNorm in Llama-3.1-8B, measured over $32\times4096$ tokens.} \label{fig:ffn-input-rms}
\end{wrapfigure}

\paragraph{Correction Fusion.}
Lazy shift and layer-wise rotation each require an additional operation with a $d\times d$ matrix for shift correction and inter-layer basis conversion, respectively. 
\revision{Let $X_l$ denote the FFN input residual before RMSNorm, with rotated representation $\bar X_l=X_lQ_l$, and define the inter-layer transition $T_l=Q_l^\top Q_{l+1}$. If $C_l^{\mathrm{orig}}$ is the correction formed from the original weights by Eq.~(\ref{eq:lazy-correction-matrix}) at layer $l$, its rotated form is $C_l=Q_l^\top D_{\gamma_l}C_l^{\mathrm{orig}}Q_{l+1}$, where $D_{\gamma_l}$ is the diagonal matrix of FFN RMSNorm channel scales. Thus, $C_l$ maps from the $Q_l$ basis to the $Q_{l+1}$ basis.} The two operations can be combined as

\vspace{-8pt}
\begin{equation}
    \color{black}
    \bar X_l
    \left[
        T_l + \frac{C_l}{\operatorname{RMS}(X_l)}
    \right].
    \label{eq:correction-fusion}
\end{equation}
\vspace{-5pt}

Although \ymyi{$T_l$} and $C_l$ are fixed matrices, \revision{the RMSNorm denominator} $\operatorname{RMS}(X_l)$ varies across tokens, preventing a single fused matrix from being precomputed and stored. However, as Figure~\ref{fig:ffn-input-rms} shows, activation RMS varies relatively little across tokens within a layer. RoShift exploits this property by approximating calibration RMS values with a small set of representatives $a_{l,r}$ and precomputing $T_l+C_l/a_{l,r}$ for each representative. At inference time, it uses a fused matrix if the relative difference between the actual RMS and its representative is within 5\%; otherwise, it falls back to computing the two operations separately using the actual RMS. Section~\ref{sec:memory-speedup-tradeoff} analyzes the memory-speedup trade-off as the number of representatives varies.

\vspace{-2.5pt}
\paragraph{Efficient GPU Execution.} To translate correction fusion into decoding acceleration, 
RoShift executes the sparse FFN and correction path in parallel on separate CUDA streams, synchronizing when their outputs are added. It also arranges weights offline according to the sparsification direction to enable coalesced memory access, and its sparse kernels compute only the retained components. Blockwise Top-K reduces selection overhead. Appendix~\ref{app:kernel-details} describes the kernel implementation and its impact on quality\revision{, while Appendix~\ref{app:rms-fallback-implementation} details RMS tolerance checking and fallback.}

\subsection{Sparsity and Rotation Optimization}
\label{sec:policy-search}
\noindent\textbf{Sparsity Optimization:}
Table~\ref{tab:sparsity-policy} compares \ymyi{the sparsity-allocation granularity} of existing methods. TEAL \ymyi{assigns different sparsities to individual projections within each layer} while keeping the total sparsity \ymyi{identical across layers}~\citep{liu2025teal}. WAS extends \ymyi{non-uniform allocation} across layers, but constrains sparsity to be monotonically non-decreasing with depth~\citep{wang2025was}.

\begin{wraptable}[9]{r}{0.44\textwidth}
    \vspace{-10pt}
    \centering
    \captionsetup{font=tablefont,skip=4pt}
    \caption{Sparsity allocation flexibility.}
    \label{tab:sparsity-policy}
    \fontsize{7.5}{9}\selectfont
    \setlength{\tabcolsep}{1.5pt}
    \renewcommand{\arraystretch}{1.12}
    \begin{tabular}{@{}lccc>{\columncolor[rgb]{0.93,0.96,0.98}[\tabcolsep][0pt]}c@{}}
        \toprule
        \textbf{Scope} & LaRoSA & TEAL & WAS & \textbf{RoShift} \\
        \midrule
        Projection-wise sparsity
            & $\triangle$ & $\checkmark$ & $\checkmark$ & \textbf{$\checkmark$} \\
        Layer-wise sparsity
            & $\times$ & $\times$ & $\triangle$ & \textbf{$\checkmark$} \\
        \bottomrule
    \end{tabular}\par\smallskip
    \begin{minipage}{\linewidth}
        $\triangle$: LaRoSA shares projection-wise allocation across layers; WAS supports layer-wise sparsity only under a monotonic non-decreasing constraint.
    \end{minipage}
    \vspace{-13pt}
\end{wraptable}

Because early and late LLM layers are particularly sensitive to pruning and quantization~\citep{ma2023llmpruner,wang2026sliderquant}, the monotonic constraint in WAS prevents assigning lower sparsity to sensitive late layers.  
RoShift initializes its allocation \ymyi{using} TEAL-style within-layer allocation \ymyi{and} WAS-style inter-layer search, \ymyi{then refines} it \ymyi{through} pairwise coordinate descent \ymyi{without such a constraint}.
Specifically, each update transfers sparsity between two layer-projection coordinates while preserving overall MAC sparsity. MAC sparsity denotes the fraction of linear-projection MACs eliminated by sparsification, following the same model-level sparsity definition as TEAL and WINA. \ymyi{We first} screen candidates using \ymyi{inexpensive} local error and PPL on a short calibration set, then evaluate \ymyi{the remaining candidates} using PPL on the full calibration set. We repeatedly accept the update \ymyi{that yields} the largest PPL improvement. 

\noindent\textbf{Rotation Optimization:}
After fixing the sparsity allocation, we search for the layer-wise coefficients $\theta_{l,i}$ that combine the six matrices in Section~\ref{sec:output-aware-rotation}. 
Starting from the same initial mixture in every layer, we perform a greedy pairwise search by transferring a fixed amount between coefficients. Each resulting candidate rotation $Q_l$ is evaluated using the relative L2 error of the block output after \ymyi{the} attention and the FFN with respect to the dense output. \ymyi{We accept the best transfer if it reduces the error.}
\ymyi{However, locally accepted rotations may not remain beneficial when combined across layers.}
Rather than immediately adopting \ymyi{the layer-wise coefficient \revision{candidates} obtained independently}, we test them sequentially from shallow to deep layers and retain each $Q_l$ only if it improves the full-model PPL on the full calibration set.
Appendix~\ref{app:search-details} details both search procedures.

\section{Experiments}
\label{sec:experiments}

\begin{table}[!t]
\centering
\captionsetup{font=tablefont,skip=5pt}
\caption{Accuracy under non-uniform sparsity at matched Transformer MAC-reduction targets, excluding the LM-head. Each model entry reports mean accuracy over 11 benchmarks.
Full results are in Appendix~\ref{app:complete-results}.}
\label{tab:model-accuracy-summary}
\begingroup
\fontsize{8}{9}\selectfont
\setlength{\tabcolsep}{0.6pt}
\renewcommand{\arraystretch}{1.15}
\begin{tabular*}{\linewidth}{@{\extracolsep{\fill}}l*{6}{c}@{}}
\toprule
Method / Model & Llama-3.1-8B & Llama-2-7B & Mistral-7B & Qwen2.5-7B & Qwen3-8B & Avg \\
\midrule
Dense (0\%) & 63.87 & 47.14 & 59.67 & 73.28 & 76.43 & 64.08 \\
\midrule
TEAL 40\% & \underline{61.83}(-2.05) & 45.89(-1.26) & 58.56(-1.11) & \textbf{71.97}\textbf{(-1.32)} & \underline{74.54}(-1.89) & \underline{62.56}(-1.52) \\
LaRoSA 40\% & 61.21(-2.66) & 44.99(-2.15) & 57.83(-1.84) & 71.04(-2.25) & 74.02(-2.40) & 61.82(-2.26) \\
WINA 40\% & 61.71(-2.16) & \textbf{46.39}\textbf{(-0.75)} & \underline{58.60}(-1.07) & 71.26(-2.02) & 73.97(-2.45) & 62.39(-1.69) \\
\textbf{RoShift 40\%} & \textbf{62.26}\textbf{(-1.62)} & \underline{46.03}(-1.12) & \textbf{59.24}\textbf{(-0.43)} & \underline{71.86}(-1.42) & \textbf{75.16}\textbf{(-1.27)} & \textbf{62.91}\textbf{(-1.17)} \\
\midrule
TEAL 50\% & \underline{59.21}(-4.67) & 44.34(-2.81) & \underline{56.95}(-2.72) & \underline{70.46}(-2.82) & \underline{73.49}(-2.94) & \underline{60.89}(-3.19) \\
LaRoSA 50\% & 56.77(-7.10) & 43.09(-4.05) & 54.77(-4.90) & 68.13(-5.16) & 70.90(-5.52) & 58.73(-5.35) \\
WINA 50\% & 58.52(-5.35) & \underline{45.49}(-1.65) & 56.44(-3.23) & 68.95(-4.33) & 71.94(-4.49) & 60.27(-3.81) \\
\textbf{RoShift 50\%} & \textbf{61.47}\textbf{(-2.40)} & \textbf{45.90}\textbf{(-1.25)} & \textbf{58.54}\textbf{(-1.13)} & \textbf{71.60}\textbf{(-1.68)} & \textbf{74.93}\textbf{(-1.50)} & \textbf{62.49}\textbf{(-1.59)} \\
\midrule
TEAL 60\% & \underline{53.44}(-10.43) & 41.46(-5.69) & \underline{54.42}(-5.25) & \underline{67.19}(-6.09) & \underline{70.41}(-6.01) & \underline{57.38}(-6.70) \\
LaRoSA 60\% & 47.49(-16.38) & 39.77(-7.37) & 50.58(-9.09) & 61.39(-11.90) & 60.35(-16.08) & 51.92(-12.16) \\
WINA 60\% & 50.62(-13.26) & \underline{42.17}(-4.97) & 52.41(-7.26) & 63.42(-9.86) & 66.47(-9.96) & 55.02(-9.06) \\
\textbf{RoShift 60\%} & \textbf{58.54}\textbf{(-5.33)} & \textbf{43.41}\textbf{(-3.73)} & \textbf{57.82}\textbf{(-1.85)} & \textbf{70.53}\textbf{(-2.75)} & \textbf{73.07}\textbf{(-3.36)} & \textbf{60.68}\textbf{(-3.40)} \\
\bottomrule\end{tabular*}
\endgroup
\end{table}

\subsection{Experimental Setup}
\label{sec:experimental-setup}
We evaluate RoShift's accuracy preservation and practical decoding acceleration. Our primary speed baseline is TEAL, which provides a publicly available end-to-end decoding acceleration implementation. \revision{WAS and WINA do not provide public end-to-end decoding implementations. Although LaRoSA reports end-to-end decoding speedups, its public release contains only PyTorch accuracy-evaluation code, without sparse GEMV kernels or a decoding-acceleration implementation. We therefore exclude LaRoSA from the main speedup comparison. Appendix~\ref{app:larosa-correction-ablation} provides a supplementary comparison with LaRoSA implemented using RoShift's GPU kernel backend.} We additionally include LaRoSA and WINA as accuracy baselines, representing rotation-based and weight-aware activation sparsification, respectively~\citep{liu2025larosa,chen2026wina}.

We evaluate five \ymyj{LLM models} on 11 benchmarks: 
\ymyi{MMLU, ARC-Easy, ARC-Challenge, BoolQ, PIQA, WinoGrande, CommonsenseQA, BBH, GSM8K, MATH-500, and MBPP}. \ymyi{End-to-end latency and decoding throughput, including correction operations, are measured on an RTX 5090 desktop GPU and a Jetson AGX Orin edge device.} Appendix~\ref{app:experimental-settings} provides the detailed experimental settings.

\vspace{-1pt}
\subsection{Accuracy}
\label{sec:accuracy-results}

The target MAC sparsity in Table~\ref{tab:model-accuracy-summary} covers Transformer blocks only, excluding LM-head sparsity. 
Among the methods in Table~\ref{tab:model-accuracy-summary}, RoShift achieves the highest mean accuracy in 13 of the 15 configurations spanning five models and three sparsity levels. At both 50\% and 60\% sparsity, it outperforms TEAL, LaRoSA, and WINA on all five models. Averaged across models, the gains over TEAL are 1.60 and 3.29 percentage points at 50\% and 60\% sparsity, respectively; 
\ymyj{the corresponding gains are 3.76 and 8.76 percentage points over LaRoSA, and 2.22 and 5.66 percentage points over WINA.}
Note that the measured MAC sparsity of TEAL's fixed-threshold implementation is, on average across models, 0.83–1.68 percentage points below its target, which may partly explain its relatively strong accuracy compared with WINA. See Appendix~\ref{app:baseline-implementations} for details. These results show that RoShift's accuracy preservation advantage becomes more pronounced as sparsity increases.

\sgs{Subsequent speedup and accuracy-speedup Pareto analyses use uniform sparsity for controlled runtime comparisons. The corresponding accuracy results are reported in Appendix~\ref{app:uniform-results}, while Appendix~\ref{app:roshift-allocation-comparison} compares uniform and non-uniform sparsity for RoShift.} \revision{Comparisons using shared calibration data and correction-inclusive MAC budgets are reported in Appendices~\ref{app:calibration-data} and~\ref{app:correction-inclusive-mac-budget}, respectively.}

\vspace{-1pt}
\subsection{Speedup}
\label{sec:end-to-end-acceleration}
Table~\ref{tab:token-generation-speed} reports RoShift's end-to-end decoding throughput with FP16, batch size~1, and uniform sparsity. RoShift accelerates decoding over Dense for every evaluated model and sparsity setting on both the RTX~5090 and Jetson AGX Orin. At 50\% sparsity, it achieves speedups of up to 1.72$\times$ and 1.69$\times$ on the RTX~5090 and Orin, respectively; at 75\% sparsity, these increase to 2.60$\times$ and 2.73$\times$. \revision{Appendix~\ref{app:free-running-generation} reports free-running speedup and fallback rates.}

RoShift additionally stores a transition matrix \(Q_l^\top Q_{l+1}\), a lazy-shift correction matrix, and four fused matrices per layer. For Llama-3.1-8B at 50\% sparsity and a 4K context length, these additions increase peak VRAM usage by 29.31\% over standard dense execution. \revision{Appendix~\ref{app:peak-gpu-memory} reports peak memory usage for all five models.} Correction and transition add an average computational overhead of 14.04\%, while LM-head sparsification saves 10.49\%, resulting in a net overhead of 3.55\%. The end-to-end decoding speedups in Table~\ref{tab:token-generation-speed} include all these overheads.

\begin{table*}[!t]
    \centering
    \fontsize{8}{9}\selectfont
    \captionsetup{font=tablefont,skip=5pt}
    \caption{Batch-size-1 decoding throughput (tokens/s; speedup over the dense baseline in parentheses) of RoShift with uniform sparsity across Transformer layers and projections. Measurement details are in Appendix~\ref{app:kernel-details}.}
    \label{tab:token-generation-speed}
    \begingroup
    \newcommand{\throughput}[2]{\mbox{#1(#2)}}
    \setlength{\tabcolsep}{0.6pt}
    \renewcommand{\arraystretch}{1.15}
    \begin{tabular*}{\linewidth}{@{\extracolsep{\fill}}lc*{5}{c}@{}}
        \toprule
        GPU & Sparsity & \shortstack{Llama-3.1\\8B}
          & \shortstack{Llama-2\\7B} & \shortstack{Mistral\\7B}
          & \shortstack{Qwen2.5\\7B} & \shortstack{Qwen3\\8B} \\
        \midrule
        \multirow{4}{*}{\shortstack{RTX\\5090}}
          & Dense & \throughput{100.58}{1.00$\times$} & \throughput{104.62}{1.00$\times$} & \throughput{94.66}{1.00$\times$}
                  & \throughput{96.72}{1.00$\times$} & \throughput{97.43}{1.00$\times$} \\
        \cmidrule(lr){2-7}
          & 25\% & \throughput{120.03}{1.19$\times$} & \throughput{121.64}{1.16$\times$} & \throughput{121.63}{1.28$\times$}
                  & \throughput{129.81}{\textbf{1.34$\times$}} & \throughput{113.52}{1.17$\times$} \\
          & 50\% & \throughput{159.86}{1.59$\times$} & \throughput{159.20}{1.52$\times$} & \throughput{162.66}{\textbf{1.72$\times$}}
                  & \throughput{159.37}{1.65$\times$} & \throughput{146.86}{1.51$\times$} \\
          & 75\% & \throughput{238.18}{2.37$\times$} & \throughput{235.79}{2.25$\times$} & \throughput{246.03}{\textbf{2.60$\times$}}
                  & \throughput{239.96}{2.48$\times$} & \throughput{217.42}{2.23$\times$} \\
        \midrule
        \multirow{4}{*}{\shortstack{Jetson\\AGX Orin\\64GB}}
          & Dense & \throughput{11.49}{1.00$\times$} & \throughput{12.58}{1.00$\times$} & \throughput{12.12}{1.00$\times$}
                  & \throughput{12.08}{1.00$\times$} & \throughput{11.34}{1.00$\times$} \\
        \cmidrule(lr){2-7}
          & 25\% & \throughput{14.09}{\textbf{1.23$\times$}} & \throughput{14.78}{1.17$\times$} & \throughput{14.25}{1.18$\times$}
                  & \throughput{14.62}{1.21$\times$} & \throughput{13.45}{1.19$\times$} \\
          & 50\% & \throughput{19.38}{\textbf{1.69$\times$}} & \throughput{20.16}{1.60$\times$} & \throughput{19.69}{1.63$\times$}
                  & \throughput{20.42}{\textbf{1.69$\times$}} & \throughput{18.18}{1.60$\times$} \\
          & 75\% & \throughput{30.86}{2.69$\times$} & \throughput{31.65}{2.52$\times$} & \throughput{31.87}{2.63$\times$}
                  & \throughput{32.98}{\textbf{2.73$\times$}} & \throughput{28.34}{2.50$\times$} \\
        \bottomrule
    \end{tabular*}
    \endgroup
\end{table*}

\begin{figure}[!t]
    \centering
    \includegraphics[width=\textwidth]{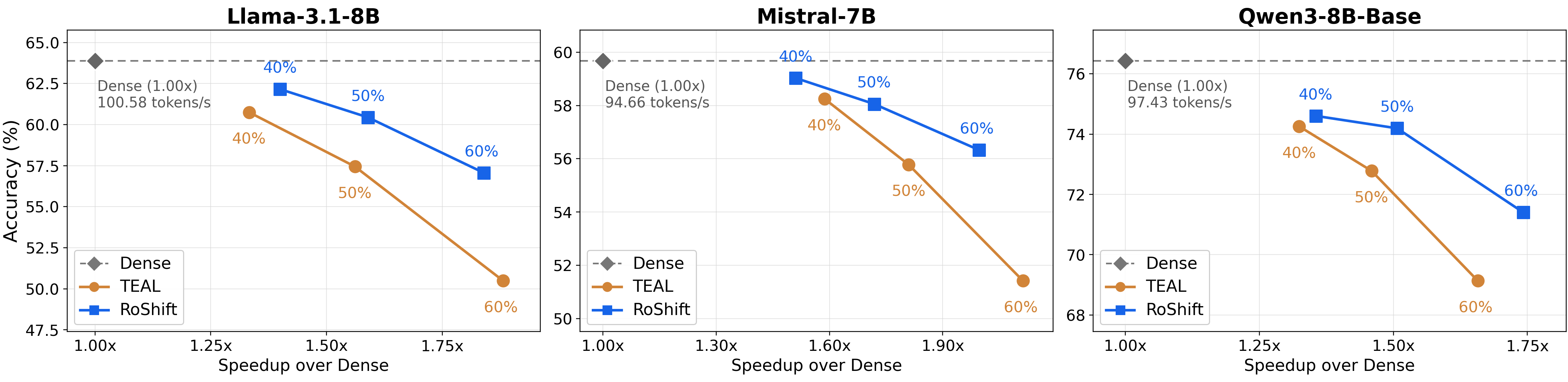}
    \captionsetup{font=small,skip=3pt}
    \caption{Accuracy-speedup trade-offs for three models on the RTX 5090. Accuracy is averaged over 11 benchmarks at uniform sparsity; speedup is the end-to-end decoding speedup over each model's Dense baseline.}
    \label{fig:quality-latency}
\end{figure}
\vspace{-1pt}

\subsection{Accuracy--Speedup Trade-off}
\label{sec:accuracy-speedup-tradeoff}
Figure~\ref{fig:quality-latency} compares the end-to-end accuracy-speedup trade-offs of RoShift and TEAL using measured end-to-end decoding speedups. Because speed is measured under uniform sparsity, we evaluate accuracy with the same uniform sparsity settings to construct each method's Pareto frontier. RoShift additionally sparsifies the LM-head, and its effect is included in both decoding speedup and generation benchmark evaluations. \ymyj{LM-head sparsification is a RoShift-specific component enabled by its rotation.}

RoShift not only preserves higher accuracy at the same sparsity, but also offers a better accuracy-speedup trade-off than TEAL when compared using measured end-to-end decoding speedups. Specifically, it achieves higher speedup at comparable accuracy or higher accuracy at comparable speedup, yielding a better Pareto frontier. 
This shows that lazy shift and \ymyj{weight-aware} rotation reduce sparsification-induced accuracy loss, while correction fusion and \ymyj{efficient} GPU \ymyj{execution} translate these benefits into greater end-to-end speedup for a given accuracy loss. \revision{Appendix~\ref{app:head-controlled-pareto} compares TEAL and RoShift under matched LM-head settings.}

\Needspace{22\baselineskip}
\subsection{Memory--Speedup Trade-off}
\label{sec:memory-speedup-tradeoff}
\begin{wrapfigure}[14]{r}{0.365\textwidth}
    \vspace{-1\intextsep}
    \vspace{-10pt}
    \centering
    \captionsetup{font=small,skip=3pt}
    \includegraphics[width=\linewidth]{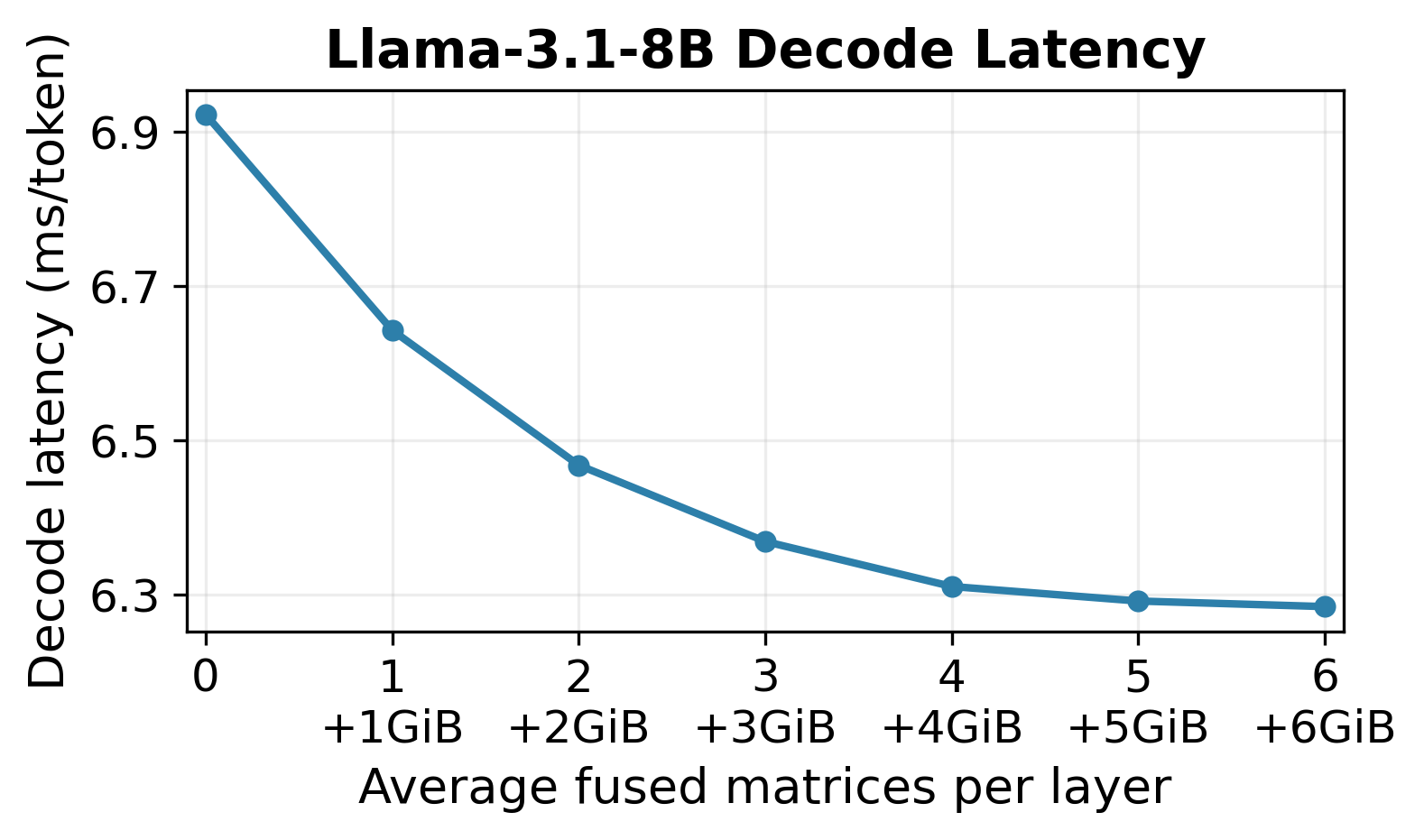}
    \caption{Decoding latency versus \ymyj{fused-matrix cache size} for Llama-3.1-8B at 50\% sparsity. The x-axis shows the average number of fused matrices per layer. The GiB annotations exclude the fixed 2 GiB for transition and correction matrices.}
\label{fig:fused-cache-tradeoff}
    \vspace{-10pt}
\end{wrapfigure}

RoShift uses a 5\% tolerance for the relative difference between the actual RMS($X$) and its representative. Figure~\ref{fig:fused-cache-tradeoff} shows end-to-end decoding latency versus fused-matrix storage under this tolerance. Storing more fused matrices increases RMS($X$) matches, reducing exact-correction fallbacks and latency. Beyond a certain point, however, most inputs already use fused matrices, so additional memory yields diminishing returns. Considering this trade-off, RoShift uses an average of four fused matrices per layer by default. Appendix~\ref{app:cache-tolerance} provides the tolerance calibration and detailed analysis.

Fused-matrix storage can be adjusted to deployment memory constraints. With limited VRAM, a smaller cache reduces additional memory use; with \ymyj{sufficient} VRAM, storing more fused matrices reduces fallbacks and increases speedup.

\subsection{Ablation Studies}
\label{sec:ablation-studies}
\paragraph{Accuracy.} Table~\ref{tab:representation-transformations} compares mean accuracy over 11 benchmarks on Llama-3.1-8B under uniform sparsity for variants of rotation, input selection, and lazy shift. \revision{All sparse variants use B128 blockwise selection and Up output sparsity, which propagates as Down input sparsity. Top-K uses activation magnitudes; WA-Top-K multiplies these magnitudes by corresponding weight-row $\ell_2$ norms for Q/K/V and Gate input selection (Section~\ref{sec:motivation-rotation}). Plain Top-K has neither rotation nor lazy shift and differs from TEAL in both blockwise selection versus fixed thresholds and gate-based Up/Down support versus projection-input thresholding.} A-Rot. and AW-Rot. denote activation-only and activation- and weight-aware rotations. The LM-head remains dense in all configurations to isolate Transformer-layer sparsification; the effect of LM-head sparsification is evaluated separately in Appendix~\ref{app:sparse-lm-head}.

\Needspace{17\baselineskip}
\begin{wraptable}[17]{r}{0.42\textwidth}
\begingroup
\setlength{\intextsep}{0pt}
\vspace{-1.2\baselineskip}
    \centering
    \captionsetup{font=tablefont,skip=5pt}
    \caption{\ymyi{Accuracy} ablation on Llama-3.1-8B at uniform sparsity. Top-K uses activation magnitudes; WA-Top-K also uses weight norms. A-Rot. and AW-Rot. denote activation-only and activation- and weight-aware rotations.}
    \label{tab:representation-transformations}
    \fontsize{8}{9}\selectfont
    \setlength{\tabcolsep}{1.4pt}
    \renewcommand{\arraystretch}{1.1}
    \begin{tabular*}{\linewidth}{@{\extracolsep{\fill}}l*{3}{c}@{}}
        \toprule
        \textbf{Method} & \textbf{40\%} & \textbf{50\%} & \textbf{60\%} \\
        \midrule
        Dense & 63.87 & 63.87 & 63.87 \\
        \midrule
        Top-K & 58.81 & 52.44 & 41.28 \\
        WA-Top-K & 58.75 & 52.97 & 42.57 \\
        A-Rot. + Top-K & 58.55 & 53.21 & 39.40 \\
        A-Rot. + WA-Top-K & 58.69 & 52.63 & 37.60 \\
        AW-Rot. + WA-Top-K & 59.47 & 54.34 & 44.92 \\
        Lazy shift + Top-K & 61.87 & 58.59 & 51.79 \\
        Lazy shift + WA-Top-K & 61.70 & 58.95 & 53.28 \\
        \textbf{RoShift} & \textbf{62.09} & \textbf{60.41} & \textbf{57.08} \\
        \bottomrule
    \end{tabular*}
\endgroup
\end{wraptable}

Without rotation, WA-Top-K achieves higher accuracy than Top-K at 50\% and 60\% sparsity, with a larger gain at 60\%. 
In contrast, A-Rot.\ + Top-K does not consistently \ymyj{outperform} Top-K without rotation. Replacing Top-K with WA-Top-K under A-Rot.\ does not \ymyj{produce a consistent gain and lowers accuracy at 50\% and 60\% sparsity.} 
However, when all configurations use WA-Top-K, AW-Rot.\ consistently achieves higher accuracy than both A-Rot.\ and no rotation, with the gains over A-Rot.\ becoming more pronounced at higher sparsity. These results highlight a limitation of simply combining activation-only rotation with weight-aware selection; RoShift addresses this limitation by incorporating weight information into the rotation itself.

Lazy shift provides substantial gains even without rotation. 
\revision{The ``Lazy shift + WA-Top-K'' configuration removes AW-Rot.\ from full RoShift and therefore applies no rotation.}
Notably, lazy shift also achieves substantial gains with standard Top-K, indicating that its effectiveness does not depend on weight-aware selection. 
Adding lazy shift to WA-Top-K yields higher accuracy than adding AW-Rot., \ymyj{indicating that} lazy shift \ymyj{is} a major contributor to RoShift's accuracy gains. 
Moreover, RoShift, which \ymyi{carefully integrates} AW-Rot.\ and lazy shift, achieves \ymyj{the highest} accuracy at every sparsity level. These results show that lazy shift provides substantial standalone gains, with activation- and weight-aware rotation offering complementary benefits that further improve accuracy.

\Needspace{15\baselineskip}
\begin{wrapfigure}{r}{0.45\textwidth}
    \vspace{-\intextsep}
    \centering
    \captionsetup{font=small,skip=3pt}
    \includegraphics[width=\linewidth]{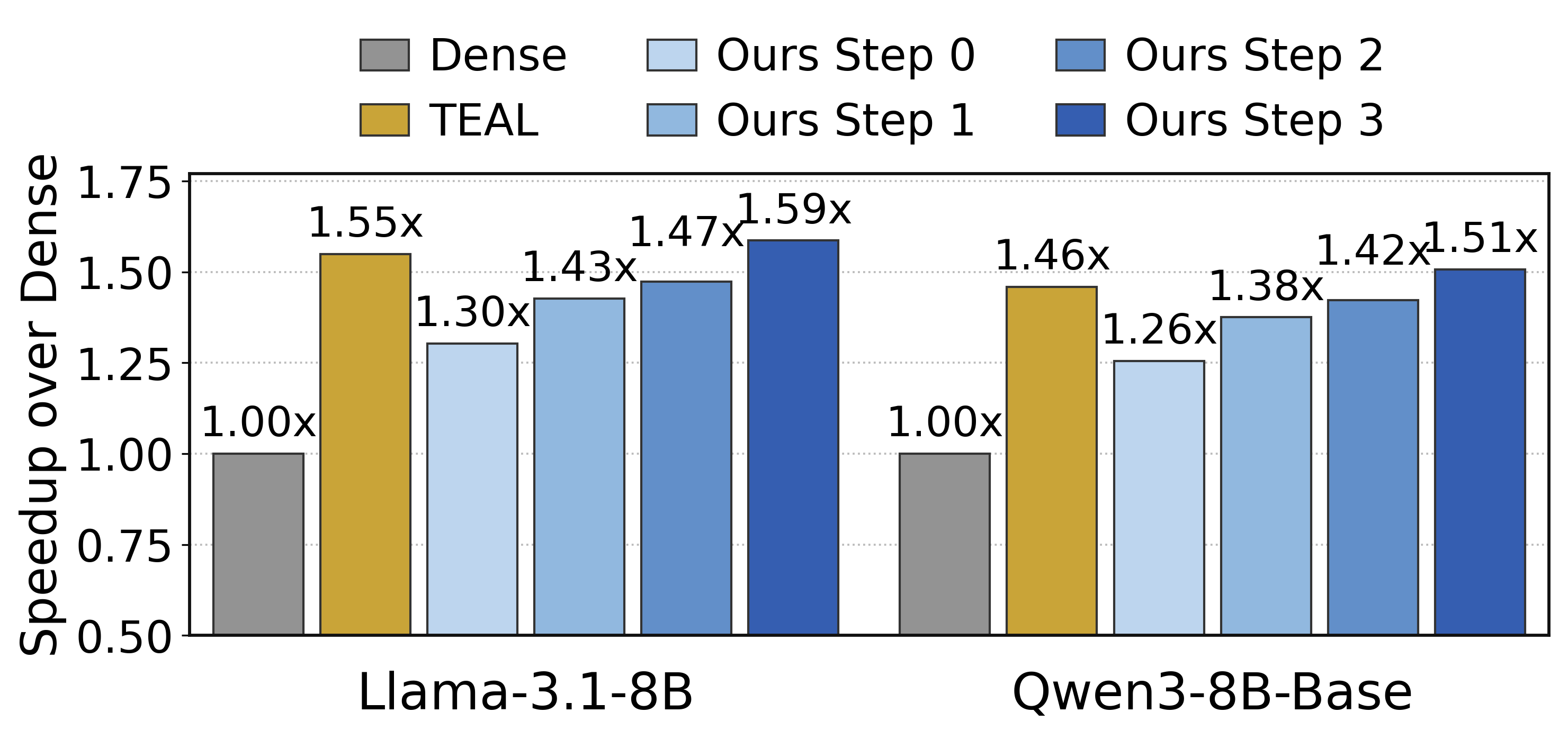}
    \caption{Cumulative decoding speedup over Dense at 50\% uniform sparsity.}
    \label{fig:speedup-ablation}
    \vspace{-10pt}
\end{wrapfigure}

\paragraph{Speedup.} Figure~\ref{fig:speedup-ablation} compares decoding speedup over Dense as RoShift's optimizations are added sequentially at 50\% sparsity. Step 0 uses sparse kernels but computes rotation transition and shift correction separately, keeping the LM-head dense. Step 1 introduces fused matrices associated with RMS representatives, and Step 2 executes the sparse FFN and correction on separate CUDA streams. Finally, Step 3 adds LM-head sparsification.

In this cumulative ordering, correction fusion provides a substantial additional speedup. When a fused matrix is selected, it replaces the two $d\times d$ matrix-vector products for rotation transition and shift correction with one, eliminating one matrix's weight reads and computations. LM-head sparsification also provides substantial acceleration because the LM-head uses a large weight matrix to project from the model dimension to the entire vocabulary. The incremental speedups differ across models, reflecting differences in factors such as the LM-head's share of computation and the frequency of fused-matrix use. \revision{Appendix~\ref{app:larosa-correction-ablation} provides a LaRoSA comparison and additional correction ablations.}

\subsection{Compatibility with Quantization}
\label{sec:quantization-compatibility}
\begin{wraptable}[10]{r}{0.39\textwidth}
\vspace{-0.7\baselineskip}
\begingroup
    \centering
    \vspace{-5pt}
    \newcommand{\pplchange}[2]{\shortstack{#1\\(#2)}}
    \captionsetup{font=tablefont,skip=4pt}
    \captionof{table}{
    PPL of RoShift under OmniQuant. $W$ and $T$ denote \ymyi{the bit widths of the model} weight and transition matrix $Q_l^\top Q_{l+1}$, respectively. \ymyi{Baseline uses} \sgs{FP16.}
    }
    \label{tab:quantization-compatibility}
    \setlength{\intextsep}{0pt}
    \fontsize{8}{9}\selectfont
    \setlength{\tabcolsep}{1pt}
    \renewcommand{\arraystretch}{1.04}
    \noindent\makebox[\linewidth][r]{\begin{tabular}{@{}lccccc@{}}
        \toprule
        Method & Baseline & W8T8 & W4T16 & W4T8 & W4T4 \\
        \midrule
        0\% & 4.99 & 5.07 & 5.38 & 5.52 & 392.60 \\
        \addlinespace[2pt]
        50\% & 5.82 & 5.73 & 6.04 & 6.14 & 282.01 \\
        \bottomrule
    \end{tabular}}
    \endgroup

\end{wraptable}
To evaluate compatibility with weight quantization, we apply 8-bit and 4-bit quantization to RoShift using RTN, GPTQ, AWQ, and OmniQuant~\citep{frantar2022gptq, lin2024awq, shao2024omniquant}. Table~\ref{tab:quantization-compatibility} reports PPL with OmniQuant.
\ymyi{
We quantize the model weights and inter-layer transition matrices $Q_l^\top Q_{l+1}$ \sgs{to either} 8 or 4 bits, or use 4-bit weights while retaining the transition matrices at 8 or 16 bits.} 
\sgs{All sparse models use 50\% uniform sparsity.}

\ymyi{With 8-bit quantization \sgs{(W8T8)}, both the dense and 50\%-sparse models retain essentially the same PPL as baseline models, indicating that RoShift is highly compatible with 8-bit quantization.}
\ymyi{In contrast, jointly quantizing both the weights and the transition matrices \(Q_l^\top Q_{l+1}\) to 4 bits causes severe degradation, }suggesting that the transition matrices are particularly sensitive to low-bit quantization.

Keeping transition matrices at 8 \ymyi{or 16} bits while quantizing the weights to 4 bits \ymyi{largely} restores the PPL. Combining RoShift with weight quantization could \ymyi{thus provide additional} inference acceleration, \sgs{but} requires kernels \ymyi{jointly} optimized for sparse and quantized computation, which we leave to future work. Appendix~\ref{app:complete-quantization-results} provides complete results, including RTN, GPTQ, and AWQ.

\section{Conclusion}
We \ymyj{proposed} RoShift, a training-free activation sparsification method for efficient LLM inference. 
Lazy shift recenters activation-function outputs and defers the required correction, \ymyj{enabling more activations to be sparsified while maintaining comparable accuracy}. Activation- and weight-aware rotation further reduces the loss of important information during sparsification. 
\ymyj{Finally, correction fusion} and \ymyj{optimized GPU execution mitigate the overhead introduced } by these transformations, translating \ymyj{the resulting activation sparsity} into practical end-to-end decoding acceleration.

Our experiments show that RoShift preserves higher accuracy than the strong baselines TEAL, LaRoSA and WINA at high sparsity and achieves decoding speedups over dense models on real hardware. 
Its accuracy-speedup Pareto frontier \ymyj{provides} a better trade-off than \ymyj{that of} TEAL, \ymyj{showing that RoShift's benefits extend beyond accuracy preservation to} practical execution efficiency. These results \ymyj{highlight RoShift's potential for} latency- and efficiency-sensitive environments, including edge and on-device settings and Physical AI platforms.

\subsection*{AI use statement}
Generative AI tools assisted with research-methodology discussion, experiment
design, software implementation, result interpretation, literature discovery,
and manuscript preparation, including translation. The research ideas and
mathematical derivations were primarily developed by the authors, with AI tools
providing assistance in reviewing the mathematical proofs. The authors review
and validate all AI-assisted
code, experiments, mathematical statements, citations, and manuscript text and
take responsibility for the final submission.

\subsection*{Reproducibility statement}
We describe the experimental setup and evaluation protocols in
Section~\ref{sec:experimental-setup} and Appendix~\ref{app:experimental-settings},
the kernel design and decoding measurement procedure in
Appendix~\ref{app:kernel-details}, and the sparsity allocation and rotation search
algorithms in Appendix~\ref{app:search-details}. Code and experimental
configurations, including those for reproducing the reported accuracy and
decoding speedup results, are provided in an anonymized ZIP archive as
supplementary material. We will release an official public code repository upon
acceptance.

\bibliography{references}
\bibliographystyle{iclr2027_conference}

\clearpage
\appendix
\section{Appendix}
\label{app:appendix}

\subsection{Related Work}
\label{app:related-work}

\noindent\textbf{LLM Activation Sparsity:}
DejaVu \ymyi{observes} contextual sparsity in LLMs, where a subset of neurons and attention heads can approximate the output for a given input, and predicts this subset using lightweight predictors~\citep{liu2023dejavu}. Building on this prediction-based approach to \ymyi{exploiting output sparsity}, SparseInfer, Grasp, and oFFN use low-bit activation and weight information to reduce prediction cost while improving accuracy~\citep{shin2025sparseinfer,shin2025grasp,song2026offn}. Other work optimizes execution based on neuron activation patterns. PowerInfer and oFFN distinguish hot and cold neurons by activation frequency and assign suitable hardware resources and execution paths to each~\citep{song2024powerinfer,song2026offn}.

\noindent\textbf{Sparsifying Non-ReLU LLMs:}
Modern LLMs using SiLU exhibit little natural activation sparsity. ReLU Strikes Back and ProSparse replace the activation function with ReLU and use additional training to obtain sparsity while preserving model performance~\citep{mirzadeh2024relu,song2025prosparse}. Training-free alternatives induce sparsity by approximating small-magnitude activations as zero. CATS applies this approach to FFNs, while TEAL extends it to attention projections for model-wide sparsification~\citep{lee2024cats,liu2025teal}. R-Sparse incorporates SVD-based low-rank approximation to compensate for sparsification-induced output error~\citep{zhang2025rsparse}.

\noindent\textbf{Weight-aware Sparsification:}
Beyond activation magnitude, several methods assess component importance using the corresponding weights. ReLU$^2$ Wins evaluates neuron importance through its output contribution and measures the cumulative error from removal using CETT~\citep{zhang2024relu2}. From this output-contribution perspective, WAS and WINA improve sparsification criteria by jointly considering activation magnitudes and the corresponding weight norms~\citep{wang2025was,chen2026wina}.

\noindent\textbf{Sparsity-friendly Transformations:}
An alternative to improving removal criteria is to transform activations into representations more amenable to sparsification. SliceGPT uses PCA-based rotation to \sgs{concentrate important activation information in a subset of dimensions, then compresses the model by removing less important dimensions and their corresponding weights}~\citep{ashkboos2024slicegpt}. LaRoSA applies rotation-based representation transformations to activation sparsification, combining layer-wise PCA rotation with Top-K selection of rotated activations~\citep{liu2025larosa}. SCAP applies mode-centering to activation-function outputs in non-GLU FFNs, shifting the distribution toward zero to increase the number of small-magnitude activations and facilitate sparsification~\citep{chua2024scap}.

Unlike prior transformation-based approaches, RoShift introduces lazy shift to exploit recentering-induced sparsity throughout the gated FFN, constructs activation- and weight-aware rotations tailored to sparsification error, and carefully integrates the two transformations through correction fusion to realize practical end-to-end decoding acceleration.

\subsection{Lazy Shift Correction Overhead}
\label{app:method-details}

Lazy shift adds a correction term $XC$ to each FFN. Let $d$ be the model dimension, $k$ the FFN intermediate dimension, and $L$ the number of layers. Each correction matrix $C$ contains $d^2$ entries per layer, whereas the Gate, Up, and Down projections together contain $3dk$ parameters. The additional number of stored matrix entries $\Delta P$ across all correction matrices and its ratio to the FFN parameter count are therefore
\[
    \Delta P = Ld^2, \qquad
    \frac{\Delta P}{P_{\mathrm{FFN}}}
    = \frac{Ld^2}{3Ldk} = \frac{d}{3k}.
\]
Because both correction and FFN parameter counts scale linearly with the number of layers, the per-layer ratio also applies to the full FFN stack. The model-wide ratio is \ymyi{lower}, however, because the full model also contains attention, embedding, LM-head, and normalization parameters. Table~\ref{tab:lazy-shift-overhead} compares these two ratios using each model's actual parameter count. Here, additional parameters refer to entries in correction matrices precomputed from existing weights, not newly learned parameters. The table reports the storage overhead of $C$ only; the rotation transition matrix introduces the same additional storage.

\begin{center}
    \begin{minipage}{\linewidth}
        \centering
        \fontsize{9}{10.8}\selectfont
        \setlength{\tabcolsep}{4pt}
        \captionof{table}{Storage overhead of lazy-shift correction matrices. Added parameters sum $C$ across all layers. Model-relative and FFN-relative increases use the full Dense model parameter count and the combined Gate, Up, and Down parameter count as denominators, respectively. Ratios are computed before rounding.}
\label{tab:lazy-shift-overhead}
        \begin{tabular}{lrrrr}
            \toprule
            Model & \shortstack{Dense params\\($A$)}
              & \shortstack{Added $C$ params\\($B$)}
              & \shortstack{Model increase\\($B/A$)}
              & \shortstack{FFN increase\\($d/(3k)$)} \\
            \midrule
            Llama-3.1-8B     & 8.03B & 0.54B & 6.69\% & 9.52\% \\
            Llama-2-7B       & 6.74B & 0.54B & 7.97\% & 12.40\% \\
            Mistral-7B       & 7.24B & 0.54B & 7.41\% & 9.52\% \\
            Qwen2.5-7B       & 7.62B & 0.36B & 4.72\% & 6.31\% \\
            Qwen3-8B        & 8.19B & 0.60B & 7.37\% & 11.11\% \\
            \bottomrule
        \end{tabular}
    \end{minipage}
\end{center}

\subsection{Rotation Derivations}
\label{app:rotation-derivation}

This section derives RoShift's rotation construction. We begin with activation-only PCA rotation, extend it to incorporate the corresponding weights for a single linear projection, and then generalize to a layer-wise rotation shared by multiple projections in a Transformer block.

\subsubsection{Activation-only Rotation}

Let $X\in\mathbb{R}^{N\times d}$ denote activation samples collected during calibration and $\mu\in\mathbb{R}^{d}$ their mean across samples. The centered activations \ymyi{$X_c$ and empirical covariance matrix $S$ are defined as follows.}
\begin{equation}
    X_c = X-\mathbf{1}\mu^\top
    \label{eq:centering}
\end{equation}
\begin{equation}
    S
    =
    \frac{1}{N}X_c^\top X_c
    \in\mathbb{R}^{d\times d}
    \label{eq:app-activation-cov}
\end{equation}

For a unit direction $q\in\mathbb{R}^{d}$ with $q^\top q=1$, the variance of the projected activations \ymyi{is}
\begin{equation}
    \operatorname{Var}(Xq)
    =
    \frac{1}{N}\|X_cq\|_2^2
    =
    q^\top S q
    \label{eq:app-pca-variance}
\end{equation}
The first principal component is therefore obtained by maximizing
\begin{equation}
    \max_{q^\top q=1} q^\top S q.
    \label{eq:app-pca-objective}
\end{equation}

Defining the Lagrangian as in Eq.~(\ref{eq:lagrangian}), the stationarity condition yields Eq.~(\ref{eq:app-pca-eigen}).
\begin{equation}
    \mathcal{L}(q,\lambda)
    =
    q^\top S q
    -
    \lambda(q^\top q-1)
    \label{eq:lagrangian}
\end{equation}
\begin{equation}
    Sq=\lambda q
    \label{eq:app-pca-eigen}
\end{equation}
The eigenvector associated with the largest eigenvalue therefore solves Eq.~(\ref{eq:app-pca-objective}). Requiring each subsequent axis to be orthogonal to the previously selected axes yields an orthogonal rotation formed from the eigenvectors of $S$ in descending eigenvalue order:
\begin{equation}
    Q=[q_1,\ldots,q_d],
    \qquad
    Q^\top Q=I
\end{equation}
Thus, in the activation-only case, a single covariance matrix $S$ suffices to construct the rotation.

\subsubsection{Activation- and Weight-aware Rotation for a Single Projection}

For activation sparsification, both activation variance and its influence on the projection output through the corresponding weights matter. Consider a linear projection with weights $W$ and an orthogonal rotation $Q=[q_1,\ldots,q_d]$. We have
\begin{equation}
    Y=XW,
    \qquad
    W\in\mathbb{R}^{d\times d_{\mathrm{out}}}
\end{equation}
\begin{equation}
    Y=(XQ)(Q^\top W)
     =\sum_{j=1}^{d}(Xq_j)(q_j^\top W)
    \label{eq:app-output-decomp}
\end{equation}
The contribution of a single rotation axis $q$ to the projection output is therefore
\begin{equation}
    Y_q=(Xq)(q^\top W)
\end{equation}

We measure this axis's output contribution by the sum of variances across all output dimensions:
\begin{equation}
    e(q)
    =
    \mathbb{E}\left[
        \left\|Y_q-\mathbb{E}[Y_q]\right\|_2^2
    \right]
    =
    \operatorname{tr}\left(\operatorname{Cov}(Y_q)\right).
\end{equation}

Since $W$ is fixed and $q^\top W$ is independent of the sample for a given $q$,

\begin{align}
    e(q)
    &=
    \operatorname{Var}(Xq)
    \|q^\top W\|_2^2
    \\
    &=
    (q^\top S q)
    (q^\top WW^\top q).
    \label{eq:app-output-contribution2}
\end{align}

Defining the weight Gram matrix $M$ as follows rewrites Eq.~(\ref{eq:app-output-contribution2}) as Eq.~(\ref{eq:app-output-contribution}).
\begin{equation}
    M=WW^\top
\end{equation}
\begin{equation}
    e(q)
    =
    (q^\top S q)(q^\top M q)
    \label{eq:app-output-contribution}
\end{equation}
Accounting for projection weights thus extends the activation-only PCA objective $q^\top S q$ to a product of two quadratic forms.

We maximize Eq.~(\ref{eq:app-output-contribution}) subject to $q^\top q=1$. Define
\begin{equation}
    \alpha(q)=q^\top Mq,
    \qquad
    \beta(q)=q^\top Sq
\end{equation}
Then
\begin{equation}
    \nabla_q e(q)
    =
    2\left[
        \alpha(q)S+\beta(q)M
    \right]q.
\end{equation}
The Lagrange stationarity condition is therefore
\begin{equation}
    \left[
        \alpha(q)S+\beta(q)M
    \right]q
    =
    \lambda q
    \label{eq:app-weight-aware-stationary}
\end{equation}

Unlike PCA, Eq.~(\ref{eq:app-weight-aware-stationary}) does not directly provide a single fixed eigensystem, because $\alpha(q)$ and $\beta(q)$ depend on the rotation axis $q$ being sought. For the full rotation $Q=[q_1,\ldots,q_d]$, the coefficients
\begin{equation}
    \alpha_j=q_j^\top Mq_j,
    \qquad
    \beta_j=q_j^\top Sq_j
\end{equation}
vary across axes, so the matrices $\alpha_jS+\beta_jM$ generally differ as well. Optimizing the axes independently does not automatically guarantee mutual orthogonality.

To obtain a single fixed eigensystem, RoShift approximates these axis-dependent coefficients with mixture coefficients shared across all axes. In the following, we use trace-normalized $S$ and $M$ to reduce scale differences between the statistics and construct
\begin{equation}
    B
    =
    \theta_S S
    +
    \theta_M M,
    \qquad
    \theta_S,\theta_M\ge0
\end{equation}

Since $B$ is a fixed symmetric matrix, its eigendecomposition
\begin{equation}
    BQ=Q\Lambda
\end{equation}
provides a complete orthogonal rotation. For a single projection, this extends PCA based on one activation covariance matrix to a construction that jointly considers two statistics: activation covariance and the weight Gram matrix.

\subsubsection{Extension to Multiple Transformer Projections}

In a Transformer block, a single layer-wise rotation $Q_l$ affects multiple projections. Since the Q/K/V and Gate projections to which RoShift applies input sparsity share the same rotation, their output contributions must be considered jointly.

Let $S_l^{\mathrm{att}}$ and $S_l^{\mathrm{ffn}}$ denote the attention-input and FFN-input covariance matrices at layer $l$, respectively. Q/K/V share the attention input, while Gate uses the FFN input. The projection-specific weight Gram matrices are defined as
\begin{align}
    M_l^Q
    &=
    W_l^Q(W_l^Q)^\top,
    &
    M_l^K
    &=
    W_l^K(W_l^K)^\top,
    \\
    M_l^V
    &=
    W_l^V(W_l^V)^\top,
    &
    M_l^G
    &=
    W_l^G(W_l^G)^\top
\end{align}

Extending the single-projection objective in Eq.~(\ref{eq:app-output-contribution}) to these four projections gives the following joint output-contribution objective for an axis $q$:
\begin{equation}
    \begin{aligned}
    E_l(q)
    &=
    (q^\top S_l^{\mathrm{att}} q)
    \left[
        q^\top
        (M_l^Q+M_l^K+M_l^V)
        q
    \right] +
    (q^\top S_l^{\mathrm{ffn}} q)
    (q^\top M_l^G q).
    \end{aligned}
    \label{eq:app-multi-projection-objective}
\end{equation}

Defining
\begin{equation}
    M_l^{\mathrm{att}}
    =
    M_l^Q+M_l^K+M_l^V
\end{equation}
the gradient of Eq.~(\ref{eq:app-multi-projection-objective}) is
\begin{equation}
    \begin{aligned}
    \nabla_q E_l(q)
    =
    2\Big[
        &\alpha_l^{\mathrm{att}}(q)S_l^{\mathrm{att}}
        +
        \beta_l^{\mathrm{att}}(q)M_l^{\mathrm{att}} +
        \alpha_l^{\mathrm{ffn}}(q)S_l^{\mathrm{ffn}}
        +
        \beta_l^{\mathrm{ffn}}(q)M_l^G
    \Big]q,
    \end{aligned}
    \label{eq:app-multi-projection-gradient}
\end{equation}
where
\begin{align}
    \alpha_l^{\mathrm{att}}(q)
    &=
    q^\top M_l^{\mathrm{att}}q,
    &
    \beta_l^{\mathrm{att}}(q)
    &=
    q^\top S_l^{\mathrm{att}}q,
    \\
    \alpha_l^{\mathrm{ffn}}(q)
    &=
    q^\top M_l^Gq,
    &
    \beta_l^{\mathrm{ffn}}(q)
    &=
    q^\top S_l^{\mathrm{ffn}}q.
\end{align}

Expanding $M_l^{\mathrm{att}}$ into the Q/K/V Gram matrices expresses Eq.~(\ref{eq:app-multi-projection-gradient}) as a $q$-dependent linear combination of six statistics:
\begin{equation}
    S_l^{\mathrm{att}},
    \quad
    S_l^{\mathrm{ffn}},
    \quad
    M_l^Q,
    \quad
    M_l^K,
    \quad
    M_l^V,
    \quad
    M_l^G
    \label{eq:app-six-statistics}
\end{equation}
Thus, the two statistics for a single projection, activation covariance and the weight Gram matrix, become six statistics when a layer-wise rotation is shared across multiple Transformer projections: two activation covariance matrices and four weight Gram matrices.

As in the single-projection case, the coefficients still depend on the rotation axis $q$, so they do not directly yield a fixed eigensystem. RoShift approximates these axis-dependent coefficients with mixture coefficients shared within each layer.

Denote the six statistics by $A_{l,i}$, $i=1,\ldots,6$. To reduce differences in scale, we apply trace normalization:
\begin{equation}
    B_{l,i}
    =
    \frac{A_{l,i}}
         {\operatorname{tr}(A_{l,i})}
\end{equation}
We then construct
\begin{equation}
    B_l
    =
    \sum_{i=1}^{6}
    \theta_{l,i}B_{l,i},
    \qquad
    \theta_{l,i}\ge0
    \label{eq:app-final-mixture}
\end{equation}
and use the eigenvectors of $B_l$ in descending eigenvalue order as the columns of $Q_l$.

To account for RMSNorm, activation covariances are computed from activations normalized by inverse RMS. The learned channel-wise scales are absorbed into the corresponding projection weights before constructing their Gram matrices.

In summary, RoShift's rotation construction extends progressively as follows.
\begin{equation}
    \underbrace{S}_{\text{Activation-only PCA}}
    \quad\longrightarrow\quad
    \underbrace{\{S,M\}}_{\text{Single projection}}
    \quad\longrightarrow\quad
    \underbrace{
        \{
        S_l^{\mathrm{att}},
        S_l^{\mathrm{ffn}},
        M_l^Q,
        M_l^K,
        M_l^V,
        M_l^G
        \}
    }_{\text{Transformer layer}}.
\end{equation}

\subsection{Sparse LM-head}
\label{app:sparse-lm-head}

\subsubsection{LM-head Sparsification}
The LM-head maps the final activations to logits over the entire vocabulary to determine the next token. This does not require completing the computation for every candidate. Greedy decoding selects the candidate with the largest logit, while Top-K and Top-P sampling primarily use a subset of high-logit candidates. Instead of applying the same input sparsity to every candidate, RoShift reduces LM-head weight reads and computations by terminating computation early for candidates unlikely to be selected.

\subsubsection{Rotation for Candidate Selection}
RoShift constructs a dedicated orthogonal LM-head rotation from final-activation statistics, concentrating energy in the leading dimensions. This rotation is also absorbed into the corresponding LM-head weights, so the full logit computation before candidate pruning is algebraically equivalent to the original computation. In this representation, partial logits computed from the leading dimensions provide rough estimates of final logits. Computation for a vocabulary candidate is therefore terminated if its intermediate logit is too small.

\subsubsection{Staged Candidate Selection}
\begin{wrapfigure}[18]{r}{0.40\textwidth}
    \centering
    \vspace{-15pt}        \captionsetup{font=small,skip=3pt}
    \includegraphics[width=\linewidth]{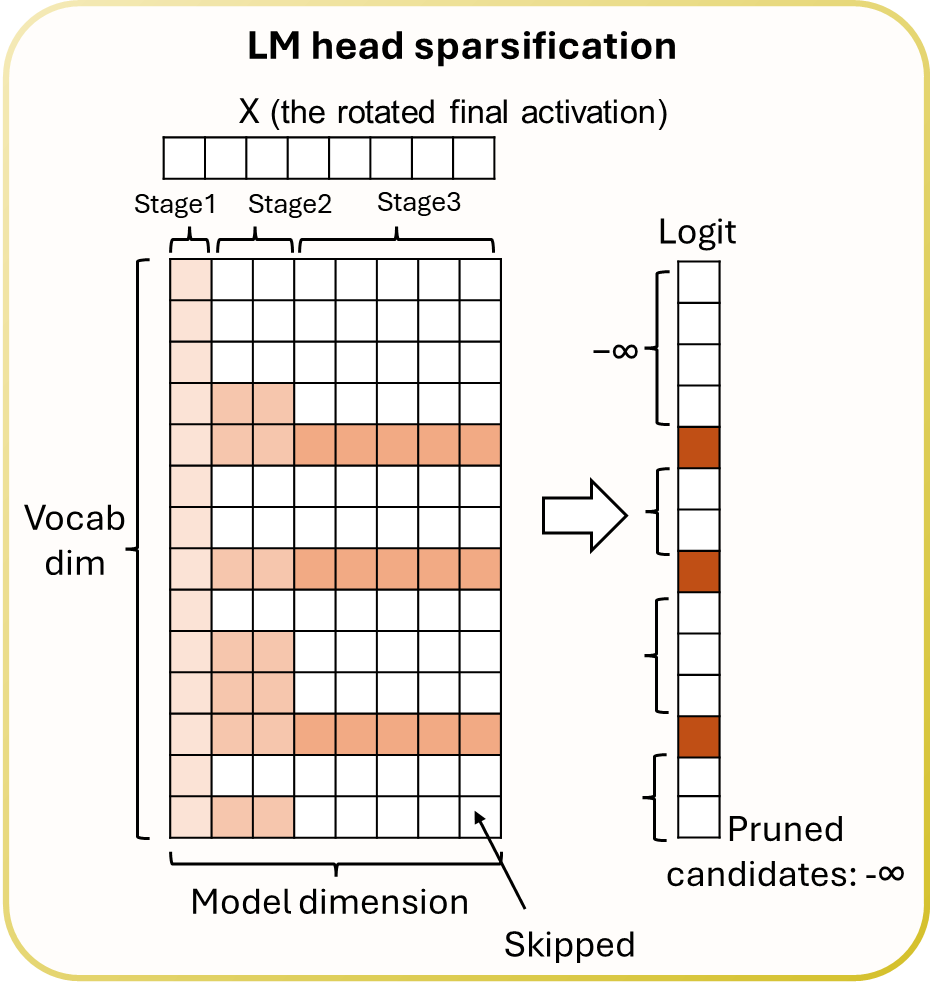}
    \caption{Three-stage LM-head candidate selection. Only shaded regions are computed. Subsequent computations for pruned candidates are skipped, and their logits are set to $-\infty$. 
    }
\label{fig:sparse-lm-head-stages}
    \vspace{-8pt}
\end{wrapfigure}
Figure~\ref{fig:sparse-lm-head-stages} illustrates the computations skipped through staged candidate selection. We divide the rotated input dimensions into three segments. The first stage computes partial logits for all vocabulary candidates. Let $m$ and $\sigma$ be the maximum and standard deviation of the accumulated partial logits among the current candidates. Only candidates with partial logits at least $m-z\sigma$ advance to the next stage. The stage-specific coefficients $z$ and segment boundaries are determined through calibration. We accumulate contributions from the next segment for the surviving candidates and repeat the selection process. Only final survivors complete the remaining dimensions to obtain full logits. Subsequent computations for pruned candidates are skipped, and their final logits are set to $-\infty$. Finer segmentation can eliminate candidates earlier and skip more computation, but incurs overhead from inspecting accumulated logits and selecting candidates after each segment. We therefore use the three-stage configuration that empirically provides the best trade-off.

\subsubsection{Experimental Validation}
For all models, we calibrate the selection criteria so that candidate pruning changes the Top-1 prediction, the candidate with the largest logit, for fewer than 0.1\% of calibration samples. Table~\ref{tab:head-calibration-summary} reports the resulting effective LM-head sparsity for each model. We define effective LM-head sparsity as the fraction of dense LM-head MACs skipped through candidate termination. 

Table~\ref{tab:head-rotation-ablation} compares effective LM-head sparsity with and without rotation. RoShift's rotation achieves higher sparsity for all five models, with particularly large gains for Llama-2-7B and Qwen2.5-7B. For Qwen3, the aggregate calibration Top-1 change is matched at 0.05371\% relative to each basis's no-pruning reference, without a per-domain cap; no-rotation inputs are reconstructed from stored rotated activations.

\begin{center}
\begin{minipage}{\linewidth}
\centering
\fontsize{9}{10.8}\selectfont
\captionof{table}{Effective LM-head sparsity and pruning-induced Top-1 prediction changes on the calibration set. Changes are measured relative to a no-pruning path with the same rotation, three-stage partition, and rounding.}
\label{tab:head-calibration-summary}
\setlength{\tabcolsep}{10pt}
\begin{tabular}{lrr}
\toprule
Model & Effective sparsity (\%) & Top-1 change (\%) \\
\midrule
Llama-3.1-8B & 83.03 & 0.049 \\
Llama-2-7B & 75.82 & 0.034 \\
Mistral-7B & 71.64 & 0.048 \\
Qwen2.5-7B & 75.06 & 0.031 \\
Qwen3-8B & 71.42 & 0.054 \\
\bottomrule
\end{tabular}
\end{minipage}
\end{center}

\begin{center}
\begin{minipage}{\linewidth}
\centering
\captionof{table}{Effective LM-head sparsity (\%) with and without RoShift's LM-head rotation. Increase is the difference between the two sparsity percentages (RoShift rotation minus no rotation).}
\label{tab:head-rotation-ablation}
\fontsize{9}{10.8}\selectfont
\setlength{\tabcolsep}{10pt}
\begin{tabular}{lrrr}
\toprule
Model & No rotation & RoShift rotation & Increase \\
\midrule
Llama-3.1-8B & 56.02 & \textbf{83.03} & +27.01 \\
Llama-2-7B & 0.65 & \textbf{75.82} & +75.17 \\
Mistral-7B & 55.17 & \textbf{71.64} & +16.47 \\
Qwen2.5-7B & 1.84 & \textbf{75.06} & +73.22 \\
Qwen3-8B-Base & 43.66 & \textbf{71.42} & +27.76 \\
\bottomrule
\end{tabular}
\end{minipage}
\end{center}

\paragraph{TEAL-style input sparsification.}
We also evaluate simple TEAL-style input sparsification~\citep{liu2025teal} applied to the LM-head without rotation. Under a Top-1 prediction change constraint of less than 0.1\%, the resulting effective LM-head sparsity ranges from 1.98\% to 2.93\% across the five models (Table~\ref{tab:head-input-sparsity}). Thus, this simple input sparsification offers limited scope for reducing LM-head computation under the evaluated constraint.

\begin{center}
\begin{minipage}{\linewidth}
\centering
\captionof{table}{TEAL-style LM-head input sparsification without rotation, subject to a Top-1 prediction change below 0.1\%. Effective sparsity is the fraction of dense LM-head MACs skipped by input sparsification.}
\label{tab:head-input-sparsity}
\fontsize{9}{10.8}\selectfont
\setlength{\tabcolsep}{10pt}
\begin{tabular}{lrr}
\toprule
Model & Effective sparsity (\%) & Top-1 change (\%) \\
\midrule
Llama-3.1-8B & 1.98 & 0.098 \\
Llama-2-7B & 2.12 & 0.090 \\
Mistral-7B & 2.93 & 0.099 \\
Qwen2.5-7B & 2.62 & 0.099 \\
Qwen3-8B-Base & 2.03 & 0.098 \\
\bottomrule
\end{tabular}
\end{minipage}
\end{center}

\paragraph{Downstream generation accuracy.}
We further evaluate LM-head sparsification on MBPP, GSM8K, and MATH-500 using Llama-3.1-8B, keeping the same RoShift Transformer at 50\% non-uniform sparsity and changing only the LM-head. In addition to greedy decoding, we test two sampling configurations using temperature scaling and Top-P/Top-K filtering (Table~\ref{tab:head-generation-accuracy}). Average accuracy is largely preserved across these decoding settings, with changes ranging from $-0.39$ to $+0.47$ percentage points relative to the matched dense LM-head.

\begin{center}
\begin{minipage}{\linewidth}
\centering
\captionof{table}{Dense versus sparse LM-head accuracy (\%) on Llama-3.1-8B with the same RoShift Transformer at 50\% non-uniform sparsity. Parentheses show differences in percentage points from the dense head under the same decoding setting. Avg is the unweighted mean of the three benchmarks.}
\label{tab:head-generation-accuracy}
\fontsize{9}{10.8}\selectfont
\setlength{\tabcolsep}{3pt}
\renewcommand{\arraystretch}{1.12}
\begin{tabular*}{\linewidth}{@{\extracolsep{\fill}}lrrrr@{}}
\toprule
Method & MBPP & GSM8K & MATH-500 & Avg \\
\midrule
Dense head (Greedy) & 44.60 & 45.03 & 18.20 & 35.94 \\
Sparse head (Greedy) & 44.60 (+0.00) & 44.66 (-0.38) & 18.00 (-0.20) & \textbf{35.75 (-0.19)} \\
\midrule
Dense head (Sampling A) & 41.73 & 42.43 & 15.27 & 33.14 \\
Sparse head (Sampling A) & 40.07 (-1.67) & 43.72 (+1.29) & 14.47 (-0.80) & \textbf{32.75 (-0.39)} \\
\midrule
Dense head (Sampling B) & 42.53 & 43.06 & 16.20 & 33.93 \\
Sparse head (Sampling B) & 43.20 (+0.67) & 43.62 (+0.56) & 16.40 (+0.20) & \textbf{34.41 (+0.47)} \\
\bottomrule
\end{tabular*}
\par\smallskip
\raggedright
Sampling A: $T=0.6$, Top-P $=0.9$, Top-K $=0$ (disabled).
Sampling B: $T=0.7$, Top-P $=0.8$, Top-K $=20$.
Generation is capped at 256 tokens. Sampling scores use each benchmark's full evaluation set and are averaged over seeds 0, 1, and 2.
\end{minipage}
\end{center}

\subsection{Kernel Details}
\label{app:kernel-details}

RoShift's kernels translate activation sparsity into actual reductions in weight reads and computations. Blockwise Top-K reduces selection cost, and sparse GEMV operates only on selected components. Sparsity-aware weight layouts and 2D tiling improve execution, while kernel fusion and multiple streams reduce the overhead of additional operations.

\subsubsection{Blockwise Top-K}
Global Top-K requires a global reduction to merge results from multiple GPU thread blocks, potentially incurring substantial selection overhead. RoShift uses B128 Blockwise Top-K to select components independently within blocks of 128 activation dimensions. A fixed retained count per block keeps the subsequent sparse GEMV workload consistent. At 25\%, 50\%, and 75\% sparsity, each block retains 96, 64, and 32 components, respectively.

\noindent
\begin{minipage}[t]{0.55\textwidth}
    \vspace{0pt}
If important components concentrate in a few blocks, however, accuracy can be lower than with Global Top-K. To mitigate this, we form selection blocks by applying a fixed random permutation to the rotated activation dimensions. This permutation can be incorporated into the corresponding rotation and weights offline. As Table~\ref{tab:blockwise-topk-ppl} shows, the PPL increase over Global Top-K is small under the evaluated conditions.
\end{minipage}\hfill
\begin{minipage}[t]{0.43\textwidth}
    \vspace{0pt}
    \centering
    \captionsetup{font=tablefont,skip=4pt}
    \captionof{table}{PPL comparison of Dense, Global Top-K, and B128 Blockwise Top-K on Llama-3.1-8B using the calibration set.}
\label{tab:blockwise-topk-ppl}
    \fontsize{9}{10.8}\selectfont
    \setlength{\tabcolsep}{5pt}
    \renewcommand{\arraystretch}{1.10}
    \begin{tabular}{@{}ccc@{}}
        \toprule
        Sparsity & Global Top-K & B128 Top-K \\
        \midrule
        Dense & 4.99 & 4.99 \\
        \midrule
        40\% & 5.32 & 5.34 \\
        50\% & 5.63 & 5.70 \\
        60\% & 6.33 & 6.40 \\
        \bottomrule
    \end{tabular}
\end{minipage}
\par

\subsubsection{Index-Based Sparse GEMV and 2D Tiling}
RoShift adapts oFFN's selective weight loading and index-based sparse execution to its input and output sparsity patterns~\citep{song2026offn}. Instead of assigning work to all dimensions and skipping pruned components, the kernels collect retained indices and perform weight reads and computations only for those components.

Input-sparse GEMV uses 2D tiling across groups of selected inputs and output segments. Each CUDA thread block computes the partial contribution of an input group to an output segment, and partial sums from different input groups are combined using atomic additions. This exposes parallelism along both the input and output dimensions. For output-sparse Up, work is assigned only to selected output channels, and each channel's dot product is computed.

Weight layouts are prepared offline to match these access directions. For input-sparse projections, weights associated with one input component are contiguous along the output dimension. For output-sparse Up, weights associated with one output channel are contiguous along the input dimension. Thus, even though selected indices are irregular, neighboring threads access contiguous weights for each selected component.

\subsubsection{Multi-Stream Execution}
The sparse FFN and correction path share the same FFN input but do not depend on each other's outputs. RoShift therefore executes them on separate CUDA streams once the input is ready, using CUDA events to synchronize immediately before summing their outputs.

\subsubsection{Decoding Measurements}
Dense, TEAL, and RoShift are evaluated in a gpt-fast-based execution environment~\citep{pytorch2023gptfast}. Dense uses dense projections, TEAL uses threshold-based Triton sparse GEMV, and RoShift integrates the kernels described above into the same environment. Dense and TEAL use dense LM-heads, whereas RoShift uses a separately calibrated sparse LM-head.

The results in Table~\ref{tab:token-generation-speed} are measured with FP16 and batch size~1. For each of WikiText-2~\citep{merity2016pointer}, Alpaca, GSM8K, and OpenBookQA (OBQA)~\citep{mihaylov2018openbookqa}, we use one fixed prompt and teacher-forced continuation. Following a 1,024-token dense prefill, CUDA events measure only the 128-token decoding interval. All methods use Math SDPA, \texttt{torch.compile}, and CUDA Graphs. We average per-token latency over four repeated measurements per dataset and across all datasets.

\clearpage
\begingroup
\raggedbottom
\subsection{Sparsity Allocation and Rotation Search}
\label{app:search-details}

RoShift first determines the sparsity allocation under a target MAC sparsity, then fixes it before searching for rotations. Both searches screen candidates using local reconstruction and select the final configuration based on full-model calibration PPL.

\subsubsection{Sparsity Allocation}
\vspace{-0.4em}
\noindent
\begin{minipage}[t]{0.53\textwidth}
\vspace{0pt}
\normalsize
TEAL assigns different sparsities to projections within each block but keeps total sparsity uniform across blocks~\citep{liu2025teal}. WAS permits inter-block allocation but restricts sparsity to be monotonically non-decreasing with depth~\citep{wang2025was}. These constraints reduce the search space but cannot express allocations with lower sparsity in later layers. RoShift is designed to explore such allocations while preserving the overall MAC sparsity constraint.

\medskip
Jointly searching all layer and projection sparsities, however, leads to a rapidly growing number of candidate configurations. We therefore initialize with TEAL-style within-layer allocation and WAS-style inter-layer allocation, then refine the result using pairwise coordinate descent that transfers equal MAC budgets between two coordinates. This limits the variables changed in each update while allowing both layer-wise total sparsity and within-layer allocation to be adjusted.
\end{minipage}
\hfill
\begin{minipage}[t]{0.44\textwidth}
\vspace{0pt}
\captionsetup{font=small,labelfont=bf,skip=4pt}
\captionof{algorithm}{Non-uniform Sparsity Allocation}
\label{alg:sparsity-allocation}
\hrule
\fontsize{8}{9.5}\selectfont
\raggedright
\algrenewcommand\textproc[1]{\textnormal{#1}}
\algrenewcommand\algorithmicindent{0.8em}
\algrenewcommand\alglinenumber[1]{\makebox[1.6em][r]{#1:}}
\begin{algorithmic}[1]
\Require Target sparsity $r$, calibration data $D$, short calibration data $D_s$, transfer grid $G$
\State $s\gets$ TEAL-style allocation followed by WAS-style allocation at target $r$
\State $p\gets\operatorname{PPL}(s,D)$
\While{search budget remains}
    \State $(\mathcal{D},\mathcal{R})\gets\Call{LocalErrorShortlist}{s,D}$
    \State $C\gets\Call{FeasiblePairTransfers}{s,\mathcal{D},\mathcal{R},G}$
    \Statex \hspace{\algorithmicindent}\textit{Preserve MAC budget}
    \State $C\gets\Call{ShortPPLScreen}{C,D_s}$
    \Statex \hspace{\algorithmicindent}\textit{Retain each pair's best grid point and neighbors.}
    \If{$C$ is empty}
        \State \textbf{break}
    \EndIf
    \State $s_{\mathrm{new}}\gets\arg\min_{c\in C}\operatorname{PPL}(c,D)$
    \State $p_{\mathrm{new}}\gets\operatorname{PPL}(s_{\mathrm{new}},D)$
    \If{the improvement criterion relative to $p$ is not met}
        \State \textbf{break}
    \EndIf
    \State $(s,p)\gets(s_{\mathrm{new}},p_{\mathrm{new}})$
\EndWhile
\State \Return $s$
\end{algorithmic}
\hrule
\end{minipage}
\par\vspace{-0.5\baselineskip}
\medskip
To reduce the cost of evaluating full calibration PPL for every candidate, we screen candidates successively using local error and short calibration PPL. Because local error alone cannot adequately capture error accumulation across layers, the final update is accepted only if it improves full calibration PPL. Algorithm~\ref{alg:sparsity-allocation} gives the procedure.

\subsubsection{Rotation Search}
\vspace{-0.4em}
\noindent
\begin{minipage}[t]{0.53\textwidth}
\vspace{0pt}
\normalsize
An \ymyi{optimal} sparsity-friendly rotation \ymyi{would vary across projections and} depend on the sparsity assigned to each projection. 
We therefore search \ymyi{for} layer-specific mixture coefficients under the fixed \ymyi{sparsity} allocation \ymyi{to derive a single rotation for each layer}. Determining allocation and rotation sequentially reduces the search space and ensures that rotation candidates are compared under identical sparsity settings.

\medskip
Rather than evaluating all mixtures of the six components, rotation search redistributes mass between pairs of coefficients while preserving their sum. For each mixture, we eigendecompose the weighted sum of the statistic matrices and evaluate the candidate using the block output's relative L2 error against the dense output, $E_l(\theta)$. We repeatedly update the mixture while local error improves, reducing the step size when improvement stops or the update limit is reached.

\medskip
Rotations that improve individual blocks need not provide the same benefit when combined. We therefore
validate the candidates from shallow to deep layers, 
\end{minipage}\hfill
\begin{minipage}[t]{0.44\textwidth}
\vspace{-3pt}
\captionsetup{font=small,labelfont=bf,skip=4pt}
\captionof{algorithm}{Layer-wise Rotation Search}
\label{alg:rotation-search}
\hrule
\fontsize{8}{9.5}\selectfont
\raggedright
\algrenewcommand\textproc[1]{\textnormal{#1}}
\algrenewcommand\algorithmicindent{0.8em}
\algrenewcommand\alglinenumber[1]{\makebox[1.6em][r]{#1:}}
\begin{algorithmic}[1]
\Require Fixed $s$, statistics $B$, calibration $D$, decreasing steps $\Delta$, update limit $K$
\State $\theta_0\gets(0.25,0.25,0.125,0.125,0.125,0.125)$
\State $Q\gets\Call{Eigenbases}{B,\theta_0}$
\For{each layer $l$}
    \State $\theta\gets\theta_0$; $e\gets E_l(\theta)$
    \For{$\delta\in\Delta$} \Comment{Coarse to fine}
        \For{$t=1,\ldots,K$}
            \State $H\gets\Call{FeasiblePairTransfers}{\theta,\delta}$
            \State \textbf{if} $H=\varnothing$ \textbf{then break}
            \State $\theta'\gets\arg\min_{h\in H}E_l(h)$; $e'\gets E_l(\theta')$
            \State \textbf{if} $e'\ge e-10^{-12}$ \textbf{then break}
            \State $(\theta,e)\gets(\theta',e')$
        \EndFor
    \EndFor
    \State $\widehat{Q}_l\gets\Call{Eigenbasis}{B_l,\theta}$
\EndFor
\State $p\gets\operatorname{PPL}(s,Q,D)$
\For{$l=1,\ldots,L$} \Comment{One sequential pass}
    \State $Q_{\mathrm{trial}}\gets Q$ with $Q_l\gets\widehat{Q}_l$
    \State $p_{\mathrm{trial}}\gets\operatorname{PPL}(s,Q_{\mathrm{trial}},D)$
    \If{$p_{\mathrm{trial}}<p-10^{-12}$}
        \State $(Q,p)\gets(Q_{\mathrm{trial}},p_{\mathrm{trial}})$
    \EndIf
\EndFor
\State \Return $Q$
\end{algorithmic}
\hrule
\end{minipage}
\par\vspace{-0.4\baselineskip}
retaining each only if it improves full-model calibration PPL with all previously accepted changes in place. 
This sequential selection is performed exactly once.

\subsubsection{Search Time}
\label{app:search-time}
On a single RTX 5090, sparsity allocation and rotation search take approximately 4.6 and 0.7 hours per sparsity setting, respectively, averaged across the five models. Table~\ref{tab:search-time} reports the per-model estimates.

\begin{table}[htbp]
    \centering
    \caption{Approximate search time (hours) per sparsity setting on a single RTX 5090, averaged over 40\%, 50\%, and 60\% sparsity using combined local and server run records.}
    \label{tab:search-time}
    \fontsize{8}{9}\selectfont
    \setlength{\tabcolsep}{10pt}
    \renewcommand{\arraystretch}{1.12}
    \begin{tabular}{lcc}
        \toprule
        Model & Sparsity allocation & Rotation search \\
        \midrule
        Llama-3.1-8B & 5.2 & 0.8 \\
        Llama-2-7B & 6.3 & 0.6 \\
        Mistral-7B & 4.0 & 0.7 \\
        Qwen2.5-7B & 4.8 & 0.7 \\
        Qwen3-8B & 2.6 & 0.8 \\
        \midrule
        Model average & 4.6 & 0.7 \\
        \bottomrule
    \end{tabular}
\end{table}

\endgroup

\subsection{Detailed Experimental Settings}
\label{app:experimental-settings}

\subsubsection{General Experimental Settings}
\label{app:general-experimental-settings}

We evaluate five SiLU-based base LLMs: \textbf{Llama-3.1-8B, Llama-2-7B, Mistral-7B-v0.1, Qwen2.5-7B, and Qwen3-8B-Base}~\citep{grattafiori2024llama,touvron2023llama,jiang2023mistral,yang2024qwen25,yang2025qwen3}. Accuracy experiments use target sparsities of 40\%, 50\%, and 60\% and report the unweighted mean over 11 downstream benchmarks. All RoShift experiments use the same calibration set of 32 sequences of 4,096 tokens each: 16 from C4, six from Alpaca, five from MetaMathQA, and five from CodeAlpaca-20K~\citep{raffel2020exploring,alpaca,yu2023metamath,codealpaca}. For the baselines, we use the calibration datasets and settings specified by their respective public implementations, following each method’s intended calibration recipe rather than forcing a common calibration distribution.

Throughout our experiments, MAC sparsity denotes the fraction of linear-projection MACs eliminated by sparsification. This definition follows the same model-level sparsity formulation used by TEAL and WINA, accounting for the computational cost of each projection and thus directly reflecting the amount of computation skipped through activation sparsification.

Benchmark accuracy evaluation is based on EleutherAI's lm-evaluation-harness (lm-eval)~\citep{gao2023evalharness}. We evaluate MMLU, ARC-Easy, ARC-Challenge, BoolQ, PIQA, WinoGrande, CommonsenseQA, BBH, GSM8K, MATH-500, and MBPP~\citep{hendryckstest2021,allenai:arc,clark2019boolq,Bisk2020,ai2:winogrande,talmor-etal-2019-commonsenseqa,suzgun2022challenging,cobbe2021gsm8k,lightman2023lets,austin2021program}. For log-likelihood evaluation, we use 0-shot prompting on PIQA, BoolQ, ARC-Easy, and CommonsenseQA, 5-shot prompting on WinoGrande and MMLU, and 25-shot prompting on ARC-Challenge. For generation, we use 3-shot prompting on MBPP and BBH, 8-shot prompting on GSM8K, and 4-shot prompting on MATH-500.

For log-likelihood benchmarks, Transformer sparsity is applied at every input-token position, while the LM-head remains dense. For generation benchmarks, the entire prompt prefill is dense, and Transformer sparsity is applied at every decoding step. RoShift uses a separately calibrated sparse LM-head during decoding, with its own sparsity setting independent of Transformer sparsity, as reported in Table~\ref{tab:head-calibration-summary}. TEAL, LaRoSA, and WINA use the same prefill and decoding sparsity scope but keep the LM-head dense. 
The target MAC sparsity reported in Table~\ref{tab:model-accuracy-summary} is defined over the Transformer blocks only and excludes LM-head sparsity; thus, RoShift’s sparse LM-head does not contribute to the sparsity target used for the accuracy comparison.

\clearpage

\subsubsection{Baseline Implementations and Modifications}
\label{app:baseline-implementations}

We use TEAL as the primary speed baseline because it applies to SiLU-based LLMs without additional training and provides public FP16 sparse kernels and an end-to-end decoding implementation. We additionally include LaRoSA and WINA as accuracy baselines~\citep{liu2025larosa,chen2026wina}. LaRoSA uses layer-wise rotations to concentrate activation energy. WINA combines activation magnitudes with the corresponding weight-column $\ell_2$ norms and uses an offline SVD-based orthogonal transformation to enforce column orthogonality for its theoretical analysis. Thus, WINA is also related to RoShift through its use of weight information and rotation for activation sparsification.

We follow each baseline's public implementation and settings as closely as possible, with adjustments to token scope, runtime precision, and sparsity allocation. Regarding token scope, TEAL~\citep{liu2025teal} and WINA~\citep{chen2026wina} exclude some initial tokens from sparsification in their original prefill evaluation settings because effects such as attention sinks~\citep{xiao2024attention} can cause greater degradation at these positions. In our log-likelihood evaluation, we instead apply sparsity at every input-token position to evaluate robustness under the common token scope described in Section~\ref{app:general-experimental-settings}. For generation benchmarks, the entire prompt prefill remains dense for all methods.

LaRoSA's public implementation applies the rotation matrix $Q$ immediately before Top-K and restores the original basis after sparsification, rather than retaining activations in the rotated basis throughout execution. These runtime rotations use FP64; we change them to FP16 to reduce experimental runtime. We retain the original precision for the offline statistics used to construct the rotations and leave other operations unchanged. For comparison, RoShift absorbs its layer-wise rotations into the model weights offline; its online rotation-related operation is the transition between consecutive rotated bases, $Q_l^\top Q_{l+1}$, which is likewise executed in FP16.

For sparsity allocation, LaRoSA's original policy measures sparsity by counting activation elements. Because projections have different computational costs, this can yield a MAC reduction below the target sparsity. We therefore modify its allocation policy to match the target MAC reduction within the Transformer, consistent with the comparison criterion in Section~\ref{app:general-experimental-settings}. TEAL and RoShift also use MAC-sparsity-based allocation. Table~\ref{tab:larosa-original-allocation} separately reports benchmark accuracy and MAC sparsity under LaRoSA's original allocation policy.

WINA's public evaluation setup~\citep{chen2026wina} uses activation-magnitude-based Top-K for its TEAL baseline, whereas we follow the original TEAL paper and official implementation~\citep{liu2025teal} by using fixed thresholds determined from calibration data. Under fixed thresholds, the achieved MAC sparsity may deviate from the target depending on the activation distribution of the evaluation data. Table~\ref{tab:teal-achieved-sparsity} therefore reports TEAL's achieved MAC sparsity averaged over the 11 benchmarks. When comparing the relative performance of TEAL and WINA reported in our study and the original WINA paper, these differences in baseline implementation and achieved sparsity should be taken into account.

\begin{table}[htbp]
    \centering
    \caption{Measured Transformer MAC sparsity (\%) of TEAL, averaged over 11 benchmarks. Column headers indicate the target MAC sparsity.}
    \label{tab:teal-achieved-sparsity}
    \fontsize{8}{9}\selectfont
    \setlength{\tabcolsep}{12pt}
    \renewcommand{\arraystretch}{1.12}
    \begin{tabular}{lccc}
        \toprule
        Model & 40\% & 50\% & 60\% \\
        \midrule
        Llama-3.1-8B & 38.78 & 48.24 & 57.89 \\
        Llama-2-7B & 40.00 & 49.43 & 59.14 \\
        Mistral-7B-v0.1 & 39.89 & 49.42 & 59.25 \\
        Qwen2.5-7B & 38.81 & 48.18 & 57.94 \\
        Qwen3-8B-Base & 38.39 & 47.89 & 57.39 \\
        \midrule
        \textbf{Model average} & \textbf{39.17} & \textbf{48.63} & \textbf{58.32} \\
        \bottomrule
    \end{tabular}
\end{table}

\clearpage
\subsection{Fused-matrix Cache and RMS Tolerance}
\label{app:cache-tolerance}

\subsubsection{Tolerance Calibration and Representative Allocation}
\label{app:rms-tolerance-calibration}

For each layer, RoShift stores fused matrices corresponding to a small set of RMS representatives obtained during calibration. At inference time, a fused matrix is used only if the relative difference between the current RMS and its representative is at most 5\%; otherwise, execution falls back to exact correction using the current RMS.

\begin{figure}[!htb]
    \centering
    \includegraphics[width=0.49\linewidth]{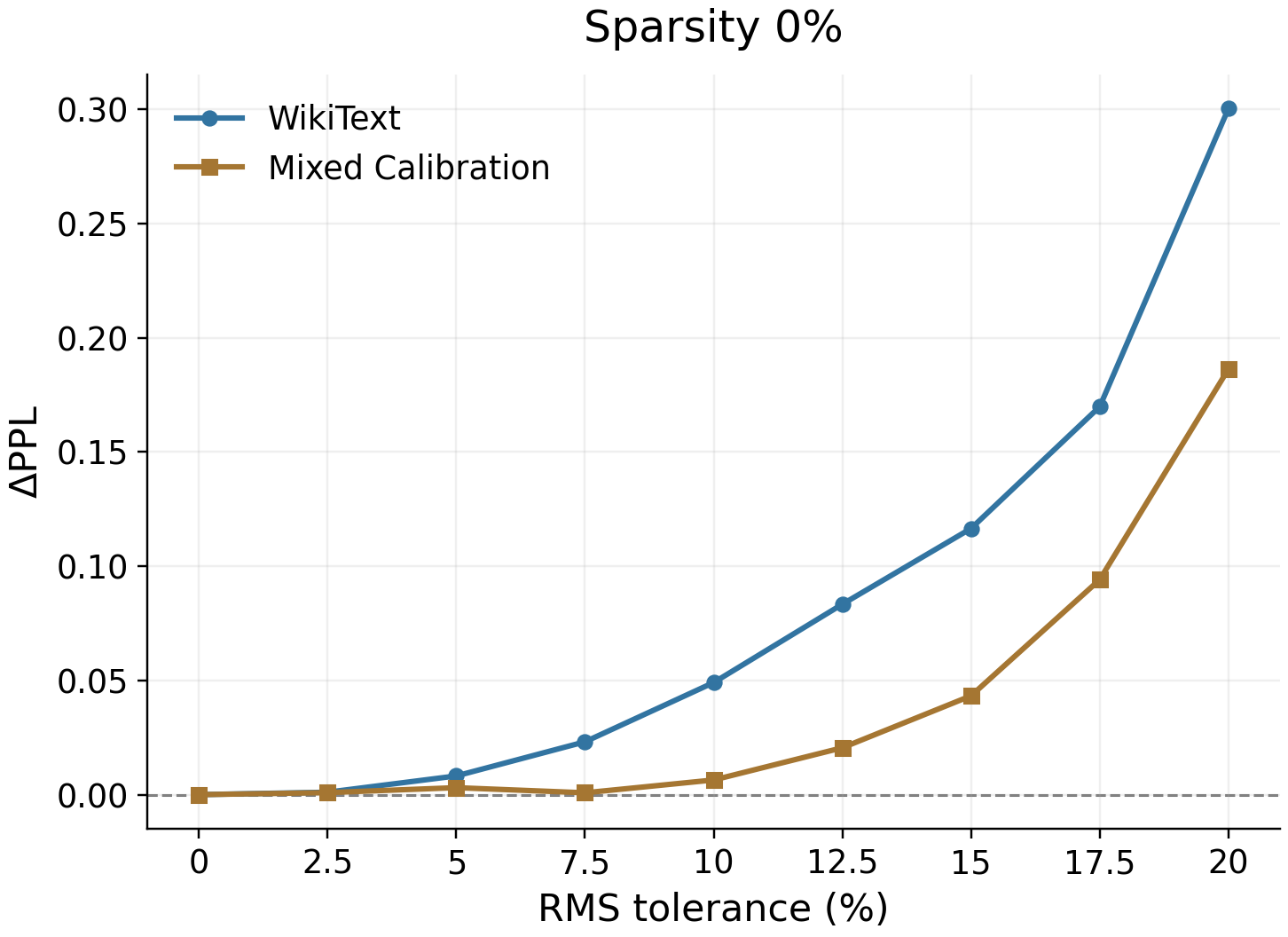}\hfill
    \includegraphics[width=0.49\linewidth]{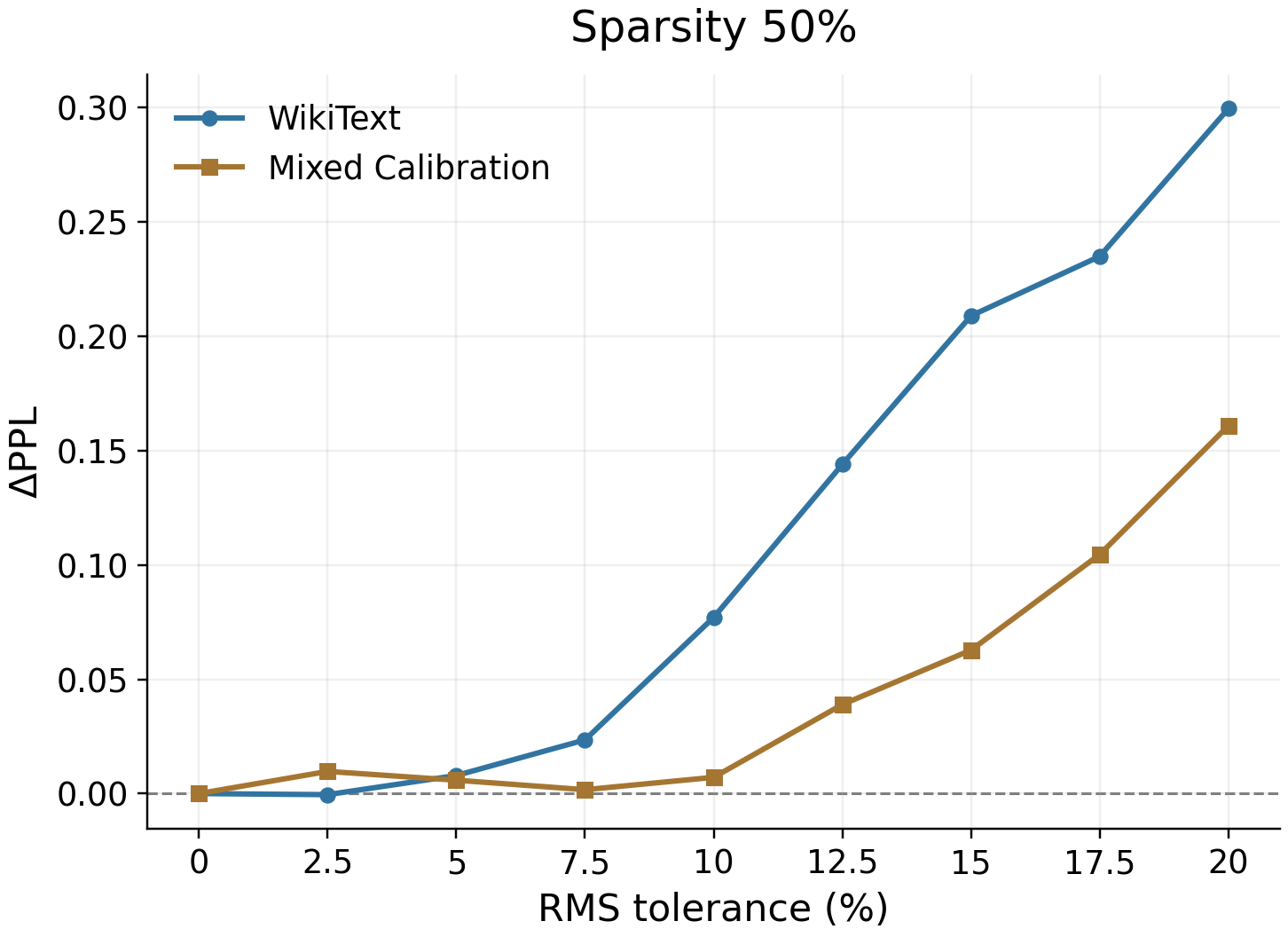}
    \caption{PPL changes versus relative RMS tolerance for Llama-3.1-8B at 0\% (left) and 50\% (right) sparsity, evaluated on WikiText~\citep{merity2016pointer} and Mixed Calibration. $\Delta$PPL is the difference from the 0\%-tolerance result in each setting.}
\label{fig:rms-tolerance-ppl}
\end{figure}

The 5\% relative tolerance was selected through calibration experiments to avoid degrading model PPL. Figure~\ref{fig:rms-tolerance-ppl} shows how PPL changes with the allowed relative error at 0\% and 50\% sparsity. Performance remains essentially unchanged at 5\% in nearly all evaluated cases, supporting our choice of this tolerance. We select the 5\% tolerance on Llama-3.1-8B and then fix it across all models.

Both the representatives and their counts per layer are determined from the calibration distribution. 
For each layer, we consider a set of candidate counts. For each fixed count, we select representative values that maximize the number of calibration samples covered within the tolerance. We then use these values in dynamic programming to allocate representative counts across layers so as to minimize the total number of fallbacks subject to the overall fused-matrix storage budget.

\Needspace{40\baselineskip}
\subsubsection{RMS Tolerance and Fallback Implementation}
\label{app:rms-fallback-implementation}

This section describes the kernel-level implementation of correction and RMS-dependent fallback. We first compare the current token's inverse RMS with all stored representatives to find the nearest one. The inverse RMS is reused from the normalization step, and a fallback flag is set if its relative difference from the nearest representative exceeds 5\%.

This flag determines which matrices the kernel reads. If the difference is within the tolerance and no fallback occurs, the kernel reads the precomputed fused matrix. Otherwise, it reads the transition and correction matrices and applies correction using the current token's inverse RMS rather than the representative. Specifically, \texttt{tl.where} selects the address of either the fused matrix or the transition matrix, while the correction matrix's \texttt{tl.load} is masked by the fallback flag. The correction matrix is therefore additionally read only when fallback occurs.

Comparing the representatives, determining whether fallback is required, and computing the selected path all take place within a single Triton kernel. There is no need to transfer the fallback decision to the CPU or launch a separate kernel depending on that decision. All checks and fallback handling are included in the reported end-to-end decoding latency.

The following excerpt from the actual kernel shows representative comparison, fallback detection, conditional matrix loads, and computation. Details such as tile setup, address calculation, and input loads are omitted.

\noindent\begin{minipage}{\linewidth}
\begin{lstlisting}[
    language=Python,
    basicstyle=\ttfamily\fontsize{8.5}{10}\selectfont,
    numbers=none,
    backgroundcolor={\color[gray]{0.96}},
    keywordstyle={\color[RGB]{40,73,109}\bfseries},
    commentstyle={\color[RGB]{76,105,86}\itshape},
    stringstyle={\color[RGB]{145,72,50}},
    emph={tl},
    emphstyle={\color[RGB]{35,105,120}},
    showstringspaces=false,
    keepspaces=true,
    columns=fullflexible,
    breaklines=true,
    frame=single,
    rulecolor={\color[gray]{0.85}},
    framerule=0.3pt,
    framesep=6pt,
    xleftmargin=6pt,
    xrightmargin=6pt,
    aboveskip=6pt,
    belowskip=4pt
]
alpha = tl.load(alpha_ptr).to(tl.float32)
best = tl.abs(alpha - tl.load(centers))
bucket = tl.full((), 0, tl.int64)

for index in tl.static_range(1, NUM_CENTERS):
    distance = tl.abs(alpha - tl.load(centers + index))
    choose = distance < best
    best = tl.where(choose, distance, best)
    bucket = tl.where(choose, index, bucket)

fallback = best / tl.maximum(tl.abs(alpha), 1.0e-12) > FALLBACK_REL

# Tile setup and input loads omitted.
ptr = tl.where(fallback, base + offsets, buckets + bucket_offsets)
matrix = tl.load(ptr, mask=valid, other=0.0).to(tl.float32)
corr = tl.load(correction + offsets, mask=valid & fallback,
               other=0.0).to(tl.float32)

# Compute this tile's contribution.
acc += tl.sum((matrix + alpha * corr) * x.to(tl.float32)[None, :], axis=1)
\end{lstlisting}
\end{minipage}

Here, \texttt{FALLBACK\_REL} is 0.05, and \texttt{NUM\_CENTERS} is the number of representatives stored for the layer.

\clearpage
\begingroup
\raggedbottom
\subsection{Complete Experimental Results}
\label{app:complete-results}

\subsubsection{Non-uniform Sparsity}
Tables~\ref{tab:benchmark-llama31}--\ref{tab:benchmark-qwen3} provide the per-model results underlying Table~\ref{tab:model-accuracy-summary}. Sparsity denotes the target MAC reduction for linear projections in the Transformer. Entries report benchmark accuracy; bold indicates the highest score among the sparse methods compared at the same sparsity. Avg is the unweighted mean over 11 benchmarks.

RoShift improves average accuracy over TEAL by 0.35 percentage points across five models at 40\% sparsity, with the gap increasing to 1.60 points at 50\% and 3.29 points at 60\%. Individual benchmark results show the same trend. Of the 55 model-benchmark combinations, RoShift outperforms TEAL in 33/55 at 40\% sparsity, increasing to 50/55 at 50\% and 54/55 at 60\%. Thus, RoShift's accuracy preservation advantage becomes both larger and more consistent at higher sparsity.

Although our log-likelihood evaluation applies sparsity to all input tokens, unlike TEAL's original setting, RoShift's accuracy advantage is also evident on generation benchmarks, where TEAL, WINA, and RoShift all use dense prompt prefill. At 50\% and 60\% sparsity, RoShift's mean gains over TEAL are 1.82 and 3.70 percentage points on generation benchmarks, respectively, compared with 1.47 and 3.06 points on log-likelihood benchmarks. Similarly, at 50\% and 60\% sparsity, RoShift's mean gains over WINA are 3.20 and 7.98 percentage points on generation benchmarks, respectively, compared with 1.66 and 4.33 points on log-likelihood benchmarks.

\medskip
\noindent\begin{minipage}{\linewidth}\centering
\captionsetup{font=tablefont,skip=5pt}
\captionof{table}{Llama-3.1-8B: complete non-uniform accuracy results. Avg is the unweighted mean of 11 benchmarks.}
\label{tab:benchmark-llama31}
\begingroup
\fontsize{7.5}{8.2}\selectfont
\setlength{\tabcolsep}{1pt}
\renewcommand{\arraystretch}{1.12}
\begin{tabular*}{\linewidth}{@{\extracolsep{\fill}}l*{12}{c}@{}}
\toprule
 & \multicolumn{7}{c}{Log-Likelihood-Based Tasks} & \multicolumn{4}{c}{Generation Tasks} & \\
\cmidrule(lr){2-8}\cmidrule(lr){9-12}
Method & MMLU & ARC-E & ARC-C & BoolQ & PIQA & \shortstack{Wino\\Grande} & CSQA & BBH & GSM8K & \shortstack{MATH-\\500} & MBPP & Avg \\
\midrule
Dense (0\%) & 65.57 & 82.49 & 58.28 & 83.12 & 81.12 & 78.22 & 70.60 & 64.52 & 48.90 & 21.60 & 48.20 & 63.87 \\
\midrule
TEAL 40\% & 62.95 & 80.22 & \textbf{56.83} & \textbf{82.51} & \textbf{80.20} & 76.87 & 68.47 & 61.91 & 46.32 & 19.40 & 44.40 & 61.83 \\
LaRoSA 40\% & 62.38 & 79.21 & 55.72 & 81.65 & 79.60 & 77.03 & \textbf{69.53} & 60.34 & 43.90 & 16.80 & \textbf{47.20} & 61.21 \\
WINA 40\% & 63.49 & \textbf{81.57} & 56.14 & 81.93 & 79.60 & 76.24 & 68.88 & 61.22 & 44.96 & \textbf{19.60} & 45.20 & 61.71 \\
RoShift 40\% & \textbf{63.70} & 80.85 & \textbf{56.83} & 82.35 & 79.43 & \textbf{77.19} & 68.71 & \textbf{62.92} & \textbf{47.23} & 18.60 & 47.00 & \textbf{62.26} \\
\midrule
TEAL 50\% & 59.98 & 78.49 & 55.46 & 81.50 & 79.05 & 74.43 & 64.13 & 58.52 & 40.94 & 15.20 & 43.60 & 59.21 \\
LaRoSA 50\% & 57.43 & 77.06 & 52.82 & 80.34 & 78.89 & 72.22 & 63.80 & 54.71 & 35.63 & 12.00 & 39.60 & 56.77 \\
WINA 50\% & 59.28 & 78.79 & 55.03 & 81.71 & 79.43 & 75.30 & 63.80 & 56.80 & 36.77 & 15.00 & 41.80 & 58.52 \\
RoShift 50\% & \textbf{62.27} & \textbf{81.02} & \textbf{57.00} & \textbf{82.32} & \textbf{79.87} & \textbf{77.51} & \textbf{69.45} & \textbf{59.75} & \textbf{44.58} & \textbf{17.80} & \textbf{44.60} & \textbf{61.47} \\
\midrule
TEAL 60\% & 52.81 & 74.20 & 49.15 & 78.10 & 76.71 & 70.40 & 56.51 & 52.14 & 30.02 & 11.00 & 36.80 & 53.44 \\
LaRoSA 60\% & 46.24 & 69.74 & 43.69 & 75.75 & 74.97 & 64.48 & 48.48 & 46.92 & 17.51 & 7.60 & 27.00 & 47.49 \\
WINA 60\% & 48.99 & 72.39 & 47.95 & 78.72 & 76.22 & 68.98 & 53.32 & 49.02 & 21.00 & 7.20 & 33.00 & 50.62 \\
RoShift 60\% & \textbf{58.88} & \textbf{78.66} & \textbf{52.47} & \textbf{81.93} & \textbf{78.62} & \textbf{73.80} & \textbf{66.83} & \textbf{56.17} & \textbf{39.80} & \textbf{15.00} & \textbf{41.80} & \textbf{58.54} \\
\bottomrule\end{tabular*}
\endgroup\end{minipage}
\par\bigskip
\noindent\begin{minipage}{\linewidth}\centering
\captionsetup{font=tablefont,skip=5pt}
\captionof{table}{Llama-2-7B: complete non-uniform accuracy results. Avg is the unweighted mean of 11 benchmarks.}
\label{tab:benchmark-llama2}
\begingroup
\fontsize{7.5}{8.2}\selectfont
\setlength{\tabcolsep}{1pt}
\renewcommand{\arraystretch}{1.12}
\begin{tabular*}{\linewidth}{@{\extracolsep{\fill}}l*{12}{c}@{}}
\toprule
 & \multicolumn{7}{c}{Log-Likelihood-Based Tasks} & \multicolumn{4}{c}{Generation Tasks} & \\
\cmidrule(lr){2-8}\cmidrule(lr){9-12}
Method & MMLU & ARC-E & ARC-C & BoolQ & PIQA & \shortstack{Wino\\Grande} & CSQA & BBH & GSM8K & \shortstack{MATH-\\500} & MBPP & Avg \\
\midrule
Dense (0\%) & 45.96 & 73.86 & 52.65 & 79.33 & 78.73 & 74.66 & 32.76 & 39.69 & 14.56 & 3.20 & 23.20 & 47.14 \\
\midrule
TEAL 40\% & 44.26 & 72.52 & 51.11 & 77.37 & 78.62 & 72.61 & 30.79 & \textbf{39.58} & \textbf{14.48} & 2.40 & 21.00 & 45.89 \\
LaRoSA 40\% & 42.28 & 72.94 & 50.17 & 77.16 & 77.09 & 71.59 & 28.83 & 38.78 & 12.51 & 3.60 & 20.00 & 44.99 \\
WINA 40\% & \textbf{44.97} & \textbf{73.86} & \textbf{52.56} & 77.65 & \textbf{78.73} & 73.95 & \textbf{31.20} & 38.15 & 13.04 & \textbf{3.80} & \textbf{22.40} & \textbf{46.39} \\
RoShift 40\% & 43.48 & 73.27 & 52.30 & \textbf{79.20} & 78.51 & \textbf{74.19} & 28.67 & 39.20 & 13.87 & 3.00 & 20.60 & 46.03 \\
\midrule
TEAL 50\% & 41.31 & 71.55 & 50.00 & 75.35 & 77.69 & 71.43 & 27.44 & 37.61 & 11.52 & \textbf{4.40} & 19.40 & 44.34 \\
LaRoSA 50\% & 38.81 & 72.22 & 47.27 & 75.84 & 76.82 & 68.67 & 27.35 & 37.26 & 10.39 & 3.80 & 15.60 & 43.09 \\
WINA 50\% & 42.88 & \textbf{72.69} & \textbf{52.39} & 76.97 & \textbf{78.73} & \textbf{73.88} & \textbf{30.06} & 37.37 & 11.68 & 4.00 & 19.80 & 45.49 \\
RoShift 50\% & \textbf{43.28} & 72.52 & 51.28 & \textbf{78.72} & 77.75 & 73.01 & 28.91 & \textbf{38.72} & \textbf{13.87} & \textbf{4.40} & \textbf{22.40} & \textbf{45.90} \\
\midrule
TEAL 60\% & 35.81 & 68.43 & 45.82 & 73.76 & 76.22 & 68.19 & 25.47 & 35.54 & 7.96 & 3.00 & 15.80 & 41.46 \\
LaRoSA 60\% & 32.35 & 67.05 & 41.81 & 72.32 & 75.03 & 68.03 & 21.87 & 35.69 & 6.37 & \textbf{4.40} & 12.60 & 39.77 \\
WINA 60\% & \textbf{38.33} & 69.53 & 47.53 & 75.44 & 76.61 & 67.88 & \textbf{26.54} & 34.08 & 6.75 & \textbf{4.40} & 16.80 & 42.17 \\
RoShift 60\% & 37.86 & \textbf{72.43} & \textbf{49.06} & \textbf{76.06} & \textbf{77.75} & \textbf{71.27} & 23.75 & \textbf{36.32} & \textbf{10.61} & 4.00 & \textbf{18.40} & \textbf{43.41} \\
\bottomrule\end{tabular*}
\endgroup\end{minipage}
\par\bigskip
\noindent\begin{minipage}{\linewidth}\centering
\captionsetup{font=tablefont,skip=5pt}
\captionof{table}{Mistral-7B: complete non-uniform accuracy results. Avg is the unweighted mean of 11 benchmarks.}
\label{tab:benchmark-mistral}
\begingroup
\fontsize{7.5}{8.2}\selectfont
\setlength{\tabcolsep}{1pt}
\renewcommand{\arraystretch}{1.12}
\begin{tabular*}{\linewidth}{@{\extracolsep{\fill}}l*{12}{c}@{}}
\toprule
 & \multicolumn{7}{c}{Log-Likelihood-Based Tasks} & \multicolumn{4}{c}{Generation Tasks} & \\
\cmidrule(lr){2-8}\cmidrule(lr){9-12}
Method & MMLU & ARC-E & ARC-C & BoolQ & PIQA & \shortstack{Wino\\Grande} & CSQA & BBH & GSM8K & \shortstack{MATH-\\500} & MBPP & Avg \\
\midrule
Dense (0\%) & 62.65 & 79.38 & 61.26 & 84.40 & 82.54 & 78.93 & 59.13 & 56.70 & 39.58 & 13.00 & 38.80 & 59.67 \\
\midrule
TEAL 40\% & 61.07 & \textbf{79.46} & 60.75 & 83.70 & 81.66 & \textbf{78.93} & 56.10 & 55.66 & 35.86 & 12.80 & \textbf{38.20} & 58.56 \\
LaRoSA 40\% & 60.78 & 79.25 & 59.13 & 83.46 & 81.72 & 78.22 & 55.28 & 55.28 & 35.18 & 11.20 & 36.60 & 57.83 \\
WINA 40\% & 60.98 & 79.34 & 61.01 & 83.91 & \textbf{82.48} & 77.11 & 56.76 & 55.68 & 36.92 & \textbf{13.20} & 37.20 & 58.60 \\
RoShift 40\% & \textbf{61.61} & 78.75 & \textbf{61.60} & \textbf{84.74} & \textbf{82.48} & 78.69 & \textbf{59.87} & \textbf{56.14} & \textbf{37.76} & 12.80 & 37.20 & \textbf{59.24} \\
\midrule
TEAL 50\% & 59.64 & 77.86 & 60.15 & 83.55 & 81.07 & 76.56 & 54.14 & 54.12 & 33.59 & \textbf{12.00} & 33.80 & 56.95 \\
LaRoSA 50\% & 57.56 & 77.90 & 56.57 & 81.68 & 80.25 & 75.69 & 48.98 & 52.39 & 29.26 & 8.60 & 33.60 & 54.77 \\
WINA 50\% & 58.92 & 77.90 & 59.90 & 82.97 & \textbf{81.56} & 76.24 & 54.95 & 52.50 & 32.30 & 10.00 & 33.60 & 56.44 \\
RoShift 50\% & \textbf{60.25} & \textbf{78.87} & \textbf{60.41} & \textbf{84.65} & 81.39 & \textbf{77.82} & \textbf{59.87} & \textbf{55.31} & \textbf{37.23} & 11.40 & \textbf{36.80} & \textbf{58.54} \\
\midrule
TEAL 60\% & 55.14 & 76.89 & 56.40 & 82.57 & 80.74 & 74.90 & 49.06 & 51.08 & 28.58 & 9.00 & 34.20 & 54.42 \\
LaRoSA 60\% & 50.78 & 73.61 & 51.71 & 80.09 & 78.89 & 72.69 & 41.20 & 47.26 & 23.96 & 8.00 & 28.20 & 50.58 \\
WINA 60\% & 54.09 & 77.19 & 55.55 & 80.76 & 79.76 & 74.03 & 47.99 & 47.14 & 22.44 & 8.00 & 29.60 & 52.41 \\
RoShift 60\% & \textbf{58.28} & \textbf{78.11} & \textbf{58.36} & \textbf{84.86} & \textbf{81.01} & \textbf{78.06} & \textbf{60.28} & \textbf{55.01} & \textbf{34.65} & \textbf{11.00} & \textbf{36.40} & \textbf{57.82} \\
\bottomrule\end{tabular*}
\endgroup\end{minipage}
\par\bigskip
\noindent\begin{minipage}{\linewidth}\centering
\captionsetup{font=tablefont,skip=5pt}
\captionof{table}{Qwen2.5-7B: complete non-uniform accuracy results. Avg is the unweighted mean of 11 benchmarks.}
\label{tab:benchmark-qwen25}
\begingroup
\fontsize{7.5}{8.2}\selectfont
\setlength{\tabcolsep}{1pt}
\renewcommand{\arraystretch}{1.12}
\begin{tabular*}{\linewidth}{@{\extracolsep{\fill}}l*{12}{c}@{}}
\toprule
 & \multicolumn{7}{c}{Log-Likelihood-Based Tasks} & \multicolumn{4}{c}{Generation Tasks} & \\
\cmidrule(lr){2-8}\cmidrule(lr){9-12}
Method & MMLU & ARC-E & ARC-C & BoolQ & PIQA & \shortstack{Wino\\Grande} & CSQA & BBH & GSM8K & \shortstack{MATH-\\500} & MBPP & Avg \\
\midrule
Dense (0\%) & 74.23 & 77.40 & 63.74 & 84.68 & 79.76 & 76.01 & 85.50 & 68.96 & 82.64 & 49.60 & 63.60 & 73.28 \\
\midrule
TEAL 40\% & 72.82 & 77.15 & 62.46 & \textbf{84.86} & 79.33 & \textbf{76.09} & 83.78 & 66.32 & \textbf{80.82} & \textbf{45.00} & \textbf{63.00} & \textbf{71.97} \\
LaRoSA 40\% & 72.67 & \textbf{78.45} & 61.01 & 83.73 & 79.38 & 74.35 & 83.37 & \textbf{66.93} & 79.30 & 41.40 & 60.80 & 71.04 \\
WINA 40\% & 72.46 & 77.57 & 61.95 & 84.68 & 78.78 & 73.95 & 83.29 & 66.72 & 79.45 & \textbf{45.00} & 60.00 & 71.26 \\
RoShift 40\% & \textbf{73.58} & 77.53 & \textbf{63.31} & 84.53 & \textbf{80.14} & 75.30 & \textbf{84.93} & \textbf{66.93} & 77.86 & 44.40 & 62.00 & 71.86 \\
\midrule
TEAL 50\% & 71.30 & 77.82 & 61.43 & \textbf{84.50} & 78.89 & 74.03 & 82.80 & 64.52 & \textbf{78.01} & 42.00 & 59.80 & 70.46 \\
LaRoSA 50\% & 69.07 & 76.89 & 59.81 & 82.20 & 78.29 & 72.14 & 79.77 & 63.40 & 74.60 & 36.80 & 56.40 & 68.13 \\
WINA 50\% & 69.72 & 77.57 & 58.19 & 83.09 & 78.84 & 72.77 & 81.82 & 63.43 & 76.42 & 39.60 & 57.00 & 68.95 \\
RoShift 50\% & \textbf{73.12} & \textbf{79.29} & \textbf{63.05} & 84.25 & \textbf{79.87} & \textbf{75.93} & \textbf{83.62} & \textbf{65.87} & 77.03 & \textbf{44.60} & \textbf{61.00} & \textbf{71.60} \\
\midrule
TEAL 60\% & 68.59 & 75.42 & 59.30 & 81.74 & 78.67 & 69.22 & 79.03 & 61.59 & 73.92 & 37.60 & 54.00 & 67.19 \\
LaRoSA 60\% & 61.95 & 72.52 & 52.56 & 78.81 & 76.33 & 70.56 & 72.65 & 57.07 & 65.20 & 25.60 & 42.00 & 61.39 \\
WINA 60\% & 65.87 & 73.23 & 54.95 & 80.83 & 78.35 & 68.82 & 77.40 & 56.54 & 66.03 & 29.20 & 46.40 & 63.42 \\
RoShift 60\% & \textbf{72.07} & \textbf{78.03} & \textbf{60.92} & \textbf{83.73} & \textbf{79.38} & \textbf{74.27} & \textbf{82.72} & \textbf{65.20} & \textbf{78.92} & \textbf{42.00} & \textbf{58.60} & \textbf{70.53} \\
\bottomrule\end{tabular*}
\endgroup\end{minipage}
\par\bigskip
\noindent\begin{minipage}{\linewidth}\centering
\captionsetup{font=tablefont,skip=5pt}
\captionof{table}{Qwen3-8B: complete non-uniform accuracy results. Avg is the unweighted mean of 11 benchmarks.}
\label{tab:benchmark-qwen3}
\begingroup
\fontsize{7.5}{8.2}\selectfont
\setlength{\tabcolsep}{1pt}
\renewcommand{\arraystretch}{1.12}
\begin{tabular*}{\linewidth}{@{\extracolsep{\fill}}l*{12}{c}@{}}
\toprule
 & \multicolumn{7}{c}{Log-Likelihood-Based Tasks} & \multicolumn{4}{c}{Generation Tasks} & \\
\cmidrule(lr){2-8}\cmidrule(lr){9-12}
Method & MMLU & ARC-E & ARC-C & BoolQ & PIQA & \shortstack{Wino\\Grande} & CSQA & BBH & GSM8K & \shortstack{MATH-\\500} & MBPP & Avg \\
\midrule
Dense (0\%) & 76.86 & 80.01 & 67.58 & 83.03 & 79.38 & 77.66 & 85.91 & 77.64 & 85.44 & 57.40 & 69.80 & 76.43 \\
\midrule
TEAL 40\% & 75.54 & 79.88 & \textbf{67.24} & 82.60 & 79.00 & \textbf{76.24} & 83.95 & 73.89 & 82.56 & \textbf{52.00} & 67.00 & 74.54 \\
LaRoSA 40\% & 75.47 & 80.85 & 66.30 & 82.26 & 79.05 & 75.69 & 83.05 & 75.36 & 81.43 & 48.00 & 66.80 & 74.02 \\
WINA 40\% & 75.30 & 78.41 & 66.47 & 81.35 & 78.89 & 75.06 & 83.13 & 75.29 & 83.02 & 50.00 & 66.80 & 73.97 \\
RoShift 40\% & \textbf{75.73} & \textbf{81.36} & \textbf{67.24} & \textbf{83.24} & \textbf{79.71} & 75.61 & \textbf{84.77} & \textbf{75.87} & \textbf{83.47} & \textbf{52.00} & \textbf{67.80} & \textbf{75.16} \\
\midrule
TEAL 50\% & 74.16 & \textbf{80.51} & 65.70 & 82.29 & 79.76 & 74.66 & 81.82 & 72.11 & 81.96 & 51.00 & 64.40 & 73.49 \\
LaRoSA 50\% & 71.52 & 80.22 & 63.82 & 79.97 & 78.02 & 72.69 & 78.38 & 71.29 & 78.62 & 42.40 & 63.00 & 70.90 \\
WINA 50\% & 72.74 & 78.62 & 64.59 & 81.25 & 77.75 & 72.69 & 80.84 & 72.17 & 80.89 & 46.80 & 63.00 & 71.94 \\
RoShift 50\% & \textbf{75.07} & 80.47 & \textbf{66.38} & \textbf{83.39} & \textbf{80.20} & \textbf{75.14} & \textbf{84.03} & \textbf{75.04} & \textbf{83.93} & \textbf{52.00} & \textbf{68.60} & \textbf{74.93} \\
\midrule
TEAL 60\% & 70.70 & 80.51 & 63.82 & 77.89 & 78.40 & 71.82 & 77.89 & 68.81 & 78.32 & 44.60 & 61.80 & 70.41 \\
LaRoSA 60\% & 60.50 & 75.13 & 56.83 & 74.19 & 74.54 & 64.48 & 65.11 & 60.19 & 62.85 & 29.00 & 41.00 & 60.35 \\
WINA 60\% & 66.44 & 77.15 & 59.30 & 77.95 & 75.79 & 69.38 & 72.73 & 64.89 & 73.54 & 37.80 & 56.20 & 66.47 \\
RoShift 60\% & \textbf{73.13} & \textbf{81.19} & \textbf{65.70} & \textbf{82.81} & \textbf{78.62} & \textbf{75.69} & \textbf{80.84} & \textbf{71.96} & \textbf{81.65} & \textbf{48.00} & \textbf{64.20} & \textbf{73.07} \\
\bottomrule\end{tabular*}
\endgroup\end{minipage}

\clearpage
\subsubsection{Uniform Sparsity}
\label{app:uniform-results}
Table~\ref{tab:uniform-accuracy-summary} summarizes accuracy under uniform sparsity, and Tables~\ref{tab:uniform-benchmark-llama31}--\ref{tab:uniform-benchmark-qwen3} provide the per-model benchmark results. All sparse results cover the same 11 benchmarks, including BBH, at 40\%, 50\%, and 60\% sparsity across five models. Bold indicates the highest score among the sparse methods compared at the same sparsity.

\medskip
\noindent\begin{minipage}{\linewidth}\centering
\captionsetup{font=tablefont,skip=5pt}
\captionof{table}{Downstream task accuracy under uniform sparsity. Each model entry reports the mean over 11 benchmarks, including BBH; Avg averages the five models. Parentheses indicate changes from Dense.}
\label{tab:uniform-accuracy-summary}
\begingroup
\fontsize{8}{9}\selectfont
\setlength{\tabcolsep}{0.6pt}
\renewcommand{\arraystretch}{1.15}
\begin{tabular*}{\linewidth}{@{\extracolsep{\fill}}l*{6}{c}@{}}
\toprule
Method / Model & Llama-3.1-8B & Llama-2-7B & Mistral-7B & Qwen2.5-7B & Qwen3-8B & Avg \\
\midrule
Dense (0\%) & 63.87 & 47.14 & 59.67 & 73.28 & 76.43 & 64.08 \\
\midrule
TEAL 40\% & 60.75(-3.13) & 45.37(-1.78) & 58.25(-1.42) & 71.52(-1.77) & 74.26(-2.17) & 62.03(-2.05) \\
\textbf{RoShift 40\%} & \textbf{62.15}\textbf{(-1.72)} & \textbf{46.12}\textbf{(-1.02)} & \textbf{59.03}\textbf{(-0.64)} & \textbf{71.60}\textbf{(-1.69)} & \textbf{74.61}\textbf{(-1.82)} & \textbf{62.70}\textbf{(-1.38)} \\
\midrule
TEAL 50\% & 57.44(-6.43) & 43.79(-3.35) & 55.78(-3.89) & 69.43(-3.85) & 72.78(-3.65) & 59.84(-4.24) \\
\textbf{RoShift 50\%} & \textbf{60.44}\textbf{(-3.43)} & \textbf{45.47}\textbf{(-1.68)} & \textbf{58.05}\textbf{(-1.62)} & \textbf{70.73}\textbf{(-2.55)} & \textbf{74.19}\textbf{(-2.23)} & \textbf{61.78}\textbf{(-2.30)} \\
\midrule
TEAL 60\% & 50.48(-13.39) & 39.82(-7.33) & 51.42(-8.25) & 64.13(-9.15) & 69.13(-7.29) & 55.00(-9.08) \\
\textbf{RoShift 60\%} & \textbf{57.06}\textbf{(-6.81)} & \textbf{43.31}\textbf{(-3.83)} & \textbf{56.33}\textbf{(-3.34)} & \textbf{67.37}\textbf{(-5.91)} & \textbf{71.40}\textbf{(-5.03)} & \textbf{59.09}\textbf{(-4.99)} \\
\bottomrule\end{tabular*}
\endgroup\end{minipage}

\par\bigskip
\noindent\begin{minipage}{\linewidth}\centering
\captionsetup{font=tablefont,skip=5pt}
\captionof{table}{Llama-3.1-8B: complete uniform accuracy results. Avg is the unweighted mean of 11 benchmarks, including BBH.}
\label{tab:uniform-benchmark-llama31}
\begingroup
\fontsize{7.5}{8.2}\selectfont
\setlength{\tabcolsep}{1pt}
\renewcommand{\arraystretch}{1.12}
\begin{tabular*}{\linewidth}{@{\extracolsep{\fill}}l*{12}{c}@{}}
\toprule
 & \multicolumn{7}{c}{Log-Likelihood-Based Tasks} & \multicolumn{4}{c}{Generation Tasks} & \\
\cmidrule(lr){2-8}\cmidrule(lr){9-12}
Method & MMLU & ARC-E & ARC-C & BoolQ & PIQA & \shortstack{Wino\\Grande} & CSQA & BBH & GSM8K & \shortstack{MATH-\\500} & MBPP & Avg \\
\midrule
Dense (0\%) & 65.57 & 82.49 & 58.28 & 83.12 & 81.12 & 78.22 & 70.60 & 64.52 & 48.90 & 21.60 & 48.20 & 63.87 \\
\midrule
TEAL 40\% & 62.33 & 79.17 & 56.23 & \textbf{82.20} & 79.54 & 75.69 & 65.93 & 60.41 & 44.12 & 17.80 & 44.80 & 60.75 \\
RoShift 40\% & \textbf{63.81} & \textbf{79.92} & \textbf{56.91} & 81.62 & \textbf{79.71} & \textbf{78.37} & \textbf{68.39} & \textbf{62.28} & \textbf{46.47} & \textbf{19.20} & \textbf{47.00} & \textbf{62.15} \\
\midrule
TEAL 50\% & 57.52 & 78.24 & 52.90 & 79.54 & 77.69 & 72.14 & 61.51 & 57.47 & 39.42 & 15.20 & 40.20 & 57.44 \\
RoShift 50\% & \textbf{61.63} & \textbf{79.34} & \textbf{56.48} & \textbf{80.80} & \textbf{79.11} & \textbf{75.37} & \textbf{65.85} & \textbf{59.98} & \textbf{44.88} & \textbf{17.20} & \textbf{44.20} & \textbf{60.44} \\
\midrule
TEAL 60\% & 48.63 & 73.15 & 43.86 & 75.50 & 76.06 & 65.75 & 50.45 & 50.64 & 26.69 & 10.40 & 34.20 & 50.48 \\
RoShift 60\% & \textbf{56.52} & \textbf{76.81} & \textbf{52.90} & \textbf{79.72} & \textbf{77.97} & \textbf{72.61} & \textbf{61.59} & \textbf{55.58} & \textbf{36.54} & \textbf{16.60} & \textbf{40.80} & \textbf{57.06} \\
\bottomrule\end{tabular*}
\endgroup\end{minipage}
\par\bigskip
\noindent\begin{minipage}{\linewidth}\centering
\captionsetup{font=tablefont,skip=5pt}
\captionof{table}{Llama-2-7B: complete uniform accuracy results. Avg is the unweighted mean of 11 benchmarks, including BBH.}
\label{tab:uniform-benchmark-llama2}
\begingroup
\fontsize{7.5}{8.2}\selectfont
\setlength{\tabcolsep}{1pt}
\renewcommand{\arraystretch}{1.12}
\begin{tabular*}{\linewidth}{@{\extracolsep{\fill}}l*{12}{c}@{}}
\toprule
 & \multicolumn{7}{c}{Log-Likelihood-Based Tasks} & \multicolumn{4}{c}{Generation Tasks} & \\
\cmidrule(lr){2-8}\cmidrule(lr){9-12}
Method & MMLU & ARC-E & ARC-C & BoolQ & PIQA & \shortstack{Wino\\Grande} & CSQA & BBH & GSM8K & \shortstack{MATH-\\500} & MBPP & Avg \\
\midrule
Dense (0\%) & 45.96 & 73.86 & 52.65 & 79.33 & 78.73 & 74.66 & 32.76 & 39.69 & 14.56 & 3.20 & 23.20 & 47.14 \\
\midrule
TEAL 40\% & 43.53 & \textbf{72.18} & 50.68 & 76.79 & 77.53 & \textbf{71.82} & 31.29 & 38.67 & 12.13 & \textbf{4.20} & 20.20 & 45.37 \\
RoShift 40\% & \textbf{45.08} & 72.05 & \textbf{51.45} & \textbf{78.56} & \textbf{78.62} & 71.59 & \textbf{33.01} & \textbf{39.04} & \textbf{12.96} & 3.40 & \textbf{21.60} & \textbf{46.12} \\
\midrule
TEAL 50\% & 40.78 & 71.63 & 49.32 & 74.50 & 77.20 & 69.93 & 27.35 & 36.51 & \textbf{12.28} & \textbf{4.40} & 17.80 & 43.79 \\
RoShift 50\% & \textbf{42.84} & \textbf{72.22} & \textbf{51.11} & \textbf{76.12} & \textbf{78.51} & \textbf{72.69} & \textbf{30.63} & \textbf{39.21} & 12.21 & 4.20 & \textbf{20.40} & \textbf{45.47} \\
\midrule
TEAL 60\% & 33.49 & 67.72 & 42.15 & 71.04 & 75.14 & 67.09 & 23.26 & 33.88 & 6.44 & \textbf{4.20} & 13.60 & 39.82 \\
RoShift 60\% & \textbf{38.22} & \textbf{70.58} & \textbf{47.18} & \textbf{75.17} & \textbf{76.33} & \textbf{70.96} & \textbf{27.76} & \textbf{37.00} & \textbf{11.22} & 3.60 & \textbf{18.40} & \textbf{43.31} \\
\bottomrule\end{tabular*}
\endgroup\end{minipage}
\par\bigskip
\noindent\begin{minipage}{\linewidth}\centering
\captionsetup{font=tablefont,skip=5pt}
\captionof{table}{Mistral-7B: complete uniform accuracy results. Avg is the unweighted mean of 11 benchmarks, including BBH.}
\label{tab:uniform-benchmark-mistral}
\begingroup
\fontsize{7.5}{8.2}\selectfont
\setlength{\tabcolsep}{1pt}
\renewcommand{\arraystretch}{1.12}
\begin{tabular*}{\linewidth}{@{\extracolsep{\fill}}l*{12}{c}@{}}
\toprule
 & \multicolumn{7}{c}{Log-Likelihood-Based Tasks} & \multicolumn{4}{c}{Generation Tasks} & \\
\cmidrule(lr){2-8}\cmidrule(lr){9-12}
Method & MMLU & ARC-E & ARC-C & BoolQ & PIQA & \shortstack{Wino\\Grande} & CSQA & BBH & GSM8K & \shortstack{MATH-\\500} & MBPP & Avg \\
\midrule
Dense (0\%) & 62.65 & 79.38 & 61.26 & 84.40 & 82.54 & 78.93 & 59.13 & 56.70 & 39.58 & 13.00 & 38.80 & 59.67 \\
\midrule
TEAL 40\% & 60.89 & \textbf{79.04} & 60.49 & 83.82 & 81.77 & 76.95 & 57.00 & 55.44 & 34.72 & 12.40 & \textbf{38.20} & 58.25 \\
RoShift 40\% & \textbf{61.18} & 78.96 & \textbf{60.84} & \textbf{84.46} & \textbf{81.94} & \textbf{77.98} & \textbf{58.72} & \textbf{55.86} & \textbf{38.36} & \textbf{13.40} & 37.60 & \textbf{59.03} \\
\midrule
TEAL 50\% & 58.70 & 77.15 & 57.94 & 83.09 & 81.23 & 75.06 & 53.07 & 53.22 & 30.71 & 9.60 & 33.80 & 55.78 \\
RoShift 50\% & \textbf{60.06} & \textbf{78.28} & \textbf{58.11} & \textbf{84.43} & \textbf{81.77} & \textbf{76.48} & \textbf{59.71} & \textbf{55.40} & \textbf{34.95} & \textbf{11.60} & \textbf{37.80} & \textbf{58.05} \\
\midrule
TEAL 60\% & 52.40 & 73.86 & 52.13 & 81.01 & 79.38 & 71.03 & 47.01 & 49.32 & 23.65 & 6.40 & 29.40 & 51.42 \\
RoShift 60\% & \textbf{57.16} & \textbf{77.74} & \textbf{56.57} & \textbf{84.10} & \textbf{80.25} & \textbf{76.01} & \textbf{56.67} & \textbf{52.85} & \textbf{33.06} & \textbf{9.40} & \textbf{35.80} & \textbf{56.33} \\
\bottomrule\end{tabular*}
\endgroup\end{minipage}
\par\bigskip
\noindent\begin{minipage}{\linewidth}\centering
\captionsetup{font=tablefont,skip=5pt}
\captionof{table}{Qwen2.5-7B: complete uniform accuracy results. Avg is the unweighted mean of 11 benchmarks, including BBH.}
\label{tab:uniform-benchmark-qwen25}
\begingroup
\fontsize{7.5}{8.2}\selectfont
\setlength{\tabcolsep}{1pt}
\renewcommand{\arraystretch}{1.12}
\begin{tabular*}{\linewidth}{@{\extracolsep{\fill}}l*{12}{c}@{}}
\toprule
 & \multicolumn{7}{c}{Log-Likelihood-Based Tasks} & \multicolumn{4}{c}{Generation Tasks} & \\
\cmidrule(lr){2-8}\cmidrule(lr){9-12}
Method & MMLU & ARC-E & ARC-C & BoolQ & PIQA & \shortstack{Wino\\Grande} & CSQA & BBH & GSM8K & \shortstack{MATH-\\500} & MBPP & Avg \\
\midrule
Dense (0\%) & 74.23 & 77.40 & 63.74 & 84.68 & 79.76 & 76.01 & 85.50 & 68.96 & 82.64 & 49.60 & 63.60 & 73.28 \\
\midrule
TEAL 40\% & 72.48 & 76.89 & 61.95 & 83.88 & 78.94 & \textbf{75.06} & 83.29 & \textbf{66.87} & \textbf{80.52} & \textbf{46.20} & 60.60 & 71.52 \\
RoShift 40\% & \textbf{73.16} & \textbf{77.82} & \textbf{62.37} & \textbf{84.89} & \textbf{79.33} & \textbf{75.06} & \textbf{84.60} & 66.06 & 77.86 & 44.20 & \textbf{62.20} & \textbf{71.60} \\
\midrule
TEAL 50\% & 70.03 & 77.19 & 60.24 & 83.67 & \textbf{79.33} & \textbf{74.11} & 80.18 & 64.00 & 76.80 & 39.60 & 58.60 & 69.43 \\
RoShift 50\% & \textbf{71.28} & \textbf{78.16} & \textbf{60.75} & \textbf{84.34} & 79.05 & 73.40 & \textbf{84.28} & \textbf{65.20} & \textbf{77.56} & \textbf{43.20} & \textbf{60.80} & \textbf{70.73} \\
\midrule
TEAL 60\% & 64.46 & 74.66 & 54.35 & 80.31 & 77.20 & 68.59 & 74.45 & 56.77 & 70.89 & 30.40 & 53.40 & 64.13 \\
RoShift 60\% & \textbf{68.41} & \textbf{76.89} & \textbf{58.11} & \textbf{83.09} & \textbf{77.37} & \textbf{72.30} & \textbf{79.69} & \textbf{60.56} & \textbf{74.91} & \textbf{35.20} & \textbf{54.60} & \textbf{67.37} \\
\bottomrule\end{tabular*}
\endgroup\end{minipage}
\par\bigskip
\noindent\begin{minipage}{\linewidth}\centering
\captionsetup{font=tablefont,skip=5pt}
\captionof{table}{Qwen3-8B: complete uniform accuracy results. Avg is the unweighted mean of 11 benchmarks, including BBH.}
\label{tab:uniform-benchmark-qwen3}
\begingroup
\fontsize{7.5}{8.2}\selectfont
\setlength{\tabcolsep}{1pt}
\renewcommand{\arraystretch}{1.12}
\begin{tabular*}{\linewidth}{@{\extracolsep{\fill}}l*{12}{c}@{}}
\toprule
 & \multicolumn{7}{c}{Log-Likelihood-Based Tasks} & \multicolumn{4}{c}{Generation Tasks} & \\
\cmidrule(lr){2-8}\cmidrule(lr){9-12}
Method & MMLU & ARC-E & ARC-C & BoolQ & PIQA & \shortstack{Wino\\Grande} & CSQA & BBH & GSM8K & \shortstack{MATH-\\500} & MBPP & Avg \\
\midrule
Dense (0\%) & 76.86 & 80.01 & 67.58 & 83.03 & 79.38 & 77.66 & 85.91 & 77.64 & 85.44 & 57.40 & 69.80 & 76.43 \\
\midrule
TEAL 40\% & 75.54 & \textbf{81.61} & 65.61 & 82.60 & \textbf{79.05} & \textbf{76.09} & 83.05 & 73.31 & \textbf{81.58} & \textbf{51.40} & 67.00 & 74.26 \\
RoShift 40\% & \textbf{75.79} & 80.39 & \textbf{66.81} & \textbf{82.69} & \textbf{79.05} & \textbf{76.09} & \textbf{84.93} & \textbf{76.09} & 80.82 & 49.80 & \textbf{68.20} & \textbf{74.61} \\
\midrule
TEAL 50\% & 73.81 & 80.39 & 64.16 & 81.35 & \textbf{79.11} & 73.56 & 81.08 & 72.60 & 80.74 & 49.40 & 64.40 & 72.78 \\
RoShift 50\% & \textbf{74.65} & \textbf{81.73} & \textbf{66.30} & \textbf{83.06} & 78.94 & \textbf{75.53} & \textbf{82.72} & \textbf{74.03} & \textbf{81.96} & \textbf{49.60} & \textbf{67.60} & \textbf{74.19} \\
\midrule
TEAL 60\% & 68.94 & 79.46 & 60.92 & 75.54 & 76.77 & 70.17 & 78.05 & 68.71 & 76.12 & 43.60 & 62.20 & 69.13 \\
RoShift 60\% & \textbf{71.12} & \textbf{80.35} & \textbf{62.37} & \textbf{82.66} & \textbf{77.53} & \textbf{72.77} & \textbf{79.44} & \textbf{70.65} & \textbf{80.14} & \textbf{45.20} & \textbf{63.20} & \textbf{71.40} \\
\bottomrule\end{tabular*}
\endgroup\end{minipage}

\clearpage
\Needspace{19\baselineskip}
\subsubsection{Uniform vs. Non-uniform Sparsity}
\label{app:roshift-allocation-comparison}
Table~\ref{tab:roshift-allocation-comparison} compares uniform and non-uniform sparsity for RoShift across the same five models and 11 benchmarks. At 40\%, 50\%, and 60\% sparsity, non-uniform allocation reduces the mean accuracy loss relative to Dense from 1.38, 2.30, and 4.99 to 1.17, 1.59, and 3.40 percentage points, respectively.

\medskip
\noindent\begin{minipage}{\linewidth}\centering
\captionsetup{font=tablefont,skip=5pt}
\captionof{table}{RoShift with uniform and non-uniform sparsity. Each model entry averages 11 benchmarks; Avg averages all 55 model-benchmark combinations. Parentheses show differences from Dense in percentage points. Bold indicates the higher accuracy within each sparsity pair.}
\label{tab:roshift-allocation-comparison}
\begingroup
\fontsize{7.5}{8.2}\selectfont
\setlength{\tabcolsep}{0.6pt}
\renewcommand{\arraystretch}{1.15}
\begin{tabular*}{\linewidth}{@{\extracolsep{\fill}}l*{6}{c}@{}}
\toprule
Method / Model & Llama-3.1-8B & Llama-2-7B & Mistral-7B & Qwen2.5-7B & Qwen3-8B & Avg \\
\midrule
Dense (0\%) & 63.87 & 47.14 & 59.67 & 73.28 & 76.43 & 64.08 \\
\midrule
Uniform 40\% & 62.15(-1.72) & \textbf{46.12(-1.02)} & 59.03(-0.64) & 71.60(-1.69) & 74.61(-1.82) & 62.70(-1.38) \\
Non-uniform 40\% & \textbf{62.26(-1.62)} & 46.03(-1.12) & \textbf{59.24(-0.43)} & \textbf{71.86(-1.42)} & \textbf{75.16(-1.27)} & \textbf{62.91(-1.17)} \\
\midrule
Uniform 50\% & 60.44(-3.43) & 45.47(-1.68) & 58.05(-1.62) & 70.73(-2.55) & 74.19(-2.23) & 61.78(-2.30) \\
Non-uniform 50\% & \textbf{61.47(-2.40)} & \textbf{45.90(-1.25)} & \textbf{58.54(-1.13)} & \textbf{71.60(-1.68)} & \textbf{74.93(-1.50)} & \textbf{62.49(-1.59)} \\
\midrule
Uniform 60\% & 57.06(-6.81) & 43.31(-3.83) & 56.33(-3.34) & 67.37(-5.91) & 71.40(-5.03) & 59.09(-4.99) \\
Non-uniform 60\% & \textbf{58.54(-5.33)} & \textbf{43.41(-3.73)} & \textbf{57.82(-1.85)} & \textbf{70.53(-2.75)} & \textbf{73.07(-3.36)} & \textbf{60.68(-3.40)} \\
\bottomrule
\end{tabular*}
\endgroup\end{minipage}

\subsubsection{LaRoSA with Original Allocation}
Table~\ref{tab:larosa-original-allocation} reports LaRoSA results under its original allocation policy, which measures sparsity by the number of activation dimensions, separately from the MAC-sparsity-based allocation search used in the main paper. Both configurations use FP16 runtime rotations. BBH results are not included in this table. Note that in RoShift, the layer-wise rotations are absorbed into the model weights offline, so no runtime rotation is required; only the transition matrix \(Q_l^\top Q_{l+1}\) is applied online in FP16. Averaged over five models and ten benchmarks excluding BBH, RoShift still outperforms LaRoSA with its original allocation policy by 0.81, 2.56, and 4.59 percentage points at nominal sparsities of 40\%, 50\%, and 60\%, respectively.

\noindent\begin{minipage}{\linewidth}\centering
\captionsetup{font=tablefont,skip=5pt}
\captionof{table}{LaRoSA with its original activation-count allocation (FP16 rotation). Target sparsity is the requested activation sparsity; MAC sparsity is the recipe-implied MAC sparsity estimate (\%). Avg averages the 10 available benchmarks, excluding BBH.}
\label{tab:larosa-original-allocation}
\begingroup
\fontsize{7.5}{8.2}\selectfont
\setlength{\tabcolsep}{1.2pt}
\renewcommand{\arraystretch}{1.10}
\begin{tabular*}{\linewidth}{@{\extracolsep{\fill}}lcc*{11}{c}@{}}
\toprule
Method & \shortstack{Target\\sparsity (\%)} & \shortstack{MAC\\sparsity (\%)} & MMLU & ARC-E & ARC-C & BoolQ & PIQA & \shortstack{Wino\\Grande} & CSQA & GSM8K & \shortstack{MATH-\\500} & MBPP & Avg \\
\midrule
\multicolumn{14}{l}{\textbf{Llama-3.1-8B}} \\
Dense & 0 & 0.00 & 65.57 & 82.49 & 58.28 & 83.12 & 81.12 & 78.22 & 70.60 & 48.90 & 21.60 & 48.20 & 63.81 \\
LaRoSA & 40 & 34.47 & 63.18 & 80.85 & 55.89 & 81.62 & 80.20 & 77.19 & 69.78 & 45.72 & 17.00 & 45.00 & 61.64 \\
LaRoSA & 50 & 45.59 & 59.42 & 77.19 & 54.44 & 79.76 & 79.16 & 72.85 & 66.01 & 38.89 & 15.00 & 42.60 & 58.53 \\
LaRoSA & 60 & 54.71 & 51.35 & 72.98 & 47.44 & 75.54 & 77.37 & 67.72 & 56.59 & 26.38 & 9.40 & 31.40 & 51.62 \\
\midrule
\multicolumn{14}{l}{\textbf{Llama-2-7B}} \\
Dense & 0 & 0.00 & 45.96 & 73.86 & 52.65 & 79.33 & 78.73 & 74.66 & 32.76 & 14.56 & 3.20 & 23.20 & 47.89 \\
LaRoSA & 40 & 37.77 & 42.35 & 73.11 & 50.09 & 76.79 & 78.18 & 71.67 & 30.47 & 12.81 & 4.00 & 21.00 & 46.05 \\
LaRoSA & 50 & 47.21 & 40.56 & 72.26 & 48.21 & 76.94 & 76.55 & 70.24 & 27.35 & 11.22 & 3.20 & 18.60 & 44.51 \\
LaRoSA & 60 & 56.66 & 33.21 & 69.15 & 46.25 & 73.18 & 75.84 & 66.77 & 22.93 & 8.95 & 3.80 & 12.60 & 41.27 \\
\midrule
\multicolumn{14}{l}{\textbf{Mistral-7B}} \\
Dense & 0 & 0.00 & 62.65 & 79.38 & 61.26 & 84.40 & 82.54 & 78.93 & 59.13 & 39.58 & 13.00 & 38.80 & 59.97 \\
LaRoSA & 40 & 35.24 & 60.98 & 79.04 & 59.13 & 83.55 & 81.28 & 77.66 & 54.46 & 35.48 & 12.40 & 35.20 & 57.92 \\
LaRoSA & 50 & 44.04 & 58.11 & 77.65 & 57.76 & 82.29 & 81.12 & 75.30 & 48.32 & 33.28 & 11.00 & 32.40 & 55.72 \\
LaRoSA & 60 & 54.71 & 52.56 & 75.63 & 55.12 & 80.34 & 79.49 & 73.80 & 40.87 & 27.60 & 7.40 & 30.20 & 52.30 \\
\midrule
\multicolumn{14}{l}{\textbf{Qwen2.5-7B}} \\
Dense & 0 & 0.00 & 74.23 & 77.40 & 63.74 & 84.68 & 79.76 & 76.01 & 85.50 & 82.64 & 49.60 & 63.60 & 73.72 \\
LaRoSA & 40 & 34.91 & 73.02 & 78.11 & 62.37 & 83.64 & 79.71 & 74.90 & 83.95 & 80.44 & 46.00 & 60.60 & 72.27 \\
LaRoSA & 50 & 42.22 & 71.62 & 76.68 & 60.15 & 81.80 & 78.51 & 74.03 & 82.72 & 76.57 & 40.60 & 60.40 & 70.31 \\
LaRoSA & 60 & 51.23 & 68.63 & 76.26 & 59.39 & 79.69 & 78.29 & 73.40 & 78.13 & 74.15 & 36.20 & 57.20 & 68.13 \\
\midrule
\multicolumn{14}{l}{\textbf{Qwen3-8B}} \\
Dense & 0 & 0.00 & 76.86 & 80.01 & 67.58 & 83.03 & 79.38 & 77.66 & 85.91 & 85.44 & 57.40 & 69.80 & 76.31 \\
LaRoSA & 40 & 35.49 & 75.49 & 80.18 & 66.47 & 82.08 & 78.73 & 74.59 & 83.78 & 81.65 & 51.60 & 65.20 & 73.98 \\
LaRoSA & 50 & 43.49 & 73.32 & 80.81 & 64.68 & 80.15 & 78.62 & 74.11 & 80.43 & 79.30 & 50.00 & 61.80 & 72.32 \\
LaRoSA & 60 & 52.97 & 69.25 & 77.40 & 62.80 & 77.77 & 78.18 & 71.27 & 73.96 & 74.00 & 45.60 & 59.80 & 69.00 \\
\bottomrule\end{tabular*}
\endgroup
\end{minipage}

\Needspace{12\baselineskip}
\subsubsection{Accuracy Impact of B128 Selection and Approximate Correction}
\label{app:blockwise-correction-accuracy}

To assess the impact of B128 Blockwise Top-K and approximate correction on benchmark accuracy, we compared RoShift's existing configuration using both techniques with a configuration using Global Top-K and separate exact correction. We evaluated both uniform and non-uniform allocation at 60\% target sparsity, keeping the sparsity allocation, rotation matrices, shift values, weights, and LM-head policy fixed within each comparison.

Tables~\ref{tab:blockwise-correction-llama31}--\ref{tab:blockwise-correction-qwen3} report results on 11 benchmarks across five models. Across the 10 model--allocation settings, the absolute difference in benchmark-average accuracy between the two configurations ranged from 0.10 to 0.55 percentage points, averaging 0.26 percentage points. This indicates that the combined use of B128 selection and approximate correction has a limited impact on average accuracy under the evaluated conditions.

\medskip
\noindent\begin{minipage}{\linewidth}\centering
\captionsetup{font=tablefont,skip=5pt}
\captionof{table}{Llama-3.1-8B: accuracy at 60\% target sparsity. Approx. uses B128 Blockwise Top-K and approximate correction fusion; Exact uses Global Top-K and separate exact correction. Values are accuracy (\%). Avg is the unweighted mean over 11 benchmarks. Bold indicates the higher accuracy within each allocation; ties at the displayed precision are both bold.}
\label{tab:blockwise-correction-llama31}
\begingroup
\fontsize{7.5}{8.2}\selectfont
\setlength{\tabcolsep}{1pt}
\renewcommand{\arraystretch}{1.12}
\begin{tabular*}{\linewidth}{@{\extracolsep{\fill}}ll*{12}{c}@{}}
\toprule
 & & \multicolumn{7}{c}{Log-Likelihood-Based Tasks} & \multicolumn{4}{c}{Generation Tasks} & \\
\cmidrule(lr){3-9}\cmidrule(lr){10-13}
Sparsity & \shortstack{RoShift\\Config.} & MMLU & ARC-E & ARC-C & BoolQ & PIQA & \shortstack{Wino\\Grande} & CSQA & BBH & GSM8K & \shortstack{MATH-\\500} & MBPP & Avg \\
\midrule
0\% & Dense & 65.57 & 82.49 & 58.28 & 83.12 & 81.12 & 78.22 & 70.60 & 64.52 & 48.90 & 21.60 & 48.20 & 63.87 \\
\midrule
\multirow{2}{*}{\shortstack[l]{Uniform\\60\%}} & Approx. & 56.52 & 76.81 & 52.90 & 79.72 & 77.97 & \textbf{72.61} & 61.59 & \textbf{55.58} & 36.54 & \textbf{16.60} & 40.80 & 57.06 \\
 & Exact & \textbf{56.75} & \textbf{77.53} & \textbf{53.84} & \textbf{79.94} & \textbf{78.02} & 72.53 & \textbf{65.52} & \textbf{55.58} & \textbf{37.68} & 12.00 & \textbf{42.20} & \textbf{57.42} \\
\midrule
\multirow{2}{*}{\shortstack[l]{Non-uniform\\60\%}} & Approx. & \textbf{58.88} & 78.66 & 52.47 & \textbf{81.93} & 78.62 & \textbf{73.80} & 66.83 & 56.17 & \textbf{39.80} & 15.00 & 41.80 & 58.54 \\
 & Exact & 58.59 & \textbf{79.21} & \textbf{52.82} & 80.86 & \textbf{79.11} & 72.93 & \textbf{67.40} & \textbf{57.16} & \textbf{39.80} & \textbf{15.20} & \textbf{43.40} & \textbf{58.77} \\
\bottomrule\end{tabular*}
\endgroup\end{minipage}
\par\bigskip
\noindent\begin{minipage}{\linewidth}\centering
\captionsetup{font=tablefont,skip=5pt}
\captionof{table}{Llama-2-7B: accuracy at 60\% target sparsity. Notation follows Table~\ref{tab:blockwise-correction-llama31}.}
\label{tab:blockwise-correction-llama2}
\begingroup
\fontsize{7.5}{8.2}\selectfont
\setlength{\tabcolsep}{1pt}
\renewcommand{\arraystretch}{1.12}
\begin{tabular*}{\linewidth}{@{\extracolsep{\fill}}ll*{12}{c}@{}}
\toprule
 & & \multicolumn{7}{c}{Log-Likelihood-Based Tasks} & \multicolumn{4}{c}{Generation Tasks} & \\
\cmidrule(lr){3-9}\cmidrule(lr){10-13}
Sparsity & \shortstack{RoShift\\Config.} & MMLU & ARC-E & ARC-C & BoolQ & PIQA & \shortstack{Wino\\Grande} & CSQA & BBH & GSM8K & \shortstack{MATH-\\500} & MBPP & Avg \\
\midrule
0\% & Dense & 45.96 & 73.86 & 52.65 & 79.33 & 78.73 & 74.66 & 32.76 & 39.69 & 14.56 & 3.20 & 23.20 & 47.14 \\
\midrule
\multirow{2}{*}{\shortstack[l]{Uniform\\60\%}} & Approx. & 38.22 & 70.58 & 47.18 & 75.17 & 76.33 & 70.96 & \textbf{27.76} & 37.00 & 11.22 & 3.60 & \textbf{18.40} & 43.31 \\
 & Exact & \textbf{38.95} & \textbf{71.09} & \textbf{49.15} & \textbf{76.33} & \textbf{77.04} & \textbf{71.03} & 26.78 & \textbf{37.23} & \textbf{11.83} & \textbf{4.60} & 17.60 & \textbf{43.78} \\
\midrule
\multirow{2}{*}{\shortstack[l]{Non-uniform\\60\%}} & Approx. & 37.86 & 72.43 & \textbf{49.06} & 76.06 & 77.75 & 71.27 & 23.75 & 36.32 & 10.61 & 4.00 & 18.40 & 43.41 \\
 & Exact & \textbf{38.21} & \textbf{72.56} & \textbf{49.06} & \textbf{76.42} & \textbf{77.80} & \textbf{71.82} & \textbf{25.31} & \textbf{36.35} & \textbf{11.07} & \textbf{4.20} & \textbf{20.80} & \textbf{43.96} \\
\bottomrule\end{tabular*}
\endgroup\end{minipage}
\par\bigskip
\noindent\begin{minipage}{\linewidth}\centering
\captionsetup{font=tablefont,skip=5pt}
\captionof{table}{Mistral-7B: accuracy at 60\% target sparsity. Notation follows Table~\ref{tab:blockwise-correction-llama31}.}
\label{tab:blockwise-correction-mistral}
\begingroup
\fontsize{7.5}{8.2}\selectfont
\setlength{\tabcolsep}{1pt}
\renewcommand{\arraystretch}{1.12}
\begin{tabular*}{\linewidth}{@{\extracolsep{\fill}}ll*{12}{c}@{}}
\toprule
 & & \multicolumn{7}{c}{Log-Likelihood-Based Tasks} & \multicolumn{4}{c}{Generation Tasks} & \\
\cmidrule(lr){3-9}\cmidrule(lr){10-13}
Sparsity & \shortstack{RoShift\\Config.} & MMLU & ARC-E & ARC-C & BoolQ & PIQA & \shortstack{Wino\\Grande} & CSQA & BBH & GSM8K & \shortstack{MATH-\\500} & MBPP & Avg \\
\midrule
0\% & Dense & 62.65 & 79.38 & 61.26 & 84.40 & 82.54 & 78.93 & 59.13 & 56.70 & 39.58 & 13.00 & 38.80 & 59.67 \\
\midrule
\multirow{2}{*}{\shortstack[l]{Uniform\\60\%}} & Approx. & 57.16 & \textbf{77.74} & \textbf{56.57} & \textbf{84.10} & 80.25 & 76.01 & 56.67 & 52.85 & \textbf{33.06} & 9.40 & \textbf{35.80} & \textbf{56.33} \\
 & Exact & \textbf{58.27} & 77.31 & 56.48 & 83.61 & \textbf{80.47} & \textbf{76.32} & \textbf{57.74} & \textbf{53.02} & 31.84 & \textbf{9.80} & 33.60 & 56.22 \\
\midrule
\multirow{2}{*}{\shortstack[l]{Non-uniform\\60\%}} & Approx. & 58.28 & 78.11 & 58.36 & \textbf{84.86} & 81.01 & \textbf{78.06} & \textbf{60.28} & \textbf{55.01} & 34.65 & 11.00 & 36.40 & 57.82 \\
 & Exact & \textbf{58.60} & \textbf{78.20} & \textbf{59.04} & 84.56 & \textbf{81.07} & 77.82 & 59.79 & 54.88 & \textbf{34.72} & \textbf{13.40} & \textbf{36.80} & \textbf{58.08} \\
\bottomrule\end{tabular*}
\endgroup\end{minipage}
\par\bigskip
\noindent\begin{minipage}{\linewidth}\centering
\captionsetup{font=tablefont,skip=5pt}
\captionof{table}{Qwen2.5-7B: accuracy at 60\% target sparsity. Notation follows Table~\ref{tab:blockwise-correction-llama31}.}
\label{tab:blockwise-correction-qwen25}
\begingroup
\fontsize{7.5}{8.2}\selectfont
\setlength{\tabcolsep}{1pt}
\renewcommand{\arraystretch}{1.12}
\begin{tabular*}{\linewidth}{@{\extracolsep{\fill}}ll*{12}{c}@{}}
\toprule
 & & \multicolumn{7}{c}{Log-Likelihood-Based Tasks} & \multicolumn{4}{c}{Generation Tasks} & \\
\cmidrule(lr){3-9}\cmidrule(lr){10-13}
Sparsity & \shortstack{RoShift\\Config.} & MMLU & ARC-E & ARC-C & BoolQ & PIQA & \shortstack{Wino\\Grande} & CSQA & BBH & GSM8K & \shortstack{MATH-\\500} & MBPP & Avg \\
\midrule
0\% & Dense & 74.23 & 77.40 & 63.74 & 84.68 & 79.76 & 76.01 & 85.50 & 68.96 & 82.64 & 49.60 & 63.60 & 73.28 \\
\midrule
\multirow{2}{*}{\shortstack[l]{Uniform\\60\%}} & Approx. & \textbf{68.41} & \textbf{76.89} & \textbf{58.11} & 83.09 & 77.37 & 72.30 & \textbf{79.69} & 60.56 & \textbf{74.91} & 35.20 & 54.60 & 67.37 \\
 & Exact & 68.09 & 76.35 & 57.51 & \textbf{83.79} & \textbf{78.07} & \textbf{72.53} & 79.36 & \textbf{61.19} & 74.37 & \textbf{35.80} & \textbf{56.20} & \textbf{67.57} \\
\midrule
\multirow{2}{*}{\shortstack[l]{Non-uniform\\60\%}} & Approx. & \textbf{72.07} & \textbf{78.03} & 60.92 & 83.73 & 79.38 & \textbf{74.27} & 82.72 & \textbf{65.20} & \textbf{78.92} & \textbf{42.00} & 58.60 & 70.53 \\
 & Exact & 71.96 & 77.99 & \textbf{61.95} & \textbf{84.07} & \textbf{79.54} & 73.56 & \textbf{83.37} & 64.97 & 78.70 & 41.20 & \textbf{60.00} & \textbf{70.66} \\
\bottomrule\end{tabular*}
\endgroup\end{minipage}
\par\bigskip
\noindent\begin{minipage}{\linewidth}\centering
\captionsetup{font=tablefont,skip=5pt}
\captionof{table}{Qwen3-8B-Base: accuracy at 60\% target sparsity. Notation follows Table~\ref{tab:blockwise-correction-llama31}.}
\label{tab:blockwise-correction-qwen3}
\begingroup
\fontsize{7.5}{8.2}\selectfont
\setlength{\tabcolsep}{1pt}
\renewcommand{\arraystretch}{1.12}
\begin{tabular*}{\linewidth}{@{\extracolsep{\fill}}ll*{12}{c}@{}}
\toprule
 & & \multicolumn{7}{c}{Log-Likelihood-Based Tasks} & \multicolumn{4}{c}{Generation Tasks} & \\
\cmidrule(lr){3-9}\cmidrule(lr){10-13}
Sparsity & \shortstack{RoShift\\Config.} & MMLU & ARC-E & ARC-C & BoolQ & PIQA & \shortstack{Wino\\Grande} & CSQA & BBH & GSM8K & \shortstack{MATH-\\500} & MBPP & Avg \\
\midrule
0\% & Dense & 76.86 & 80.01 & 67.58 & 83.03 & 79.38 & 77.66 & 85.91 & 77.64 & 85.44 & 57.40 & 69.80 & 76.43 \\
\midrule
\multirow{2}{*}{\shortstack[l]{Uniform\\60\%}} & Approx. & 71.12 & 80.35 & 62.37 & 82.66 & 77.53 & \textbf{72.77} & \textbf{79.44} & \textbf{70.65} & 80.14 & 45.20 & \textbf{63.20} & 71.40 \\
 & Exact & \textbf{71.44} & \textbf{80.43} & \textbf{63.82} & \textbf{82.97} & \textbf{78.67} & 71.98 & 79.12 & 70.56 & \textbf{81.20} & \textbf{45.40} & 62.00 & \textbf{71.60} \\
\midrule
\multirow{2}{*}{\shortstack[l]{Non-uniform\\60\%}} & Approx. & 73.13 & \textbf{81.19} & \textbf{65.70} & \textbf{82.81} & 78.62 & \textbf{75.69} & 80.84 & \textbf{71.96} & \textbf{81.65} & 48.00 & \textbf{64.20} & 73.07 \\
 & Exact & \textbf{73.76} & \textbf{81.19} & 65.44 & 82.48 & \textbf{78.94} & 75.37 & \textbf{82.88} & 71.92 & 80.67 & \textbf{49.40} & 63.00 & \textbf{73.19} \\
\bottomrule\end{tabular*}
\endgroup\end{minipage}

\clearpage
\Needspace{12\baselineskip}
\subsubsection{Baseline Comparisons with Shared Calibration Data}
\label{app:calibration-data}

To assess whether RoShift's accuracy advantage persists under shared calibration data, we recalibrated the baselines using RoShift's Mixed dataset and compared the results with their original-calibration configurations. All configurations use 60\% target sparsity with uniform allocation to exclude the effect of sparsity-allocation policy search. Tables~\ref{tab:calibration-llama31}--\ref{tab:calibration-qwen3} report the results.

With the same Mixed calibration data and uniform sparsity allocation, RoShift achieved an average accuracy of 59.09\% across five models, exceeding TEAL (54.60\%) and LaRoSA (54.10\%) by 4.49 and 4.99 percentage points, respectively.

We additionally compared each method's original calibration dataset with the common Mixed dataset for Llama-3.1-8B at 60\% target sparsity with non-uniform allocation (Table~\ref{tab:calibration-llama31-nonuniform}). After aligning calibration data, RoShift retained the highest benchmark-average accuracy of 58.54\%, outperforming the second-best method, TEAL (52.75\%), by 5.79 percentage points.

\medskip
\noindent\begin{minipage}{\linewidth}\centering
\captionsetup{font=tablefont,skip=5pt}
\captionof{table}{Llama-3.1-8B: calibration-data comparison at 60\% target sparsity with uniform allocation. All values are accuracy (\%). Avg is the unweighted arithmetic mean over 11 benchmarks. Bold indicates the highest accuracy among the sparse configurations in the table.}
\label{tab:calibration-llama31}
\begingroup
\fontsize{7.5}{8.2}\selectfont
\setlength{\tabcolsep}{1pt}
\renewcommand{\arraystretch}{1.12}
\begin{tabular*}{\linewidth}{@{\extracolsep{\fill}}ll*{12}{c}@{}}
\toprule
 & & \multicolumn{7}{c}{Log-Likelihood-Based Tasks} & \multicolumn{4}{c}{Generation Tasks} & \\
\cmidrule(lr){3-9}\cmidrule(lr){10-13}
Method & Calibration & MMLU & ARC-E & ARC-C & BoolQ & PIQA & \shortstack{Wino\\Grande} & CSQA & BBH & GSM8K & \shortstack{MATH-\\500} & MBPP & Avg \\
\midrule
Dense & -- & 65.57 & 82.49 & 58.28 & 83.12 & 81.12 & 78.22 & 70.60 & 64.52 & 48.90 & 21.60 & 48.20 & 63.87 \\
\midrule
TEAL & Alpaca & 48.63 & 73.15 & 43.86 & 75.50 & 76.06 & 65.75 & 50.45 & 50.64 & 26.69 & 10.40 & 34.20 & 50.48 \\
TEAL & Mixed & 46.96 & 71.80 & 43.94 & 73.55 & 74.37 & 65.51 & 47.91 & 48.96 & 25.25 & 10.00 & 33.00 & 49.20 \\
LaRoSA & WikiText-2 & 47.00 & 70.75 & 45.05 & 76.02 & 75.08 & 66.69 & 50.53 & 47.78 & 17.36 & 7.60 & 28.60 & 48.41 \\
LaRoSA & Mixed & 48.83 & 72.18 & 45.05 & 76.67 & 75.14 & 65.98 & 53.07 & 48.75 & 24.41 & 11.20 & 33.80 & 50.46 \\
RoShift & Mixed & \textbf{56.52} & \textbf{76.81} & \textbf{52.90} & \textbf{79.72} & \textbf{77.97} & \textbf{72.61} & \textbf{61.59} & \textbf{55.58} & \textbf{36.54} & \textbf{16.60} & \textbf{40.80} & \textbf{57.06} \\
\bottomrule\end{tabular*}
\endgroup\end{minipage}
\par\bigskip

\noindent\begin{minipage}{\linewidth}\centering
\captionsetup{font=tablefont,skip=5pt}
\captionof{table}{Llama-2-7B: calibration-data comparison at 60\% target sparsity with uniform allocation. All values are accuracy (\%). Avg is the unweighted arithmetic mean over 11 benchmarks. Bold indicates the highest accuracy among the sparse configurations in the table.}
\label{tab:calibration-llama2}
\begingroup
\fontsize{7.5}{8.2}\selectfont
\setlength{\tabcolsep}{1pt}
\renewcommand{\arraystretch}{1.12}
\begin{tabular*}{\linewidth}{@{\extracolsep{\fill}}ll*{12}{c}@{}}
\toprule
 & & \multicolumn{7}{c}{Log-Likelihood-Based Tasks} & \multicolumn{4}{c}{Generation Tasks} & \\
\cmidrule(lr){3-9}\cmidrule(lr){10-13}
Method & Calibration & MMLU & ARC-E & ARC-C & BoolQ & PIQA & \shortstack{Wino\\Grande} & CSQA & BBH & GSM8K & \shortstack{MATH-\\500} & MBPP & Avg \\
\midrule
Dense & -- & 45.96 & 73.86 & 52.65 & 79.33 & 78.73 & 74.66 & 32.76 & 39.69 & 14.56 & 3.20 & 23.20 & 47.14 \\
\midrule
TEAL & Alpaca & 33.49 & 67.72 & 42.15 & 71.04 & 75.14 & 67.09 & 23.26 & 33.88 & 6.44 & \textbf{4.20} & 13.60 & 39.82 \\
TEAL & Mixed & 32.79 & 66.92 & 44.03 & 69.72 & 75.79 & 64.25 & 24.65 & 33.70 & 5.84 & 3.60 & 13.60 & 39.54 \\
LaRoSA & Mixed & 35.23 & 69.36 & 44.37 & 73.64 & 75.08 & 65.98 & 25.39 & 35.57 & 9.33 & 3.20 & 14.00 & 41.01 \\
RoShift & Mixed & \textbf{38.22} & \textbf{70.58} & \textbf{47.18} & \textbf{75.17} & \textbf{76.33} & \textbf{70.96} & \textbf{27.76} & \textbf{37.00} & \textbf{11.22} & 3.60 & \textbf{18.40} & \textbf{43.31} \\
\bottomrule\end{tabular*}
\endgroup\end{minipage}
\par\bigskip

\noindent\begin{minipage}{\linewidth}\centering
\captionsetup{font=tablefont,skip=5pt}
\captionof{table}{Mistral-7B: calibration-data comparison at 60\% target sparsity with uniform allocation. All values are accuracy (\%). Avg is the unweighted arithmetic mean over 11 benchmarks. Bold indicates the highest accuracy among the sparse configurations in the table.}
\label{tab:calibration-mistral}
\begingroup
\fontsize{7.5}{8.2}\selectfont
\setlength{\tabcolsep}{1pt}
\renewcommand{\arraystretch}{1.12}
\begin{tabular*}{\linewidth}{@{\extracolsep{\fill}}ll*{12}{c}@{}}
\toprule
 & & \multicolumn{7}{c}{Log-Likelihood-Based Tasks} & \multicolumn{4}{c}{Generation Tasks} & \\
\cmidrule(lr){3-9}\cmidrule(lr){10-13}
Method & Calibration & MMLU & ARC-E & ARC-C & BoolQ & PIQA & \shortstack{Wino\\Grande} & CSQA & BBH & GSM8K & \shortstack{MATH-\\500} & MBPP & Avg \\
\midrule
Dense & -- & 62.65 & 79.38 & 61.26 & 84.40 & 82.54 & 78.93 & 59.13 & 56.70 & 39.58 & 13.00 & 38.80 & 59.67 \\
\midrule
TEAL & Alpaca & 52.40 & 73.86 & 52.13 & 81.01 & 79.38 & 71.03 & 47.01 & 49.32 & 23.65 & 6.40 & 29.40 & 51.42 \\
TEAL & Mixed & 53.23 & 74.45 & 51.45 & 80.34 & 78.67 & 71.43 & 47.34 & 48.26 & 20.39 & 6.60 & 29.80 & 51.09 \\
LaRoSA & WikiText-2 & 49.26 & 73.32 & 48.81 & 78.69 & 78.35 & 70.96 & 39.89 & 45.71 & 17.97 & 6.20 & 22.80 & 48.36 \\
LaRoSA & Mixed & 50.11 & 74.33 & 50.43 & 80.28 & 76.93 & 70.80 & 47.99 & 47.89 & 21.46 & 7.40 & 27.00 & 50.42 \\
RoShift & Mixed & \textbf{57.16} & \textbf{77.74} & \textbf{56.57} & \textbf{84.10} & \textbf{80.25} & \textbf{76.01} & \textbf{56.67} & \textbf{52.85} & \textbf{33.06} & \textbf{9.40} & \textbf{35.80} & \textbf{56.33} \\
\bottomrule\end{tabular*}
\endgroup\end{minipage}
\par\bigskip

\noindent\begin{minipage}{\linewidth}\centering
\captionsetup{font=tablefont,skip=5pt}
\captionof{table}{Qwen2.5-7B: calibration-data comparison at 60\% target sparsity with uniform allocation. All values are accuracy (\%). Avg is the unweighted arithmetic mean over 11 benchmarks. Bold indicates the highest accuracy among the sparse configurations in the table.}
\label{tab:calibration-qwen25}
\begingroup
\fontsize{7.5}{8.2}\selectfont
\setlength{\tabcolsep}{1pt}
\renewcommand{\arraystretch}{1.12}
\begin{tabular*}{\linewidth}{@{\extracolsep{\fill}}ll*{12}{c}@{}}
\toprule
 & & \multicolumn{7}{c}{Log-Likelihood-Based Tasks} & \multicolumn{4}{c}{Generation Tasks} & \\
\cmidrule(lr){3-9}\cmidrule(lr){10-13}
Method & Calibration & MMLU & ARC-E & ARC-C & BoolQ & PIQA & \shortstack{Wino\\Grande} & CSQA & BBH & GSM8K & \shortstack{MATH-\\500} & MBPP & Avg \\
\midrule
Dense & -- & 74.23 & 77.40 & 63.74 & 84.68 & 79.76 & 76.01 & 85.50 & 68.96 & 82.64 & 49.60 & 63.60 & 73.28 \\
\midrule
TEAL & Alpaca & 64.46 & 74.66 & 54.35 & 80.31 & 77.20 & 68.59 & 74.45 & 56.77 & 70.89 & 30.40 & 53.40 & 64.13 \\
TEAL & Mixed & 64.73 & 74.96 & 55.72 & 80.73 & 77.15 & 67.88 & 75.51 & 56.38 & 71.80 & 29.40 & 53.80 & 64.37 \\
LaRoSA & WikiText-2 & 60.99 & 74.92 & 52.99 & 79.11 & 75.63 & 67.72 & 69.37 & 55.97 & 66.03 & 24.80 & 46.40 & 61.27 \\
LaRoSA & Mixed & 61.73 & 75.80 & 53.84 & 78.99 & 76.66 & 67.72 & 72.81 & 56.93 & 67.25 & 30.80 & 50.80 & 63.03 \\
RoShift & Mixed & \textbf{68.41} & \textbf{76.89} & \textbf{58.11} & \textbf{83.09} & \textbf{77.37} & \textbf{72.30} & \textbf{79.69} & \textbf{60.56} & \textbf{74.91} & \textbf{35.20} & \textbf{54.60} & \textbf{67.37} \\
\bottomrule\end{tabular*}
\endgroup\end{minipage}
\par\bigskip

\noindent\begin{minipage}{\linewidth}\centering
\captionsetup{font=tablefont,skip=5pt}
\captionof{table}{Qwen3-8B-Base: calibration-data comparison at 60\% target sparsity with uniform allocation. All values are accuracy (\%). Avg is the unweighted arithmetic mean over 11 benchmarks. Bold indicates the highest accuracy among the sparse configurations in the table.}
\label{tab:calibration-qwen3}
\begingroup
\fontsize{7.5}{8.2}\selectfont
\setlength{\tabcolsep}{1pt}
\renewcommand{\arraystretch}{1.12}
\begin{tabular*}{\linewidth}{@{\extracolsep{\fill}}ll*{12}{c}@{}}
\toprule
 & & \multicolumn{7}{c}{Log-Likelihood-Based Tasks} & \multicolumn{4}{c}{Generation Tasks} & \\
\cmidrule(lr){3-9}\cmidrule(lr){10-13}
Method & Calibration & MMLU & ARC-E & ARC-C & BoolQ & PIQA & \shortstack{Wino\\Grande} & CSQA & BBH & GSM8K & \shortstack{MATH-\\500} & MBPP & Avg \\
\midrule
Dense & -- & 76.86 & 80.01 & 67.58 & 83.03 & 79.38 & 77.66 & 85.91 & 77.64 & 85.44 & 57.40 & 69.80 & 76.43 \\
\midrule
TEAL & Alpaca & 68.94 & 79.46 & 60.92 & 75.54 & 76.77 & 70.17 & 78.05 & 68.71 & 76.12 & 43.60 & 62.20 & 69.13 \\
TEAL & Mixed & 69.45 & 79.25 & 61.35 & 75.57 & \textbf{77.69} & 70.24 & 76.66 & 69.14 & 77.18 & 41.20 & 59.20 & 68.81 \\
LaRoSA & WikiText-2 & 63.03 & 76.68 & 58.36 & 76.51 & 74.70 & 67.32 & 71.83 & 62.43 & 67.17 & 34.20 & 51.80 & 64.00 \\
LaRoSA & Mixed & 63.59 & 78.03 & 57.59 & 77.98 & 75.84 & 67.80 & 74.37 & 63.91 & 73.24 & 37.80 & 51.40 & 65.60 \\
RoShift & Mixed & \textbf{71.12} & \textbf{80.35} & \textbf{62.37} & \textbf{82.66} & 77.53 & \textbf{72.77} & \textbf{79.44} & \textbf{70.65} & \textbf{80.14} & \textbf{45.20} & \textbf{63.20} & \textbf{71.40} \\
\bottomrule\end{tabular*}
\endgroup\end{minipage}
\par\bigskip

\noindent\begin{minipage}{\linewidth}\centering
\captionsetup{font=tablefont,skip=5pt}
\captionof{table}{Llama-3.1-8B: calibration-data comparison at 60\% target sparsity with non-uniform allocation. All values are accuracy (\%). Avg is the unweighted arithmetic mean over 11 benchmarks. Bold indicates the highest accuracy among the sparse configurations in the table.}
\label{tab:calibration-llama31-nonuniform}
\begingroup
\fontsize{7.5}{8.2}\selectfont
\setlength{\tabcolsep}{1pt}
\renewcommand{\arraystretch}{1.12}
\begin{tabular*}{\linewidth}{@{\extracolsep{\fill}}ll*{12}{c}@{}}
\toprule
 & & \multicolumn{7}{c}{Log-Likelihood-Based Tasks} & \multicolumn{4}{c}{Generation Tasks} & \\
\cmidrule(lr){3-9}\cmidrule(lr){10-13}
Method & Calibration & MMLU & ARC-E & ARC-C & BoolQ & PIQA & \shortstack{Wino\\Grande} & CSQA & BBH & GSM8K & \shortstack{MATH-\\500} & MBPP & Avg \\
\midrule
Dense & -- & 65.57 & 82.49 & 58.28 & 83.12 & 81.12 & 78.22 & 70.60 & 64.52 & 48.90 & 21.60 & 48.20 & 63.87 \\
\midrule
TEAL & Alpaca & 52.81 & 74.20 & 49.15 & 78.10 & 76.71 & 70.40 & 56.51 & 52.14 & 30.02 & 11.00 & 36.80 & 53.44 \\
TEAL & Mixed & 51.79 & 74.87 & 49.15 & 77.43 & 76.77 & 69.30 & 54.79 & 50.05 & 28.28 & 11.60 & 36.20 & 52.75 \\
LaRoSA & WikiText-2 & 46.24 & 69.74 & 43.69 & 75.75 & 74.97 & 64.48 & 48.48 & 46.92 & 17.51 & 7.60 & 27.00 & 47.49 \\
LaRoSA & Mixed & 50.82 & 72.77 & 48.89 & 73.88 & 76.61 & 67.80 & 54.38 & 49.61 & 26.69 & 10.40 & 33.40 & 51.39 \\
WINA & Alpaca & 48.99 & 72.39 & 47.95 & 78.72 & 76.22 & 68.98 & 53.32 & 49.02 & 21.00 & 7.20 & 33.00 & 50.62 \\
WINA & Mixed & 50.09 & 73.32 & 49.74 & 78.26 & 77.20 & 68.43 & 54.46 & 48.40 & 19.26 & 9.40 & 32.80 & 51.03 \\
RoShift & Mixed & \textbf{58.88} & \textbf{78.66} & \textbf{52.47} & \textbf{81.93} & \textbf{78.62} & \textbf{73.80} & \textbf{66.83} & \textbf{56.17} & \textbf{39.80} & \textbf{15.00} & \textbf{41.80} & \textbf{58.54} \\
\bottomrule\end{tabular*}
\endgroup\end{minipage}

\clearpage
\Needspace{12\baselineskip}
\subsubsection{Correction-inclusive MAC-Budget Comparison}
\label{app:correction-inclusive-mac-budget}

This section compares accuracy with TEAL under the same MAC budget. Using the measured Transformer-projection MAC sparsity of TEAL at a 50\% target, we adjusted RoShift's target sparsity to match the budget including its additional correction and transition operations. To isolate the effects of Transformer sparsification, we kept RoShift's LM-head dense and included the matrix operations required for transition and fused correction in the budget, increasing Transformer sparsity to offset these costs. We assumed a 90\% fused-path rate and a 10\% fallback rate, accounting for an average of $1.1d^2$ MACs per layer for correction and transition, where $d$ is the model dimension. Both methods use non-uniform sparsity allocation.

Even after accounting for these additional costs, RoShift achieved an average accuracy of 61.22\% across five models and 11 benchmarks, compared with 60.89\% for TEAL, a gain of 0.33 percentage points. The per-model RoShift-minus-TEAL accuracy differences ranged from $-0.57$ to 1.21 percentage points, with TEAL achieving higher accuracy on Llama-2 (Table~\ref{tab:correction-inclusive-accuracy}).

\medskip
\noindent\begin{minipage}{\linewidth}
\centering
\captionsetup{font=tablefont,skip=5pt}
\captionof{table}{Accuracy comparison under MAC budgets including correction and transition costs. Each model's accuracy is the unweighted arithmetic mean over 11 benchmarks, and Avg is the mean over five models. TEAL Actual Sparsity is the measured Transformer-projection MAC sparsity at a 50\% target. Qwen3-8B accuracy is shown to three decimal places to display the small difference.}
\label{tab:correction-inclusive-accuracy}
\begingroup
\fontsize{8}{9}\selectfont
\setlength{\tabcolsep}{3pt}
\renewcommand{\arraystretch}{1.12}
\begin{tabular}{@{}l*{6}{c}@{}}
\toprule
Method & Llama-3.1-8B & Llama-2-7B & Mistral-7B & Qwen2.5-7B & Qwen3-8B & Avg \\
\midrule
TEAL Actual Sparsity & 48.24\% & 49.43\% & 49.42\% & 48.18\% & 47.89\% & -- \\
RoShift Target Sparsity & 56.70\% & 58.55\% & 57.88\% & 54.24\% & 57.46\% & -- \\
\midrule
TEAL Avg.\ Accuracy & 59.21 & \textbf{44.34} & 56.95 & 70.46 & 73.489 & 60.89 \\
RoShift Avg.\ Accuracy & \textbf{59.37} & 43.77 & \textbf{58.16} & \textbf{71.29} & \textbf{73.494} & \textbf{61.22} \\
\bottomrule
\end{tabular}
\endgroup
\end{minipage}
\medskip

This comparison matches MAC budgets rather than measured bytes read per token. Moreover, equal bytes read per token do not necessarily imply equal GPU execution efficiency. RoShift is designed to improve GPU execution efficiency by reusing Up output sparsity in Down, keeping per-block workloads consistent with B128 Blockwise Top-K, and executing correction in parallel with the sparse FFN. These structural advantages can contribute to GPU execution efficiency beyond reducing memory reads; therefore, measured end-to-end decoding performance should be considered alongside MAC counts and memory-read volume.

To examine actual decoding performance, we also evaluated end-to-end decoding speed in a separate uniform configuration targeting similar MAC budgets (Table~\ref{tab:correction-inclusive-speedup}). RoShift achieved higher speedups than TEAL on four models, while both methods were equal on Qwen2.5 to two decimal places. Although the accuracy and speed evaluations use different configurations, RoShift shows higher average accuracy in the former and generally higher decoding speedup in the latter in these MAC-budget-aware comparisons.

\medskip
\noindent\begin{minipage}{\linewidth}
\centering
\captionsetup{font=tablefont,skip=5pt}
\captionof{table}{End-to-end decoding speedup relative to Dense on an RTX 5090 with FP16, batch size 1, a dense LM-head, and uniform sparsity. Sparsity values are obtained from separate instrumentation. Bold indicates higher speedup.}
\label{tab:correction-inclusive-speedup}
\begingroup
\fontsize{8}{9}\selectfont
\setlength{\tabcolsep}{3pt}
\renewcommand{\arraystretch}{1.12}
\begin{tabular}{@{}l*{5}{c}@{}}
\toprule
Method & Llama-3.1-8B & Llama-2-7B & Mistral-7B & Qwen2.5-7B & Qwen3-8B \\
\midrule
TEAL Actual Sparsity & 48.25\% & 49.44\% & 49.42\% & 48.18\% & 47.87\% \\
RoShift Uniform Sparsity & 57.03\% & 58.59\% & 57.81\% & 53.91\% & 57.81\% \\
\midrule
TEAL Speedup & 1.55$\times$ & 1.57$\times$ & 1.62$\times$ & \textbf{1.60$\times$} & 1.43$\times$ \\
RoShift Speedup & \textbf{1.65$\times$} & \textbf{1.65$\times$} & \textbf{1.70$\times$} & \textbf{1.60$\times$} & \textbf{1.55$\times$} \\
\bottomrule
\end{tabular}
\endgroup
\end{minipage}

\clearpage
\endgroup

\clearpage
\subsection{Complete Quantization Results}
\label{app:complete-quantization-results}

Table~\ref{tab:full-8bit-quantization-results} reports the 8-bit quantization results for RoShift. Tables~\ref{tab:full-4bit-quantization-q8-results} and \ref{tab:full-4bit-quantization-q16-results} report RoShift's 4-bit weight quantization results with the transition matrices $Q_l^\top Q_{l+1}$ retained in 8 and 16 bits, respectively. Table~\ref{tab:full-4bit-quantization-results} compares TEAL, LaRoSA, and RoShift under 4-bit quantization. For Table~\ref{tab:full-4bit-quantization-results}, unlike the public implementation used in the accuracy experiments, we implement LaRoSA following its paper formulation, retaining activations in the rotated basis and applying \(Q_l^\top Q_{l+1}\) between layers. All tables report PPL on Llama-3.1-8B using RTN, GPTQ, AWQ, and OmniQuant, measured on the datasets specified in Appendix~\ref{app:kernel-details}.
We evaluate all sparsified models under uniform sparsity.

With 8-bit quantization, the PPL gap between sparse RoShift and Dense remains nearly unchanged across all quantization methods, demonstrating high compatibility between sparsification and 8-bit quantization.

In contrast, compatibility with 4-bit quantization depends strongly on the precision of the transition matrices $Q_l^\top Q_{l+1}$. TEAL, which does not use transition matrices, retains a similar PPL gap and is robust to quantization. LaRoSA and RoShift, however, degrade substantially across all quantization methods when both weights and transition matrices are quantized to 4 bits. We attribute this sensitivity to quantization errors in the transition matrices, which act on the entire rotated FFN input residual $X_l$, allowing errors to accumulate across layers.

Keeping RoShift's transition matrices at 8 bits while quantizing the weights to 4 bits yields PPL comparable to that obtained without quantizing the transition matrices across all evaluated quantizers. This demonstrates that 8-bit precision for the transition matrices largely preserves model performance under 4-bit weight quantization.

\medskip
\noindent\begin{minipage}{\textwidth}
\begingroup
    \centering
    \captionsetup{font=tablefont,skip=5pt,justification=justified,singlelinecheck=false}
    \captionof{table}{PPL comparison of RoShift across 8-bit quantization methods.}
    \label{tab:full-8bit-quantization-results}
    \fontsize{8}{9}\selectfont
    \setlength{\tabcolsep}{2pt}
    \renewcommand{\arraystretch}{1.1}
    \begin{tabular*}{0.65\linewidth}{@{\extracolsep{\fill}}l*{5}{c}@{}}
        \toprule
        Method & \shortstack{Baseline\\(16-bit)} & RTN & GPTQ & AWQ & OmniQuant \\
        \midrule
        RoShift 0\% & 4.99 & 5.07 & 5.07 & 5.05 & 5.07 \\
        \addlinespace[3pt]
        RoShift 50\% & 5.82 & 5.73
          & 5.75 & 5.72 & 5.73 \\
        \bottomrule
    \end{tabular*}\par
\endgroup

\medskip
\begingroup
    \centering
    \captionsetup{font=tablefont,skip=5pt,justification=justified,singlelinecheck=false}
    \captionof{table}{PPL comparison of RoShift under 4-bit weight quantization with transition matrices retained in 8-bit.}
    \label{tab:full-4bit-quantization-q8-results}
    \fontsize{8}{9}\selectfont
    \setlength{\tabcolsep}{2pt}
    \renewcommand{\arraystretch}{1.1}
    \begin{tabular*}{0.65\linewidth}{@{\extracolsep{\fill}}l*{5}{c}@{}}
        \toprule
        Method & \shortstack{Baseline\\(16-bit)} & RTN & GPTQ & AWQ & OmniQuant \\
        \midrule
        RoShift 0\% & 4.99 & 5.57 & 5.56 & 5.37 & 5.52 \\
        \addlinespace[3pt]
        RoShift 50\% & 5.82 & 6.24
          & 6.19 & 5.97 & 6.14 \\
        \bottomrule
    \end{tabular*}\par
\endgroup

\medskip
\begingroup
    \centering
    \captionsetup{font=tablefont,skip=5pt,justification=justified,singlelinecheck=false}
    \captionof{table}{PPL comparison of RoShift under 4-bit weight quantization with transition matrices retained in 16-bit.}
    \label{tab:full-4bit-quantization-q16-results}
    \fontsize{8}{9}\selectfont
    \setlength{\tabcolsep}{2pt}
    \renewcommand{\arraystretch}{1.1}
    \begin{tabular*}{0.65\linewidth}{@{\extracolsep{\fill}}l*{5}{c}@{}}
        \toprule
        Method & \shortstack{Baseline\\(16-bit)} & RTN & GPTQ & AWQ & OmniQuant \\
        \midrule
        RoShift 0\% & 4.99 & 5.50 & 5.48 & 5.29 & 5.38 \\
        \addlinespace[3pt]
        RoShift 50\% & 5.82 & 6.15
          & 6.09 & 5.91 & 6.04 \\
        \bottomrule
    \end{tabular*}\par
\endgroup

\medskip
\begingroup
    \centering
    \captionsetup{font=tablefont,skip=5pt,justification=justified,singlelinecheck=false}
    \captionof{table}{PPL comparison of TEAL, LaRoSA, and RoShift under 4-bit quantization.}
    \label{tab:full-4bit-quantization-results}
    \fontsize{8}{9}\selectfont
    \setlength{\tabcolsep}{1pt}
    \renewcommand{\arraystretch}{1.1}
    \begin{tabular*}{0.65\linewidth}{@{\extracolsep{\fill}}l*{5}{c}@{}}
        \toprule
        Method & \shortstack{Baseline\\(16-bit)} & RTN & GPTQ & AWQ & OmniQuant \\
        \midrule
        TEAL 0\% & 4.99 & 5.50 & 5.48 & 5.29 & 5.38 \\
        \addlinespace[3pt]
        TEAL 50\% & 6.22 & 6.62
          & 6.78 & 6.92 & 6.68 \\
        \midrule
        LaRoSA 0\% & 4.99 & 5528.44 & 336.47 & 4447.38 & 1464.26 \\
        \addlinespace[3pt]
        LaRoSA 50\% & 6.16 & 7368.24
          & 1415.94 & 7593.46
          & 6201.85 \\
        \midrule
        RoShift 0\% & 4.99 & 331.88 & 231.45 & 2331.75 & 392.60 \\
        \addlinespace[3pt]
        RoShift 50\% & 5.82 & 481.92
          & 349.68 & 1187.89
          & 282.01 \\
        \bottomrule
    \end{tabular*}\par
\endgroup

\end{minipage}

\clearpage
\subsection{LaRoSA Comparison and Shift Correction Analysis}
\label{app:larosa-correction-ablation}

\subsubsection{Experimental Setup and Configurations}
\label{app:larosa-correction-setup}

This section analyzes decoding latency relative to LaRoSA, the cost of RoShift's shift correction, and the effects of correction fusion and parallel execution in reducing this cost.

LaRoSA's publicly released code does not include the optimized sparse GEMV kernels and decoding acceleration implementation used for its reported speed measurements~\citep{liu2025larosa}. We therefore implement LaRoSA using RoShift's GPU kernel backend. RoShift's backend supports only B128 Blockwise Top-K, whereas LaRoSA's original paper uses a kernel that fuses Top-K selection with sparse GEMV. Differences may also exist in sparse mask generation and GEMV implementation, so the results below may differ from the performance of LaRoSA's original implementation.

Experiments use an RTX~5090 GPU, FP16, batch size~1, and 50\% uniform sparsity, with a dense LM-head in all configurations. For the RoShift ablation, we perform a 1,024-token dense prefill for each of four prompts, then measure latency over 128 teacher-forced decode steps and average the results across three repetitions. Prefill time is excluded.

To distinguish the effects of shift correction and execution optimizations, we compare four configurations:
\begin{itemize}
    \item \textbf{Correction Off:} Shift correction is omitted. This timing-only configuration is not used for accuracy comparison.
    \item \textbf{Unfused Serial:} Correction and transition are not fused and execute serially with the sparse FFN.
    \item \textbf{Fused Serial:} Correction and transition are fused but execute serially with the sparse FFN.
    \item \textbf{Fused Parallel:} Correction fusion and parallel execution with the sparse FFN on separate CUDA streams are both enabled.
\end{itemize}

\subsubsection{LaRoSA Comparison and RoShift Ablation}
\label{app:larosa-correction-results}

\paragraph{Comparison with the Implemented LaRoSA.} Table~\ref{tab:larosa-correction-latency} compares the end-to-end decoding latency of LaRoSA implemented using RoShift's backend with that of RoShift, including its execution-optimization configurations. All values are in ms/token, and lower is better.

\medskip
\noindent\begin{minipage}{\linewidth}
\centering
\captionof{table}{End-to-end decoding latency (ms/token) at 50\% uniform sparsity with a dense LM-head. Lower is better.}
\label{tab:larosa-correction-latency}
\small
\setlength{\tabcolsep}{4pt}
\renewcommand{\arraystretch}{1.1}
\begin{tabular*}{\linewidth}{@{\extracolsep{\fill}}lrrrrrr@{}}
\toprule
Model & Dense & \shortstack{LaRoSA\\(RoShift\\backend)} & \shortstack{Correction\\Off} & \shortstack{Unfused\\Serial} & \shortstack{Fused\\Serial} & \shortstack{Fused\\Parallel} \\
\midrule
Llama-3.1-8B & 9.94 & 6.91 & 6.91 & 7.64 & 6.95 & \textbf{6.75} \\
Llama-2-7B & 9.56 & 6.39 & 6.31 & 7.07 & 6.38 & \textbf{6.18} \\
Mistral-7B & 10.56 & 6.42 & 6.37 & 7.11 & 6.44 & \textbf{6.23} \\
Qwen2.5-7B & 10.34 & 6.34 & 6.30 & 6.79 & 6.38 & \textbf{6.19} \\
Qwen3-8B & 10.26 & \textbf{7.29} & 7.57 & 8.30 & 7.59 & 7.35 \\
\bottomrule
\end{tabular*}
\end{minipage}
\medskip

With both correction fusion and parallel execution, RoShift achieves lower decoding latency than LaRoSA implemented on the shared backend on four of the five models. LaRoSA has lower latency on Qwen3-8B.

\paragraph{RoShift Ablation.} Separately from the LaRoSA comparison, we analyze the additional cost of shift correction within RoShift and the effects of the two execution optimizations that reduce it. The change from Correction Off to Unfused Serial shows the latency increase when correction is added; the change from Unfused Serial to Fused Serial shows the effect of fusion; and the change from Fused Serial to Fused Parallel shows the effect of parallel execution.

Table~\ref{tab:correction-ablation-deltas} summarizes the latency differences between these configurations. Values are computed from the latencies displayed to two decimal places in Table~\ref{tab:larosa-correction-latency} and are reported in ms/token.

\medskip
\noindent\begin{minipage}{\linewidth}
\centering
\captionof{table}{Latency differences between RoShift configurations (ms/token), computed from the displayed values in Table~\ref{tab:larosa-correction-latency}.}
\label{tab:correction-ablation-deltas}
\fontsize{9}{10.8}\selectfont
\setlength{\tabcolsep}{10pt}
\renewcommand{\arraystretch}{1.1}
\begin{tabular}{lrrr}
\toprule
Model & Correction Overhead & Fusion Reduction & Parallel Reduction \\
\midrule
Llama-3.1-8B & 0.73 & 0.69 & 0.20 \\
Llama-2-7B & 0.76 & 0.69 & 0.20 \\
Mistral-7B & 0.74 & 0.67 & 0.21 \\
Qwen2.5-7B & 0.49 & 0.41 & 0.19 \\
Qwen3-8B & 0.73 & 0.71 & 0.24 \\
\bottomrule
\end{tabular}
\end{minipage}
\medskip

Executing shift correction separately increases decoding latency by 0.49--0.76~ms/token. Correction fusion reduces latency by 0.41--0.71~ms/token, offsetting most of this additional cost, while parallel execution reduces latency by a further 0.19--0.24~ms/token. This ablation quantifies the effects of correction fusion and parallel execution in reducing the additional cost of shift correction. With both optimizations, RoShift achieves competitive decoding latency relative to LaRoSA implemented on the shared backend.

\subsection{End-to-End Accuracy--Speedup Trade-offs under Different LM-head Settings}
\label{app:head-controlled-pareto}

\begin{figure}[!htb]
    \centering
    \includegraphics[width=\linewidth]{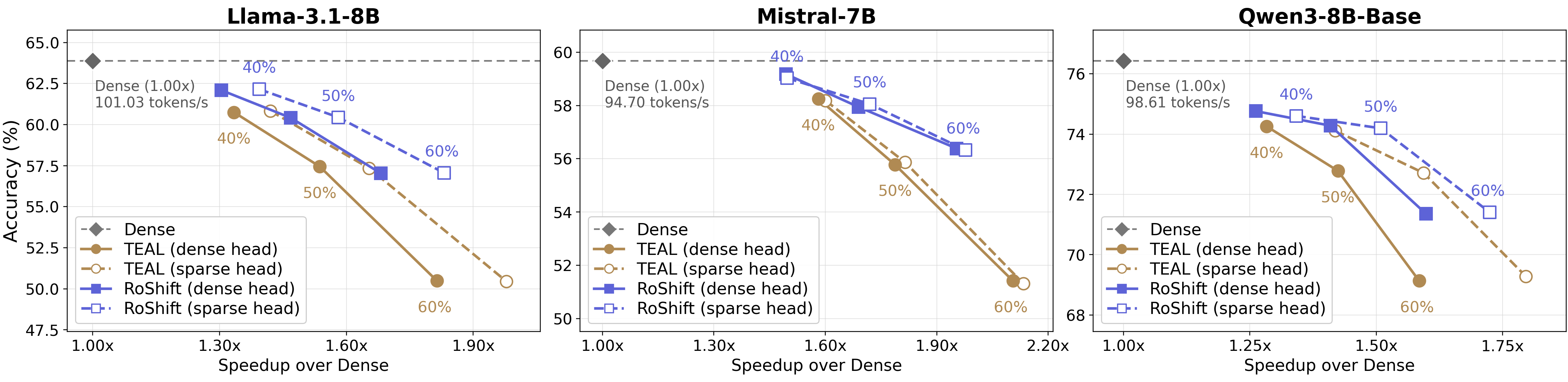}
    \captionsetup{font=small,skip=3pt}
    \caption{End-to-end accuracy--speedup trade-offs for TEAL and RoShift with dense and rotated sparse LM-heads on an RTX~5090 at batch size~1. Accuracy is averaged over 11 benchmarks; labels indicate target uniform sparsity.}
    \label{fig:head-controlled-pareto}
\end{figure}

To distinguish the effect of LM-head sparsification on the end-to-end accuracy--speedup trade-off, we compare TEAL and RoShift under two LM-head settings. We evaluate both methods with a dense LM-head and with the same rotated sparse LM-head. Figure~\ref{fig:head-controlled-pareto} shows that RoShift offers an overall better accuracy--speedup trade-off than TEAL under both settings. This indicates that RoShift's end-to-end advantage does not depend solely on LM-head sparsification.

\subsection{Peak GPU Memory Usage}
\label{app:peak-gpu-memory}

Table~\ref{tab:peak-gpu-memory} compares the peak GPU memory usage of Dense and RoShift across five models. Measurements use batch size~1, a 4K-token prefill, and one decode step, with 50\% uniform sparsity for RoShift. For all models, peak memory usage occurs during prefill for both Dense and RoShift.

\medskip
\noindent\begin{minipage}{\linewidth}
\centering
\captionof{table}{Peak GPU memory usage of Dense and RoShift across five models.}
\label{tab:peak-gpu-memory}
\fontsize{9}{10.8}\selectfont
\setlength{\tabcolsep}{5pt}
\begin{tabular}{lrrrrr}
\toprule
Memory Metric & \shortstack{Llama3.1\\8B} & \shortstack{Llama2\\7B} & \shortstack{Mistral\\7B} & \shortstack{Qwen2.5\\7B} & \shortstack{Qwen3\\8B} \\
\midrule
Dense Weights (GiB) & 14.96 & 12.55 & 13.49 & 14.19 & 15.26 \\
$A$: Dense Peak (GiB) & 20.48 & 19.62 & 19.01 & 18.86 & 20.85 \\
RoShift Peak (GiB) & 26.49 & 25.63 & 25.02 & 22.87 & 27.60 \\
$B$: Additional Memory (GiB) & 6.00 & 6.00 & 6.00 & 4.01 & 6.75 \\
$B/A$: \textbf{Increase (\%)} & \textbf{29.31\%} & \textbf{30.59\%} & \textbf{31.57\%} & \textbf{21.27\%} & \textbf{32.40\%} \\
\bottomrule
\end{tabular}
\end{minipage}
\medskip

\subsection{Decoding Speed and Fallback in Free-running Generation}
\label{app:free-running-generation}

\begin{figure}[!htb]
    \centering
    \includegraphics[width=\linewidth]{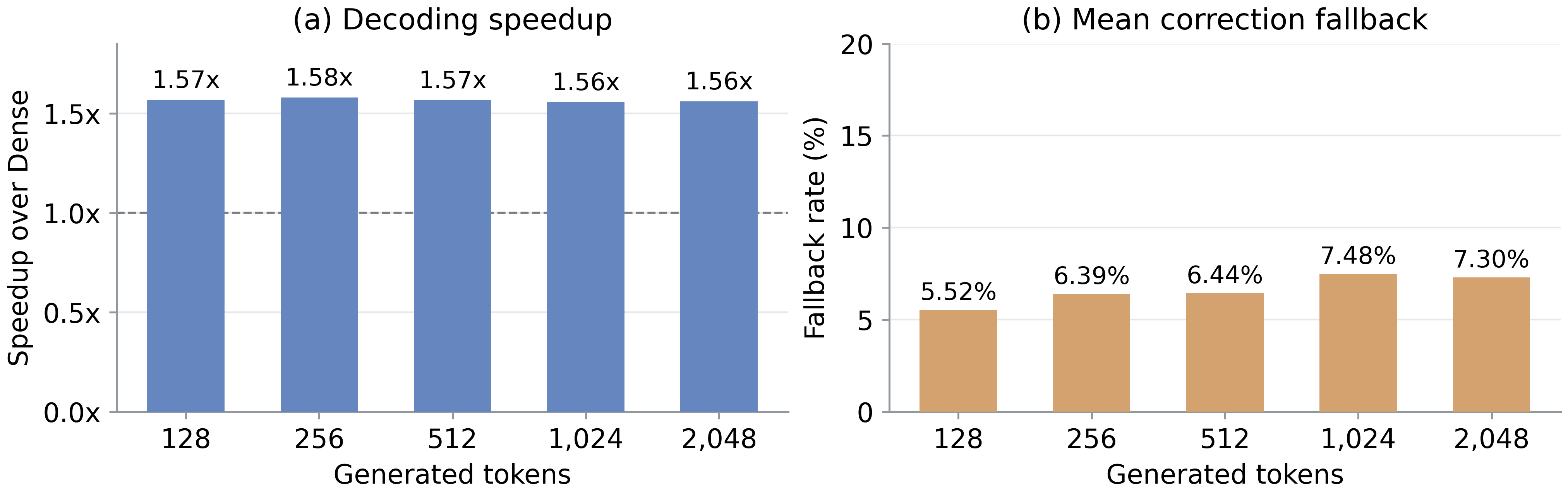}
    \caption{(a) Decoding speedup over Dense and (b) mean correction fallback rate of Llama-3.1-8B across generation lengths. Measurements used an RTX~5090, FP16, batch size~1, and 50\% uniform sparsity for RoShift. The 16 prompts comprise the four original prompts plus three additional prompts from each of the four generation benchmarks. Decoding speed was measured over three repetitions per configuration, excluding dense prefill time. Fallback rates were obtained from separate replay diagnostics.}
    \label{fig:free-running-generation}
\end{figure}

To assess whether RoShift's decoding speedup is maintained during free-running generation across diverse prompts, we analyzed how speedup and correction fallback rates vary with generation length. As shown in Figure~\ref{fig:free-running-generation}, RoShift maintained a 1.56--1.58$\times$ speedup over Dense as the generation length increased from 128 to 2,048 tokens. The mean fallback rate ranged from 5.52\% to 7.48\%, remaining below 10\% at all evaluated generation lengths.

However, it is difficult to precisely control TEAL's achieved sparsity during free-running generation. We therefore used teacher-forced continuations with the same token sequences for the cross-method speed comparisons in the main text.

\end{document}